\documentclass[twoside,11pt]{article}
\usepackage[T1]{fontenc}
\usepackage{jair, theapa, rawfonts}
\usepackage{wrapfig}
\usepackage{subcaption}
\usepackage{booktabs}       
\usepackage{microtype}      
\usepackage{nicefrac}       
\usepackage[usenames, dvipsnames]{color}
\usepackage{amsfonts}        
\usepackage{amsmath,amssymb} 
\usepackage{mathtools}
\usepackage{algorithm}
\usepackage[noend]{algpseudocode}
\usepackage{graphicx}
\graphicspath{ {images/} }

\ShortHeadings{Optimizing for Interpretability in Deep Neural Networks with Tree Regularization}
{Wu et. al.}

\begin{document}

\title{Optimizing for Interpretability in Deep Neural Networks with Tree Regularization}

\author{\name Mike Wu \email wumike@stanford.edu \\
        \addr Stanford University,
        Stanford, CA 94305 USA
        \AND
       \name Sonali Parbhoo \email sonali.parbhoo@unibas.ch \\
       \addr University of Basel,
       Basel, Switzerland
       \AND
       \name Michael C. Hughes \email mike@michaelchughes.com \\
       \addr Tufts University,
       Medford, MA 02153 USA
       \AND
       \name Volker Roth \email volker.roth@unibas.ch \\
       \addr University of Basel,
       Basel, Switzerland
       \AND
       \name Finale Doshi-Velez \email finale@seas.harvard.edu \\
       \addr Harvard University SEAS,
       Cambridge, MA 02138 USA}

\maketitle

\begin{abstract}
  \noindent Deep models have advanced prediction in many domains, but their lack of interpretability remains a key barrier to the adoption in many real world applications. There exists a large body of work aiming to help humans understand these black box functions to varying levels of granularity -- for example, through distillation, gradients, or adversarial examples. These methods however, all tackle interpretability as a separate process after training. In this work, we take a different approach and explicitly regularize deep models so that they are well-approximated by processes that humans can step-through in little time.  Specifically, we train several families of deep neural networks to resemble compact, axis-aligned decision trees without significant compromises in accuracy. The resulting \textit{axis-aligned} decision functions uniquely make tree regularized models easy for humans to interpret. Moreover, for situations in which a single, global tree is a poor estimator, we introduce a \emph{regional} tree regularizer that encourages the deep model to resemble a compact, axis-aligned decision tree in predefined, human-interpretable contexts. Using intuitive toy examples as well as medical tasks for patients in critical care and with HIV, we demonstrate that this new family of tree regularizers yield models that are
  easier for humans to simulate than simpler L1 or L2 penalties without sacrificing predictive power.
\end{abstract}



\section{Introduction}
\label{Introduction}

Deep models have become the de-facto approach for prediction in many applications like image classification (e.g. \cite{krizhevsky2012imagenet}) and machine translation (e.g. \cite{bahdanau2014neural,sutskever2014sequence}) and further seem poised to advance prediction in real-world domains \cite{miotto2016deep,gulshan2016development,ghassemi2017predicting}.
However, many practitioners still are reluctant to adopt deep models because their predictions are difficult to interpret. Without interpretability, humans are unable to incorporate their domain knowledge and effectively audit predictions.

In this work, we shall seek a specific form of interpretability known as \emph{human-simulability}. A human-simulable model is one in which a human user can ``take in input data together with the parameters of the model and in reasonable time step through every calculation required to produce a prediction'' \cite{lipton2016interpretability}.
For example, small decision trees with only a few nodes are easy for humans to simulate and thus understand.  Human-simulability is valuable in many domains.  In particular, despite advances in deep learning for clinical decision support (e.g. \cite{miotto2016deep,choi2016doctor,che2015deep}), the clinical community remains skeptical (and rightfully so) of machine learning systems \cite{chen2017machine}. The black box nature of neural networks prevents the checks-and-balances and quality control that we expect from healthcare providers.
Meanwhile, a simulable model would enable clinicians to audit predictions easily: they can manually inspect changes to outputs under perturbed inputs, check substeps against their expert knowledge, and reason about external factors influencing prediction like systemic bias in the data.
Similar needs for simulability exist in many decision-critical domains such as disaster response or recidivism prediction.

Despite the appeal and need for human-simulability, many popular models are not simulable.  Even simple deep models like multi-layer perceptrons with a few dozen units can have far too many parameters and connections for a human to easily step through (successive matrix multiplications quickly becomes difficult to think about). Richer families of neural networks such as those for sequences are essentially impossible for humans to simulate. However, with added non-linearities and many more free parameters, these rich families often allow for significantly more accurate predictions than a small decision tree.
Thus, the primary question we consider in this work is the following: \emph{Is it possible for a powerful model such as a deep network to be human-simulable, or at least frequently human-simulable?}

Simulability is a rather strict definition for interpretability as it requires full transparency in prediction. As such, current work on the interpretability of black-box models struggle to balance being both simulable and faithful to the model.
For instance, \citeauthor{craven1996extracting} \shortcite{craven1996extracting} train decision trees that mimic the predictions of a fixed, pre-trained neural network. Other post-hoc interpretations typically evaluate the sensitivity of predictions to local perturbations of inputs or the input gradient \cite{ribeiro2016should,selvaraju2017grad,adler2016auditing,lundberg2016unexpected,erhan2009visualizing}.  While the post-hoc interpretations come in many sophisticated forms--- others include \cite{singh2016programs}, who uses programs to explain a model's predictions as a post-hoc step, and \cite{lakkaraju2016interpretable}, who learn decision sets based on a learned model--- it is difficult to simplify the complex logic of an unregularized neural network to a simulable (simple) tree, set, or program. As a result, many of these methods only explain local behavior or a lower resolution (noisy) depiction of global logic. In general, the problem of distilling the decision function of a trained and unregularized neural network to a simple family of decision functions is somewhat ill-posed: unregularized neural networks have no incentive to be simulable or any other notion of human-interpretability.
Instead, they will learn complex decision boundaries fit to succeed at the target task. Trying to enforce interpretability post-hoc must understandably make strong assumptions that over-simplify the model's logic.

In contrast, we begin with the observation that since it is well-known that deep models often have multiple optima of similar predictive accuracy \cite{Goodfellow-et-al-2016} one might hope to directly find ``more interpretable" minima with equal predictive accuracy. In other words, if we consider interpretability from the very start i.e. add an ``interpretability term" in the objective function, it might be possible to train neural networks to be both performant and simulable. In general however, the field of \emph{optimizing} deep models for interpretability remains largely nascent. In this vein, \citeauthor{ross2017right} \shortcite{ross2017right} penalize input sensitivity to features marked as less relevant, while \citeauthor{lei2016rationalizing} \shortcite{lei2016rationalizing} train deep models that make predictions from text and simultaneously highlight contiguous subsets of words, called a ``rationale,'' to justify each prediction. Unfortunately, while both works optimize deep models to expose relevant features, these lists of features alone are not sufficient to \emph{simulate} the prediction. We draw a stark distinction between explanation and simulation: the former may describe interpretable features whereas the latter requires defining both features and a procedure for translating them into output. In the following, we introduce two contributions: we first discuss how to optimize deep models to expose prediction logic (not just features) using decison trees, and second, how to generalize this method to incorporate human prior knowledge.

\paragraph{Tree Regularization}
To optimize for interpretability, we must define an objective function that finds deep models that are both accurate and simulable.
To do this, we introduce the notion of \emph{tree-regularization.}  Specifically, we define a novel model-complexity penalty function that favors model optima whose decision boundaries can be well-approximated by small decision trees. In effect, this penalizes models that would require many calculations to simulate predictions. Similar to many popular regularizers such as L2 or L1, the tree regularizer is a function on the weights of the neural network. Several of our technical contributions surround making this regularizer differentiable such that it is compatible with stochastic gradient descent.
Experimentally, we first exemplify how this technique can be used to train simple multi-layer perceptrons to have tree-like decision boundaries. We then focus on time-series applications and show that gated recurrent unit (GRU) models trained with strong tree-regularization reach a high-accuracy-at-low-complexity sweet spot that is not possible with any strength of L1 or L2 regularization. Furthermore, we will show that the decision trees (produced during training) can be used as tools for human simulation -- they act as distillations of the deep model and can be give to domain experts. Choosing several real world applications, we demonstrate these features of our approach on a speech recognition task and two medical treatment prediction tasks for patients with sepsis in the intensive care unit (ICU) and for patients with human immunodeficiency virus (HIV). Throughout, we also show that standalone decision trees as a baseline are noticeably less accurate than our tree-regularized deep models.

\paragraph{Granularity of Explanation}
Thus far, we have implicitly assumed that there exists an optima for a deep model that is simulable while maintaining high performance. For many  domains, this may not be true -- we may rely on the complexity of a deep model where any strong regularization greatly increases error. In such cases, it may not be possible to have a model that is both accurate and well-approximated by a simple decision tree. To remedy this, we consider \textit{regional} explanations that constrain the model independently across a partitioning of the input space. Coincidentally, this form of explanation is consistent with those of humans, whose models are typically context-dependent \cite{miller2018explanation}. For example, physicians in the intensive care unit do not expect treatment rules to be the same across different categories of patients. Constraining each region to be interpretable allows the deep model more flexibility than a global constraint, while still revealing prediction logic that can generalize to nearby inputs (in contrast to works on local explanation---\cite{ribeiro2016should,selvaraju2016grad,ross2017right}---which cannot indicate whether the same logic revealed for an input $x$ can be used for nearby inputs $x'$, an ambiguity that can lead to mistaken assumptions and poor decisions). In other words, we assume that even the most complex decision boundaries can be decomposed into an ensemble of simpler regional boundaries, each of which can be well-approximated by a decision tree. Furthermore, in many domains like medicine, human experts have very good intuitions for how to partition the input space. For example, an intensivist may care for patients in the surgical unit differently than patients in (non)-surgical units. By generalizing tree regularization to support regions, we can incorporate prior knowledge from domain experts to train simulable models.

While a straightforward conceptual leap, optimizing for simulable explanations across many regions poses a difficult technical challenge, facing issues with differentiability, efficiency, and a delicate balance of constraints between regions of varying size and complexity. In the methods, we will describe a computationally tractable and reliable approach to do so. Specifically, we show how to jointly train a deep model that both has high accuracy and is regionally simulable, and introduce innovations for stability in optimization. We first present a few synthetic experiments to build intuition and then, revisiting the clinical domain, we demonstrate that \textit{regional tree regularization} achieves better performance while learning a much simpler decision function than any other regularizer.

\section{Related work}
\label{Related_Work}

\paragraph{Global Interpretability}
Given a \emph{trained} black box model, many approaches exist to explain what the model has learned.  Works such as \cite{mordvintsev2015inceptionism} expose the features a representation encodes but not the logic. \cite{amir2018highlights,kim2014bayesian} provide an informative set of examples that summarize the system. Model distillation compress a source network into a smaller target neural network \cite{frosst2017distilling}. However, even a small neural model may not be interpretable. Activation maximisation of neural networks \cite{montavon2018methods} tries to find input patterns that produce the maximum response for a quantity of interest. However, a set of input patterns is not necessarily adequate to simulate a model's predictions. Similarly, Layerwise-Relevance Propagation \cite{binder2016layer,bach2015pixel} produces a heatmap of relevant information for prediction based on the aggregating the weights of a neural network. Again, learning a heatmap of the important information for predicting outcomes does not always enable human simulability, since we cannot necessarily step through each calculation that produces a decision.

\paragraph{Local Interpretability}
In contrast, local approaches provide explanation for a specific input. \citeauthor{ribeiro2016should} \shortcite{ribeiro2016should} show that using the weights of a sparse linear model, one can explain the decisions of a black box  model in a small area near a fixed data point. This captures the intuition that even nonlinear functions are locally linear. Similarly, instead of a linear model, \citeauthor{singh2016programs} \shortcite{singh2016programs} and \citeauthor{koh2017understanding} \shortcite{koh2017understanding} output a simple program or an influence function, respectively. Other approaches have used input gradients (which can be thought of as infinitesimal perturbations) to characterize the local space \cite{maaten2008visualizing,selvaraju2016grad}.  However, the notion of a local region in these works is both very small and often implicit; it does not match with human notions of contexts \cite{miller2018explanation}: a user may have difficulty knowing when local explanations apply and how they generalize to nearby inputs.

\paragraph{Optimizing for Interpretability}
While there is little work on optimizing models for interpretability, there are some related threads. The first is \emph{model compression}, which trains smaller models that perform similarly to large, black-box models (e.g. \cite{buciluǎ2006model,hinton2015distilling,balan2015bayesian,han2015learning}).
Other efforts specifically train very sparse networks via L1 penalties \cite{zhang2016l1} or even \emph{binary} neural
networks \cite{tang2017train,rastegari2016xnor} with the goal of faster computation. Edge and node regularization is commonly used to improve prediction accuracy \cite{drucker1992improving,ochiai2017automatic}, and recently \citeauthor{hu2016harnessingLogic} \shortcite{hu2016harnessingLogic} improve prediction accuracy by training neural networks so that predictions match a small list of known domain-specific first-order logic rules.  Sometimes, these regularizations---which all smooth or simplify decision boundaries---\textit{can} have the effect of also improving interpretability. However, there is no guarantee that these regularizations will do so; we emphasize that specifically \emph{training} deep models to have easily-simulatable decision boundaries is (to our best knowledge) novel.

\section{Background and Models}
\label{Background}

We consider supervised learning tasks given datasets of $N$ labeled examples, $\mathcal{D} = \{(\mathbf{x}_n, \mathbf{y}_n)\}_{n=1}^N$, where each example (indexed by $n$) has an input feature vector $\mathbf{x}_n \in \mathcal{X}^P$ and a target output vector $\mathbf{y}_n \in \mathcal{Y}^Q$. $P$ and $Q$ are the dimensionalities. For example, we will sometimes write $\mathbf{x}_n = [x_{n}(1), ..., x_{n}(p)]$, using $(\cdot)$ to indicate indexing into the vector. We shall assume the targets $\mathbf{y}_n$ are binary, though it is simple to extend to other types. When modeling time-series, each example sequence $n$ contains $T_n$ timesteps indexed by $t$ which each have a feature vector
$\mathbf{x}_{nt}$ and an output $\mathbf{y}_{nt}$. Formally, we write: $\mathbf{x}_n = (\mathbf{x}_{n1} \ldots \mathbf{x}_{nT_n})$ and $\mathbf{y}_n = (\mathbf{y}_{n1} \ldots \mathbf{y}_{nT_n})$. Each value $\mathbf{y}_{nt}$ could be a prediction about the next timestep (e.g. the character at time $t+1$) or some other task-related annotation (e.g. if the patient became septic at time $t$).

We will primarily consider two kinds of deep models: multi-layer perceptrons and recurrent neural networks.  That said, our approach is compatible with any architecture.

\paragraph{Multi-Layer Perceptrons.}
A multi-layer perceptron (MLP) makes predictions $\mathbf{\hat{y}}_n$ of the target $\mathbf{y}_n$ via a function $f: \mathcal{X}^P \times \Theta \rightarrow \mathcal{Y}^Q$ such that $\hat{\mathbf{y}}_n = f(\mathbf{x}_n; \theta)$, where the vector $\theta \in \Theta$ represents all parameters of the network.  Given a data set $\mathcal{D}$, our goal is to learn the optimal parameters $\theta^*$ to minimize the objective
\begin{align}
\theta^* = \arg\min_{\theta \in \Theta} \sum_{n=1}^N \mathcal{L}( \mathbf{y}_n , \mathbf{\hat{y}}_n ) + \lambda \Psi(\theta)
\label{eqn:orig_loss}
\end{align}
For binary targets $\mathbf{y}_n$, the logistic loss (binary cross entropy) is an effective choice for $\mathcal{L}(\cdot)$. The regularization term $\Psi(\theta)$ can represent L1, L2 penalties (e.g. \cite{drucker1992improving,Goodfellow-et-al-2016,ochiai2017automatic}) or our new family of regularizers.

\begin{figure}
  \centering
  \begin{subfigure}[b]{0.3\linewidth}
      \centering
      \includegraphics[width=\linewidth]{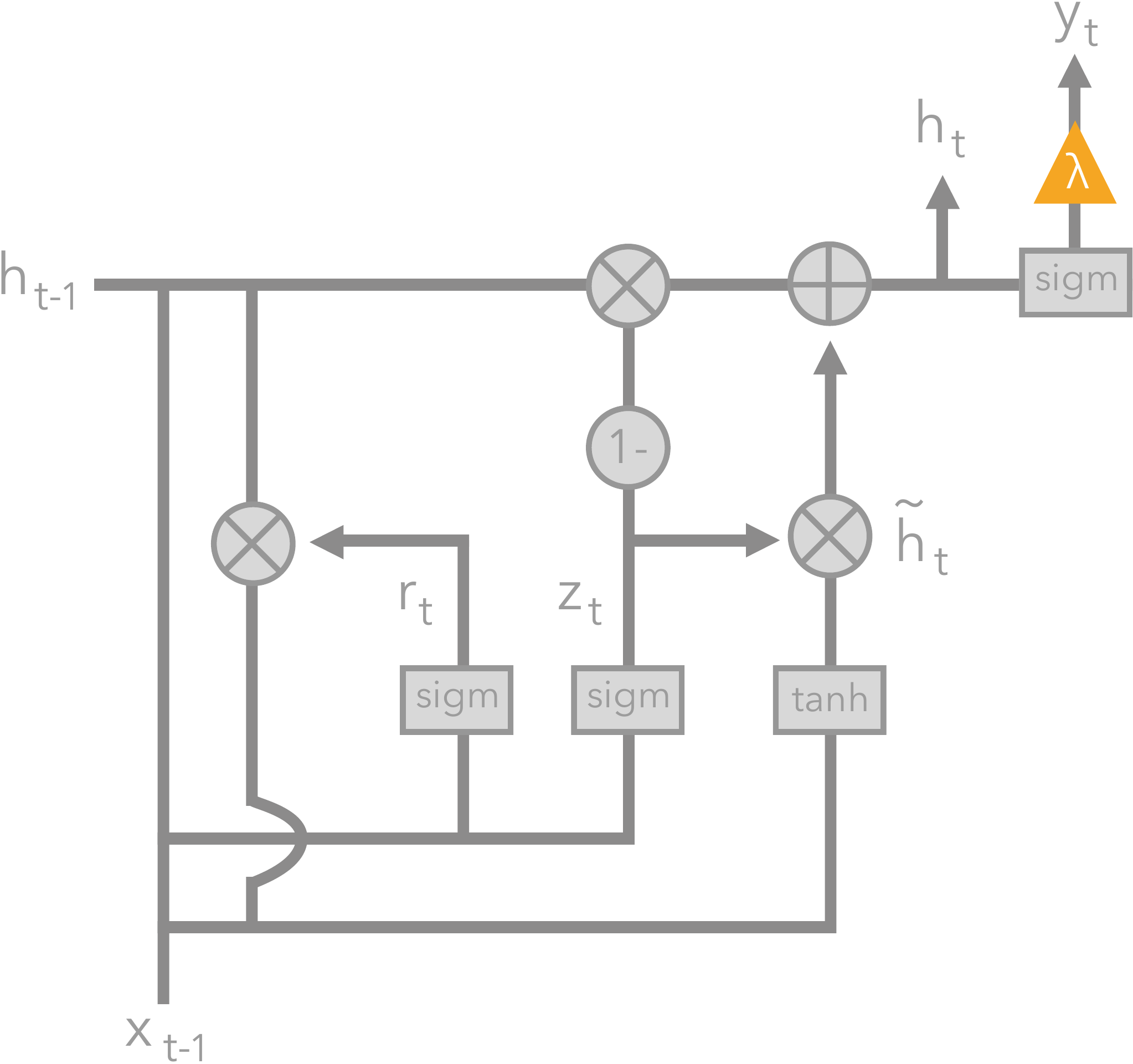}
      \caption{GRU}
  \end{subfigure}
  \qquad
  \begin{subfigure}[b]{0.55\linewidth}
      \centering
      \includegraphics[width=\linewidth]{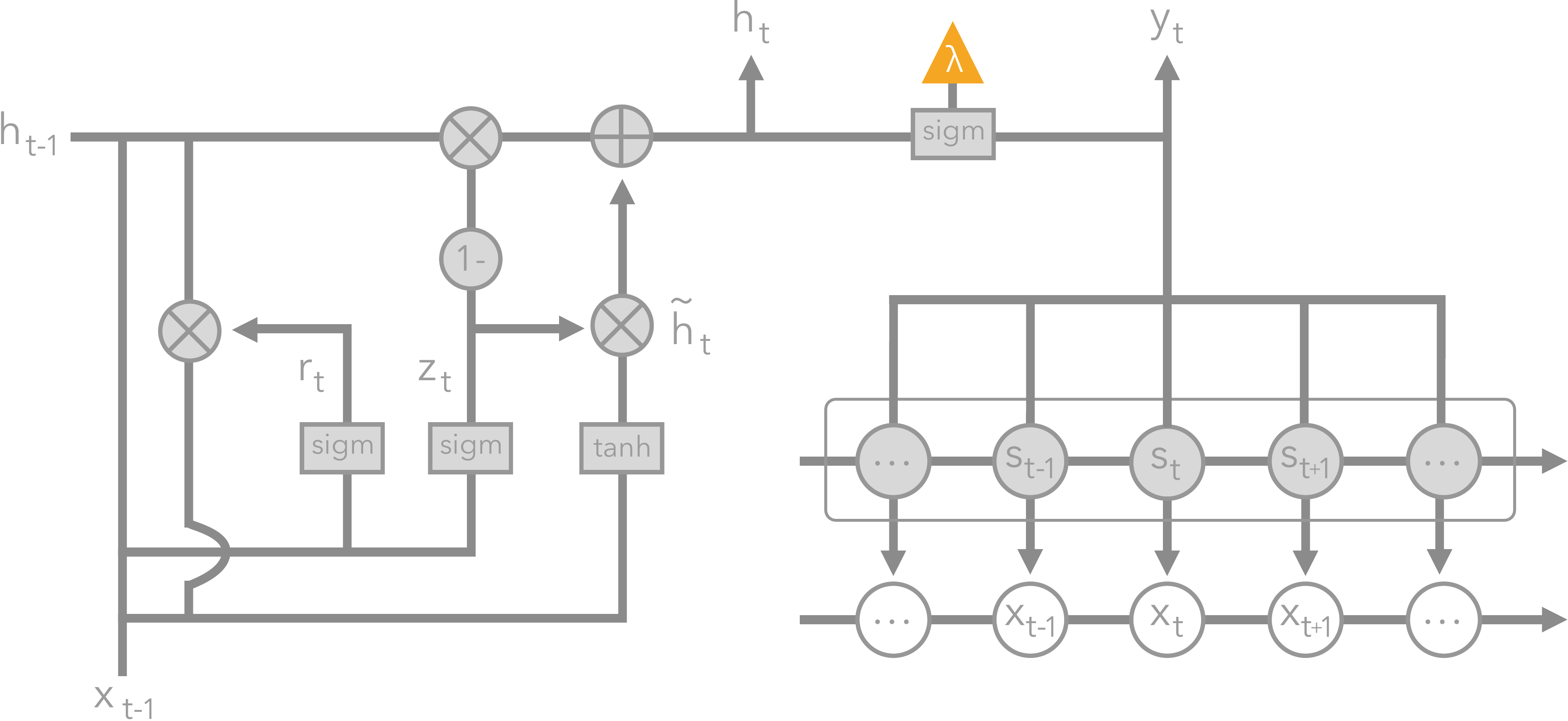}
      \caption{GRU-HMM}
  \end{subfigure}
  \caption{Architecture diagrams for (a) gated recurrent units (GRU) and (b) a GRU and hidden markov model (HMM) hybrid. The orange triangle indicates the output used in surrogate training for tree regularization.}
  \label{fig:what-we-regularize}
\end{figure}

\paragraph{Recurrent Neural Networks with Gated Recurrent Units.}
A recurrent neural network (RNN) takes as input an arbitrary length sequence $\mathbf{x}_n = (\mathbf{x}_{n1} \ldots \mathbf{x}_{nT_n})$ and produces a ``hidden state'' sequence $\mathbf{h}_n = (\mathbf{h}_{n1} \ldots \mathbf{h}_{nT_n})$ of the same length as the input. Each hidden state vector at timestep $t$ represents a location in a (possibly low-dimensional) ``state space'' with $K$ dimensions: $\mathbf{h}_{nt} \in \mathbb{R}^K$ ($K$ is often chosen as a hyperparameter). RNNs perform
sequential \emph{nonlinear} embedding of the form $\mathbf{h}_{nt} = f(\mathbf{x}_{nt}, \mathbf{h}_{nt-1}; \theta)$ in hope that the state space location $\mathbf{h}_{nt}$ is a useful summary statistic for making predictions of the target $\mathbf{y}_{nt}$ at timestep $t$. As written, $f: \mathcal{X}^P \times \mathbb{R}^K \times \Theta \rightarrow \mathbb{R}^K$ is called a \textit{transition} function parameterized by $\theta \in \Theta$. Many different variants of the transition function architecture have been proposed to solve the challenge of capturing long-term dependencies. In this paper, we use gated recurrent units (GRUs) \cite{cho2014gru}, which are simpler than other alternatives such as long short-term memory units (LSTMs) \cite{hochreiter1997long}. While GRUs are convenient, any differentiable RNN architecture is compatible with our new tree-regularization approach.

As review, we describe the evolution of a single
GRU sequence, dropping the sequence index $n$ for readability. The GRU transition function $f$ produces the state vector $\mathbf{h}_{t} =
(\mathbf{h}_{t1} \ldots \mathbf{h}_{tT})$ (let $T$ denote the number of timesteps) from a previous state $\mathbf{h}_{t-1}$ and an input vector $\mathbf{n}_t$, via the following feed-forward architecture:
\begin{align}
\textup{output state}: \mathbf{h}_{t} &= (1 - \mathbf{z}_{t}) \mathbf{h}_{t-1} + \mathbf{z}_{t,k}\mathbf{\tilde{h}}_{t}
\\
\textup{candidate state}:
\mathbf{\tilde{h}}_{t} &= \textup{tanh}( \mathbf{V}_h \mathbf{x}_t + \mathbf{U}_h
    (\mathbf{r}_{t} \odot \mathbf{h}_{t-1}) )
\\
\textup{update gate}:
\mathbf{z}_{t} &= \sigma( \mathbf{V}_z \mathbf{x}_t + \mathbf{U}_z \mathbf{h}_{t-1} )
\\
\textup{reset gate}:
\mathbf{r}_{t} &= \sigma(\mathbf{V}_r \mathbf{x}_t + \mathbf{U}_r \mathbf{h}_{t-1})
\end{align}
The internal network nodes include candidate state gates $\mathbf{\tilde{h}}$, update gates $\mathbf{z}$ and reset gates $\mathbf{r}$ which have the same cardinality as the state vector $\mathbf{h}$. Reset gates allow the network to forget past state vectors when set near zero via the logistic sigmoid nonlinearity $\sigma(\cdot)$, which critically adds a multiplicative expressivity to this model class. Update gates allow the network to either pass along the previous state vector unchanged or use the new candidate state vector instead. This architecture is diagrammed in Figure~\ref{fig:what-we-regularize}.

The predicted probability of the binary target $\mathbf{y}_t$ for timestep $t$ is a sigmoid transformation of the state at time $t$, $\mathbf{\hat{y}}_t = \sigma(\mathbf{w}^T \mathbf{h}_t)$.
Here, weight vector $w \in \mathbb{R}^K$ represents the parameters of this individual output layer. We denote the parameters for the entire GRU-RNN model as $\theta = (\mathbf{w}, \mathbf{U}_h, \mathbf{U}_z, \mathbf{U}_r, \mathbf{V}_h, \mathbf{V}_z, \mathbf{V}_r) \in \Theta$, concatenating all component parameters. We can train GRU-RNN timeseries models (hereafter often just called GRUs) via the following loss minimization objective, sharing many similarities to the MLP's loss (Eqn.~\ref{eqn:orig_loss}):
\begin{align}
\theta^* = \arg\min_{\theta \in \Theta} \sum_{n=1}^N \sum_{t=1}^{T_n} \mathcal{L}( \mathbf{y}_{nt}, \mathbf{\hat{y}}_{nt}) + \lambda \Psi(\theta)
\label{eqn:gru_loss}
\end{align}
where again $\Psi(\theta)$ defines a regularization cost, and $\theta^*$ represents the optimal parameters.

\paragraph{Hidden Markov Models with Stochastic Gradient Descent.} Besides recurrent neural networks, hidden markov models (or HMMs) are another class of sequence models that are commonly used to describe stochastic processes. Often, (as with RNNs) we are given a sequence of $T_n$ observed variables $\mathbf{x}_n = (\mathbf{x}_{n1} \ldots \mathbf{x}_{nT_n})$, and wish to derive a sequence of $T_n$ latent (or hidden) variables $\mathbf{s}_n = (\mathbf{s}_{n1} \ldots \mathbf{s}_{nT_n})$. We assume each latent variable, $\mathbf{s}_{nt}$ can take one of $K$ discrete states. In practice, these latent variables can be interpreted as an unsupervised clustering over the observed sequence. For our purposes, one can view the HMM as a stochastic RNN (added noise), making it a probabilistic generative model. To be tractable, the HMM makes a set of simplifying assumptions. The free parameters of an HMM define a \textit{prior}, $p(\mathbf{s}_{n0})$, the probability distribution over $K$ states for timestep 0; a \textit{transition matrix}, $p(\mathbf{s}_{nt}|\mathbf{s}_{n,t-1})$ which specifies a probability distribution over states for timestep $t$ given the state at timestep $t-1$; and an \textit{emission matrix}, $p(\mathbf{x}_{nt}|\mathbf{s}_{nt})$ which specifies a probability distribution over (possibly continuous) observations at timestep $t$ given only the latent at timestep $t$. Critically, this setup makes the Markov assumption -- all information required to make a decision at timestep $t$ is present at timestep $t-1$.

In our setting, we also have a sequence of known outputs, $\mathbf{y}_n = (\mathbf{y}_{n1} \ldots \mathbf{y}_{nT_n})$. In some sense, we are not interested not in the latent states themselves but using them to classify an observation into output. If we decide upfront to specify a simple classifier on top of the latent variables (such as logistic regression), then we explicitly write the joint distribution over latents, observations, and outputs as:
\begin{equation}
  p(\mathbf{x}_n, \mathbf{y}_n, \mathbf{s}_n) = p(\mathbf{s}_{n0};\phi)\prod_{t=1}^T p(\mathbf{s}_{nt}|\mathbf{s}_{n,t-1};\phi)p(\mathbf{x}_{nt}|\mathbf{s}_{nt};\phi)p(\mathbf{y}_{nt}|\sigma(\sum_{k} w_k f(\mathbf{s}_{nt})))
\end{equation}
where $\phi$ are the parameters specifying the prior, transition, and emission probabilities; $\{ w_k \}_{k=1}^K$ are the parameters used in logistic regression; $f(\mathbf{s}_{nt}) = p(\mathbf{s}_{nt}|\mathbf{s}_{n,t-1}, \mathbf{x}_{nt};\phi)$, the posterior distribution over states at timestep $t$; $\sigma$ represents a Sigmoid function. Therefore, we can train the HMM with stochastic gradient descent using the objective:

\begin{equation}
  \theta^* = \arg \max_{\theta \in \Theta} p(\mathbf{x}_n, \mathbf{y}_n, \mathbf{s}_n)
\end{equation}

where $\theta = \{\phi, w_1, ..., w_K \}$ contain all trainable parameters from a high-dimensional space of parameters $\Theta$. In other words, because we only desire maximum-a-posteriori (MAP) inference, we never need to sample from any of the distributions and therefore can differentiate this objective with standard techniques. Note that this is quite similar to the forward pass in the forward-backward algorithm.

\paragraph{Modeling the Residuals of a Hidden Markov Models}

One strength of the HMM is that it is a fairly interpretable model. Often, the discrete latent states have contextual meaning such that we can analyze the predictions of HMM as conditioned completely on its state. However, for complex domains, discrete states (even for large $K$) might not be able to fully capture the true decision function, resulting in high prediction error. One option is to add a recurrent neural network, which are known to be high performing but un-interpretable, to model the residual errors when predicting the target outputs using the HMM belief (latent) states. If we can properly penalize the complexity of the deep model, then high quality predictions do not come at the price of a less interpretable model. In practice, the GRU and HMM can be trained jointly where the parameters of each model are kept independent. We call this model a \textit{GRU-HMM} and use it in several experiments. Figure~\ref{fig:what-we-regularize}(b) recap the model architecture.

\section{(Decision) Tree-Regularization}

As presented in Eqns.~\ref{eqn:orig_loss} and \ref{eqn:gru_loss}, the regularizer $\Psi(\theta)$ is arbitrary. Common choices include $L_2$ norms to manage the sizes of $\theta$ and $L_1$ norms to manage the sparsity of $\theta$. We now come to our core contribution: we replace $\Psi(\theta)$ with a novel \emph{tree-regularizer}, denoted $\Omega(\theta)$, that encourages the model $\theta$ to be \emph{simulable}. Specifically, we shall encourage our deep models to be well-approximated by (small) decision trees. For clarity, we refer to the deep neural network that we are trying to regularize as the \textit{target neural model} or target network.

To do so, we first fit a binary decision tree which \textit{accurately} reproduces the target network's thresholded binary predictions $\mathbf{\hat{y}}_{n}$ given input $\mathbf{x}_n$. The accuracy parameter is always kept fixed, so that the tree is forced to model the network well. Next, we penalize the network based on the complexity of learnt tree: a simple decision function can be explained with only a few branches whereas a complex function may need exceedingly large trees. With this in mind, we quantify complexity as the \emph{average decision path length} (shorthand APL) ---the average number of decision nodes that must be touched to make a prediction for an input $\mathbf{x}_n$ (i.e. the number of nodes from root to leaf). We compute the \emph{average} with respect to some designated reference dataset of example inputs $\mathcal{D} = \{\mathbf{x}_n\}$ from the training set.  Thus, our regularizer is
\begin{equation}
  \Omega(\theta) \triangleq \text{APL}(\{\mathbf{x}_n\}_{n=1}^N, f(\cdot; \theta), h)
  \label{eqn:gapl}
\end{equation}
where the APL function is detailed in Algorithm~\ref{alg:true_tree_regularization}; $f(\cdot; \theta)$ represents the neural model; $h$ is a hyperparameter for training decision trees that controls the minimum number of training examples to define a leaf node. This definition of APL generalizes when the input data represents a timeseries. Algorithm~\ref{alg:true_tree_regularization} requires two subroutines, \textsc{TrainTree} and \textsc{PathLength}. Firstly, \textsc{TrainTree} trains a binary decision tree to accurately reproduce the provided labeled examples $\{\mathbf{x}_n, \mathbf{\hat{y}}_n \}$ (recall $\mathbf{\hat{y}}_n = f(\mathbf{x}_n; \theta)$). For this we use the \texttt{DecisionTree} module distributed in Python's scikit-learn \cite{scikit-learn}, which fits a tree by maximizing information gain with Gini impurity. Generally, the runtime cost of this module scales superlinearly with the number of examples $N$ and linearly with the number of features $F$ for a total complexity of $O(FN\log N)$. In practice, we found that with $N = 1000$, $F=10$, fitting a decision tree takes 15.3 microseconds. These trees can give probabilistic predictions at each leaf. Next, \textsc{PathLength} counts how many nodes are needed to make a specific input to an output node in the provided decision tree (this is done programmatically by storing traversals).

We consider average path length a good proxy for simulability because human simulation requires stepping through every calculation required to make a prediction. Average path length (or APL) exactly counts the number of true-or-false boolean calculations needed to make an average prediction, assuming the model is a binary decision tree.  In contrast, a metric such as the total number of nodes might penalize more accurate trees that have short paths for most examples but need more involved logic for few outliers. While a sensible choice, a few technical innovations are required to efficiently optimize the APL loss.

\begin{algorithm}[!t]
\caption{Average-Path-Length (APL) Cost Function}
\begin{algorithmic}[1]
\Require{
\Statex $f(\cdot; \theta)$ : binary prediction function, with parameters $\theta$
\Statex $\mathcal{D} = \{ \mathbf{x}_n \}_{n=1}^N$ : reference dataset with $N$ examples
\Statex $h$ : minimum number of samples required to be a leaf node; a higher $h$ regularizes the tree, resulting in a smaller tree
}
\Function{\textup{APL}}{$\{\mathbf{x}_n\}, f(\cdot; \theta), h$}
\State $\mbox{tree} \gets \textsc{TrainTree}( \{ \mathbf{x}_n, f(\mathbf{x}_n, \theta) \}_{n=1}^N)$
\State \Return $\frac{1}{N} \sum_{n} \textsc{PathLength}(\mbox{tree}, \mathbf{x}_n)$
\EndFunction
\end{algorithmic}
\label{alg:true_tree_regularization}
\end{algorithm}

\paragraph{Making Tree Regularization Differentiable}
Training decision trees is not differentiable, and thus the tree regularization loss $\Omega(\theta)$ from Equation~\ref{eqn:gapl} is not differentiable with respect to the network parameters $\theta$ (unlike standard regularizers such as $L_1$ or $L_2$). While one could resort to derivative-free optimization techniques \cite{audet2016blackbox} e.g. search algorithms, gradient descent has been an extremely fast and robust way of training neural networks \cite{Goodfellow-et-al-2016}.

\begin{figure}[h!]
  \centering
  \begin{subfigure}[b]{0.47\linewidth}
      \centering
      \includegraphics[width=\linewidth]{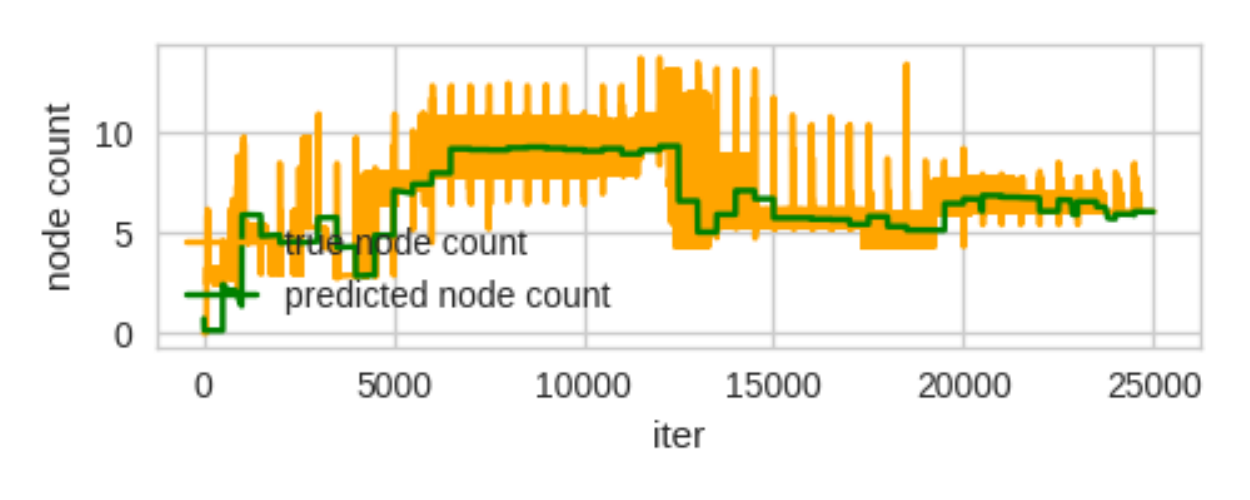}
      \caption{Surrogate APL vs True APL}
  \end{subfigure}
  \begin{subfigure}[b]{0.45\linewidth}
      \centering
      \includegraphics[width=\linewidth]{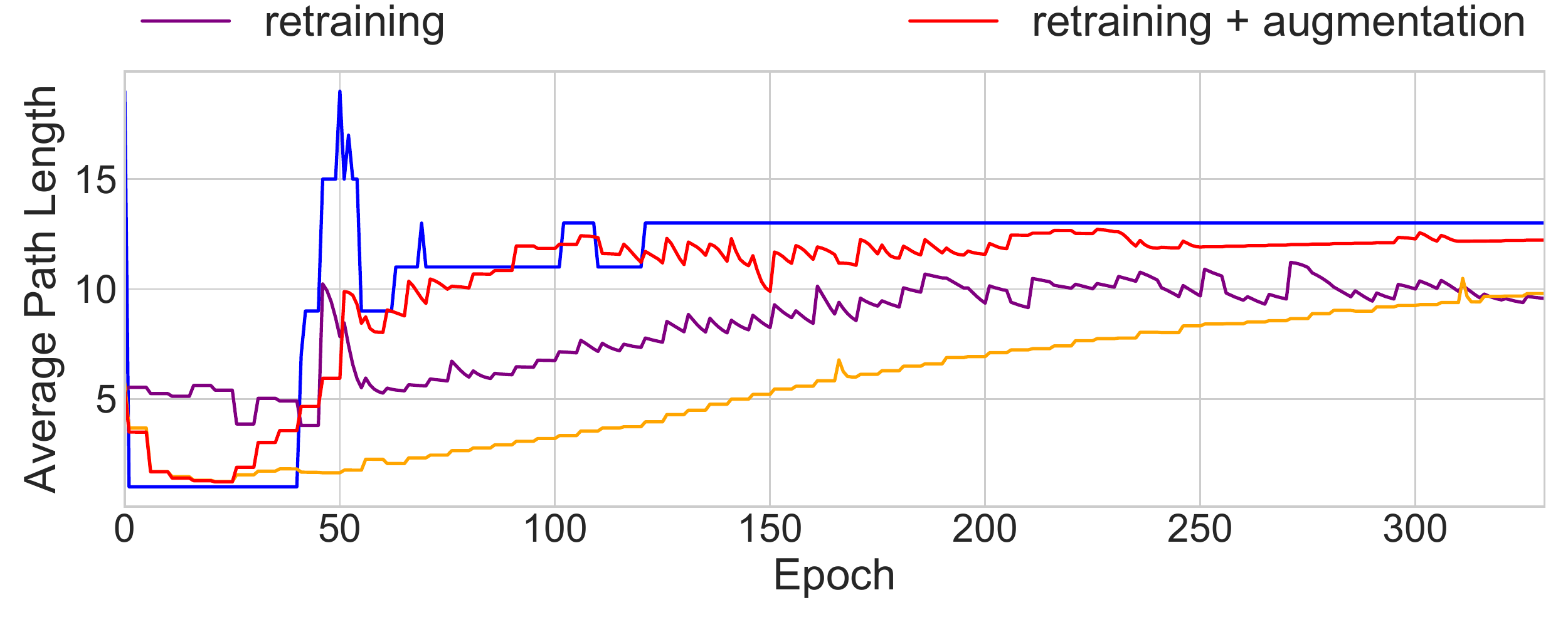}
      \caption{Effect of Restarts and Augmentation}
  \end{subfigure}
  \caption{\emph{(a)} True average path lengths (yellow) and surrogate estimates $\hat{\Omega}$ (green) across many iterations of network parameter
  training iterations (on 2D Parabola). \emph{(b)} Compares the effects of parameter augmentation and random restarts (retraining): The blue line shows the true APL of the decision tree at each epoch. All other lines show predicted APL using the surrogate MLP. By augmenting and restarting, we significantly improve the ability of the surrogate model to track the changes in the ground truth.}
  \label{fig:tricks}
\end{figure}

A key technical contribution of our work is introducing and training a \emph{surrogate} regularization function
$\hat{\Omega}(\theta): \mbox{supp}(\theta) \rightarrow \mathbb{R}^+$ to map each parameter vector $\theta \in \Theta$ of the target neural model to an \emph{estimate} of the APL. Our approximate function $\hat{\Omega}$ is implemented as a standalone multi-layer perceptron network and is critically \emph{differentiable}. Let vector $\xi \in \Xi$ denote the trainable parameters of this chosen MLP surrogate. We can train $\hat{\Omega}$ to be a good estimator by minimizing a squared error
loss function:
\begin{align}
\min_{\xi \in \Xi} \textstyle \sum_{j=1}^J (\Omega( \theta_j ) - \hat{\Omega}( \theta_j, \xi ) )^2 + \epsilon || \xi ||_{2}^2
\label{eqn:surrogate_loss}
\end{align}
where each $\theta_j$ is an instance of the \emph{entire} set of parameters for the target neural model, $\epsilon > 0$ is a regularization strength, and we assume we have a dataset of $J$ known parameter vectors and their associated true APLs: $\mathcal{D}^{\theta} = \{\theta_j, \Omega(\theta_j) \}_{j=1}^J$. This dataset can be assembled using the candidate parameter vectors obtained every gradient step while training our target neural model $f(\cdot, \theta)$. Importantly, one can train the surrogate function $\hat{\Omega}$ in parallel with our network. In Figure~\ref{fig:tricks}(a), we show evidence that our surrogate predictor $\hat{\Omega}(\cdot)$ tracks the true average path length as we train the target predictor $f(\cdot, \theta)$.

\paragraph{Training the Surrogate Loss} In this section, we describe a few more considerations to improve surrogate quality. Firstly, even moderately-sized neural models can have parameter vectors $\theta$ with thousands of dimensions.  Our labeled dataset for surrogate training -- $\{ \theta_j, \Omega(\theta_j) \}_{j=1}^J$---will only have one $\theta_j$ example from each target network training iteration. Even with small batch sizes (more gradient steps), this dataset is too small. Thus, in early iterations, we will have only few examples from which to learn a good surrogate function $\hat{\Omega}(\theta)$.  We resolve this challenge via \emph{augmenting} our training set with additional examples: We randomly sample weight vectors $\theta$ and calculate the true APL $\Omega(\theta)$, and we also perform several random restarts (initializing parameters with different random seeds) on the unregularized target network and use those weights in our training set.

A second challenge arises later in training: as the model parameters $\theta$ shift away from their initial values, parameters from earlier in optimization may not be as relevant in characterizing the current decision function of the target neural model. In practice, this is a function of the learning rate: a high step size will quickly render recent parameters ineffective for training a surrogate. To address this, for each epoch, we use examples only from the past $E$ iterations, where in practice, $E$ is empirically chosen.  Consequently, using examples from a fixed window of iterations also speeds up training. Figure~\ref{fig:tricks}(b) shows a comparison of the importance of these heuristics for efficient and accurate training---empirically, data augmentation for stabilizing surrogate training allows us to scale to neural networks with 100s of nodes. MLPs and GRUs of this size are already sufficient for many real problems, such as those we encounter in healthcare domains.

\begin{figure}[h!]
  \centering
  \begin{subfigure}[b]{0.3\linewidth}
      \centering
      \includegraphics[width=0.60\linewidth]{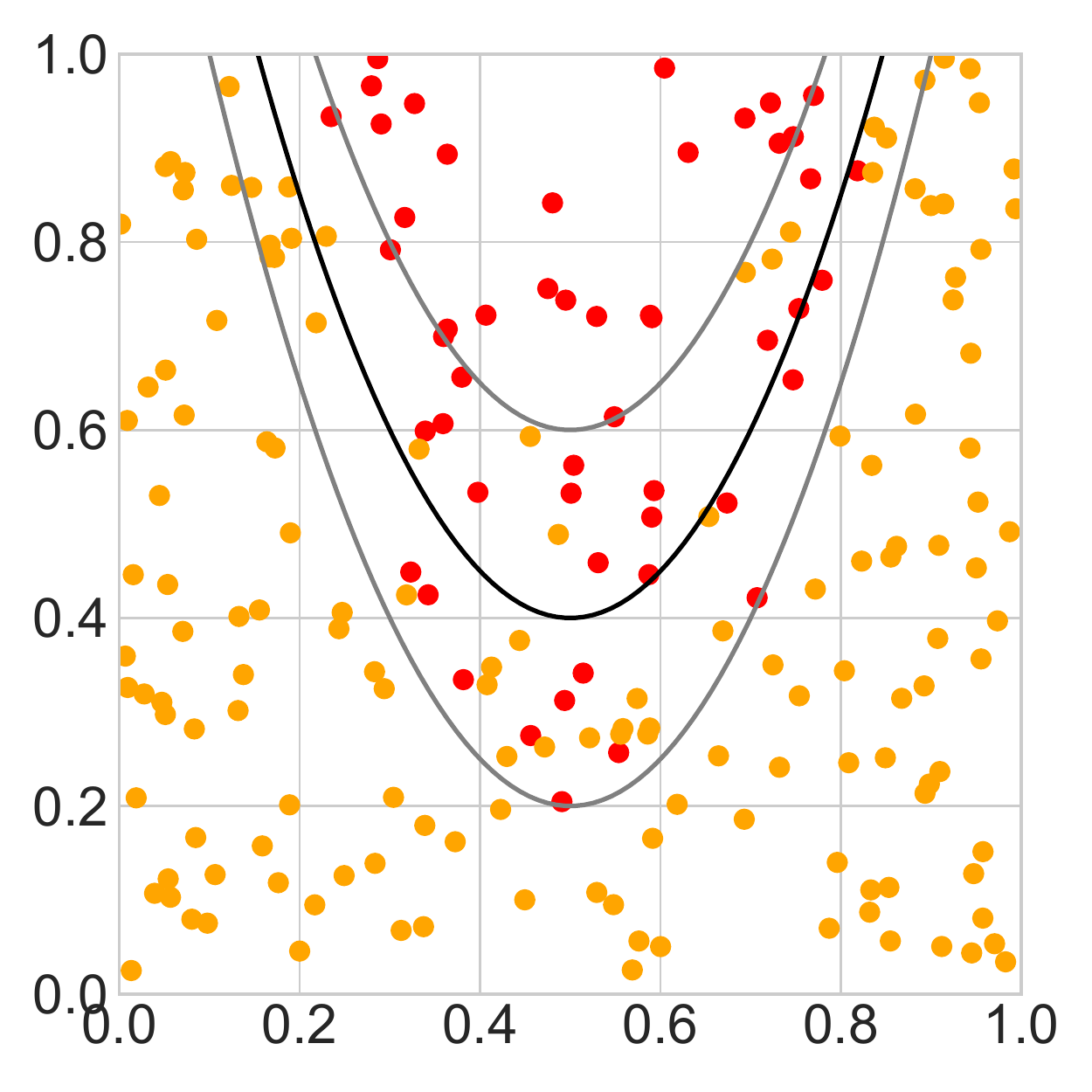}
      \caption{2D Parabola}
  \end{subfigure}
  \begin{subfigure}[b]{0.5\linewidth}
      \includegraphics[width=\linewidth]{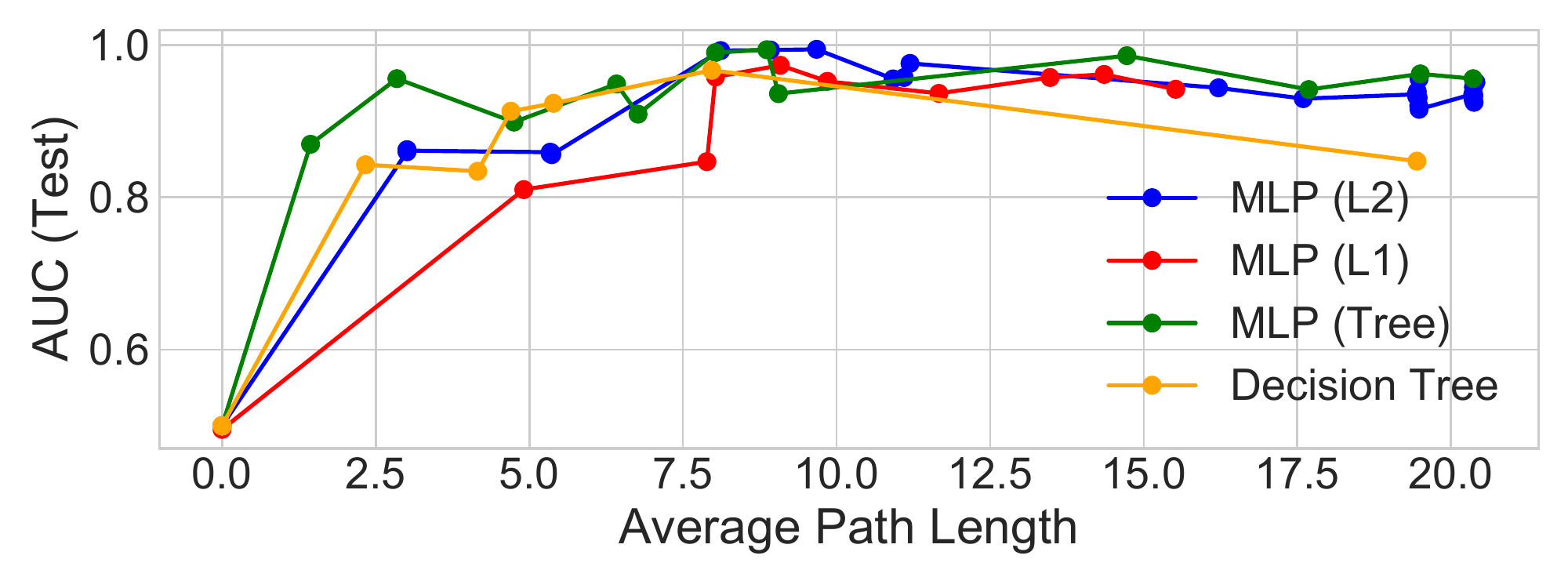}
      \caption{Prediction vs Complexity for many $\lambda$}
  \end{subfigure}
  \caption{\emph{(a)} 2D parabola dataset. The black line shows the true decision boundary; the gray lines define areas where noise is added. \emph{(b)} A comparision of APL versus AUC for many regularizers. In the small average path length regime (0-5), tree-regularization produces models with higher AUC than L$_1$ or L$_2$.}
  \label{fig:parabola}
\end{figure}

\section{Demonstration: A Tree-Regularized MLP and RNN}

We start by exploring two simple domains intended to build intuition for the tree regularization method. We first test the regularizer on MLPs in a two-dimensional classification task followed by a second prediction task with sequential data.

\paragraph{Tree-Regularized MLP: Noisy Parabola}

We first show a binary classification task as demonstration.  We call this task the \textit{2D Parabola problem}, because as Figure~\ref{fig:parabola}(a) shows, the training data consists of 2D input points whose two-class decision boundary is roughly shaped like a parabola.  The true decision function is defined by $y = 5*(x-0.5)^{2} + 0.4$.
\begin{wrapfigure}{r}{0.5\textwidth}
\centering
    \begin{subfigure}[b]{\linewidth}
        \includegraphics[width=\linewidth]{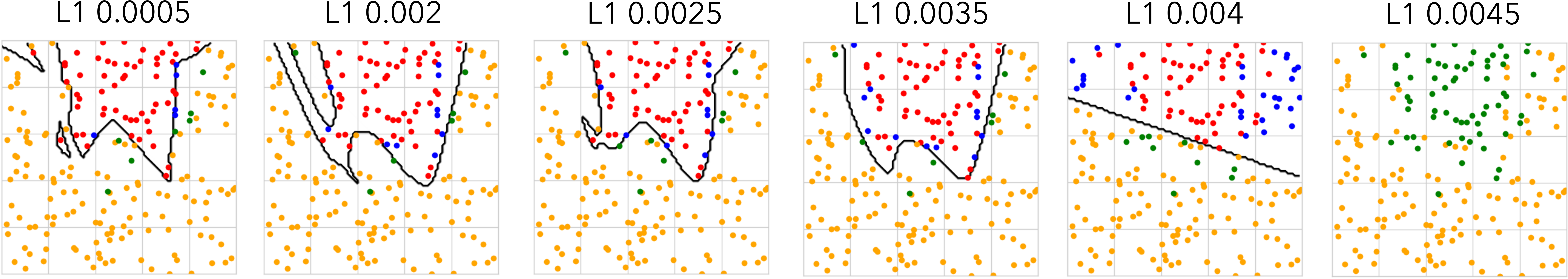}
        \caption{Decision Boundaries with L1 regularization\\}
        \label{fig:decision_function:l1}
    \end{subfigure}
    \begin{subfigure}[b]{\linewidth}
        \includegraphics[width=\linewidth]{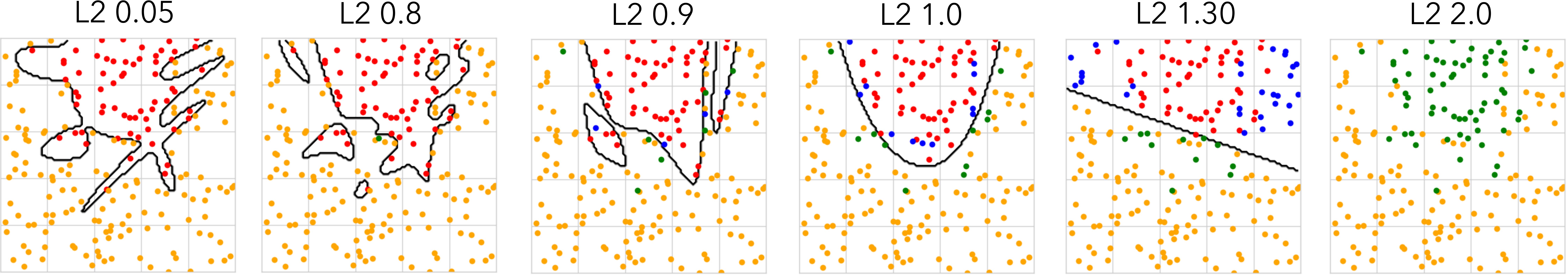}
        \caption{Decision Boundaries with L2 regularization\\}
        \label{fig:decision_function:l2}
    \end{subfigure}
    \begin{subfigure}[b]{\linewidth}
        \includegraphics[width=\linewidth]{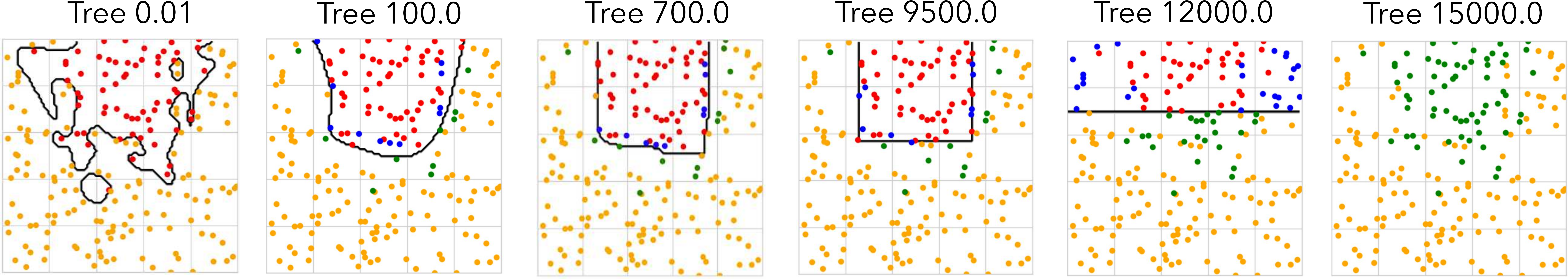}
        \caption{Decision Boundaries Tree regularization}
        \label{fig:decision_function:tree}
    \end{subfigure}
    \caption{Decision boundaries (black lines) have qualitatively different shapes
    for different regularization schemes, as regularization strength $\lambda$ increases. We color each prediction as true positive (red), true negative (yellow), false negative (green), and false positive (blue). The L$_1$ boundary appears more sharp, whereas L$_2$ is more round, and tree reg. is axis-aligned.}
    \label{fig:parabola2}
\end{wrapfigure}
We sampled 500 input points $\mathbf{x}_n$ uniformly within the unit square $[0,1] \times [0,1]$ and labeled those above the decision function as positive.  To make it easy for models to overfit to more complex decision boundaries, we flipped 10\% of the points in a region near the boundary.  A random 30\% were held out for testing. For the classifier, we train a 3-layer MLP with 100 first layer nodes, 100 second layer nodes, and 10 third layer nodes. This MLP is intentionally overly expressive to encourage overfitting and expose the impact of different forms of regularization: our proposed tree regularization $\Psi(\theta) = \hat{\Omega}(\theta)$, an L$_2$ penalty on the weights $\Psi(\theta) = ||\theta||_2$, and an L$_1$ penalty on the weights $\Psi(\theta) = ||\theta||_1$. For each regularization function, we train models at many different regularization strengths $\lambda$ chosen to explore the full range of decision boundary complexities possible under each technique. For tree regularization, we model a  surrogate $\hat{\Omega}(\theta)$ with a 1-hidden layer MLP with 25 units. The surrogate is intentionally chosen to be small with few parameters. In practice, we bias towards simpler surrogate networks to ensure faster training -- additionally, too complex of a surrogate would no longer preserve intepretability. The objective in Equation~\ref{eqn:orig_loss} was optimized via Adam gradient descent \cite{kingma2014adam} using a batch size of 100 and a learning rate of 1e-3 for 250 epochs. These hyperparameters were set via cross validation using grid search.

To evaluate model simulability, we use APL. Since Algorithm~\ref{alg:true_tree_regularization} can compute the APL for \textit{any} fixed deep model given its parameters, we use it to measure decision boundary complexity under any regularization, including L$_1$ or L$_2$. Figure~\ref{fig:parabola2}(b) shows each trained model as a single point in a 2D fitness space: the x-axis measures model complexity with APL, and the y-axis measures AUC (area under the ROC curve) prediction performance. These results show that simple L$_1$ or L$_2$ regularization does \emph{not} produce models with both small node count and good
predictions at \emph{any} value of the regularization strength $\lambda$. As expected, large $\lambda$ values for L$_1$ and L$_2$ only produce far-too-simple linear decision boundaries with poor accuracies.  In contrast, our proposed tree regularization directly optimizes the MLP to have simple tree-like boundaries at high $\lambda$ values which can still yield good predictions. The lower panes of Figure~\ref{fig:parabola2} shows these boundaries. Our tree regularization is uniquely able to create \textit{axis-aligned} functions, because decision trees by definition parameterize functions with axis-aligned splits. Critically, these axis-aligned functions require very few nodes but are more effective than L$_1$ and L$_2$ counterparts.

\paragraph{Tree-Regularized GRU: Signal-and-noise HMM}

\begin{figure}[t]
    \centering
    \begin{subfigure}[b]{0.18\linewidth}
        \includegraphics[width=0.76\linewidth]{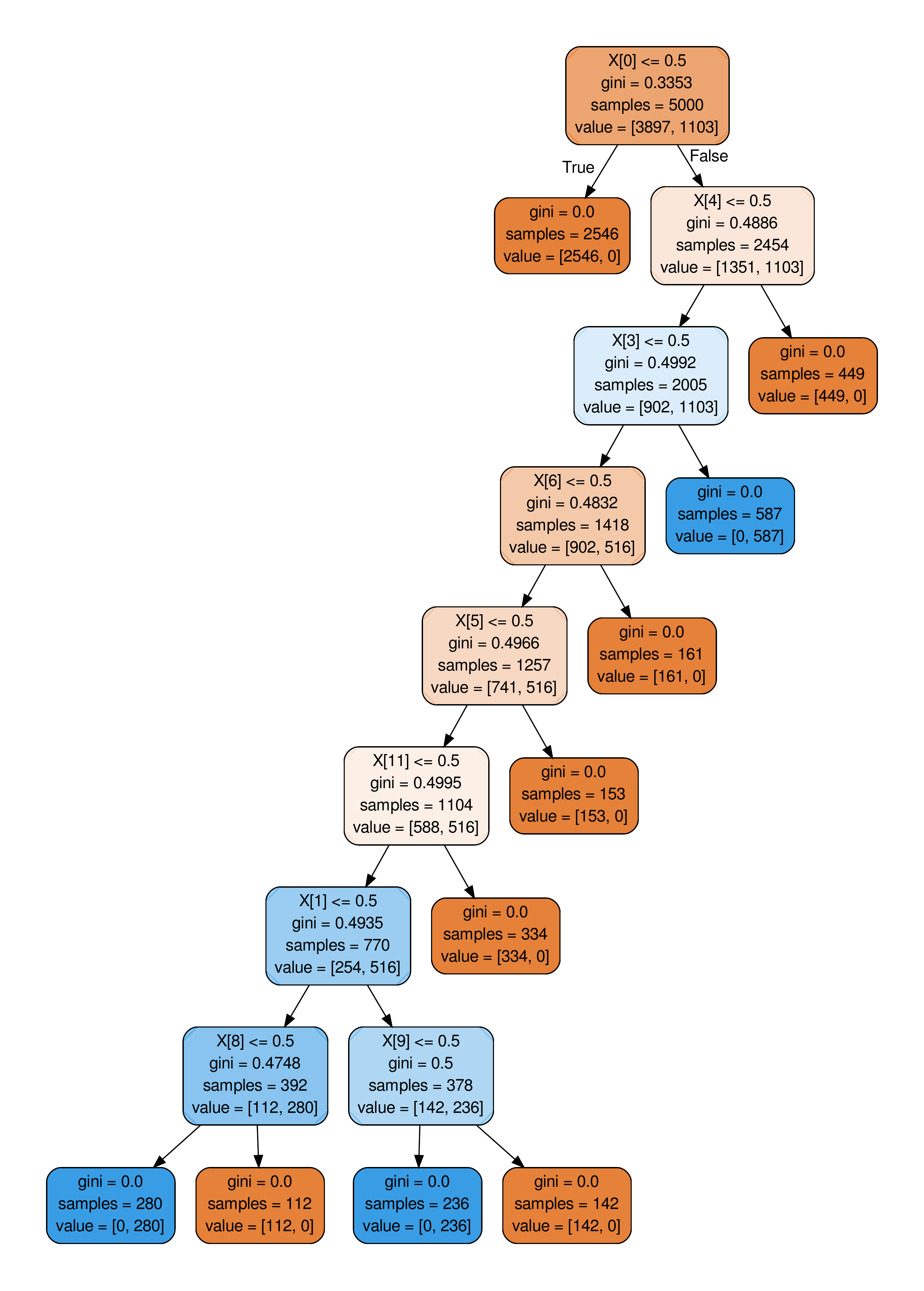}
        \label{fig:hmm2:tree:1}
        \caption{GRU $\lambda = 1$}
    \end{subfigure}
    \begin{subfigure}[b]{0.23\linewidth}
        \includegraphics[width=0.8\linewidth]{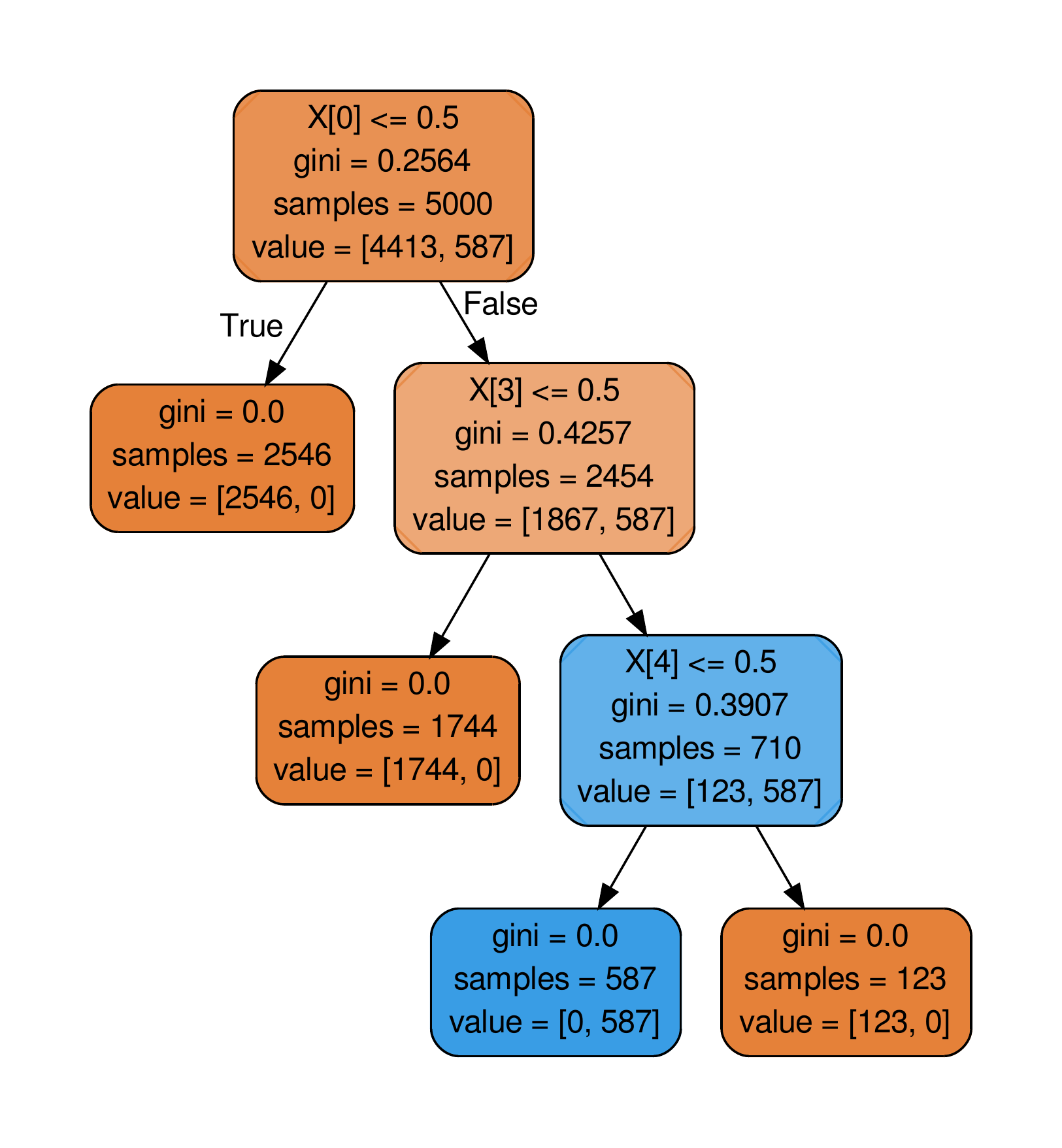}
        \label{fig:hmm2:tree:1000}
        \caption{GRU $\lambda = 1\,000$}
    \end{subfigure}
    \begin{subfigure}[b]{0.28\linewidth}
        \includegraphics[width=\linewidth]{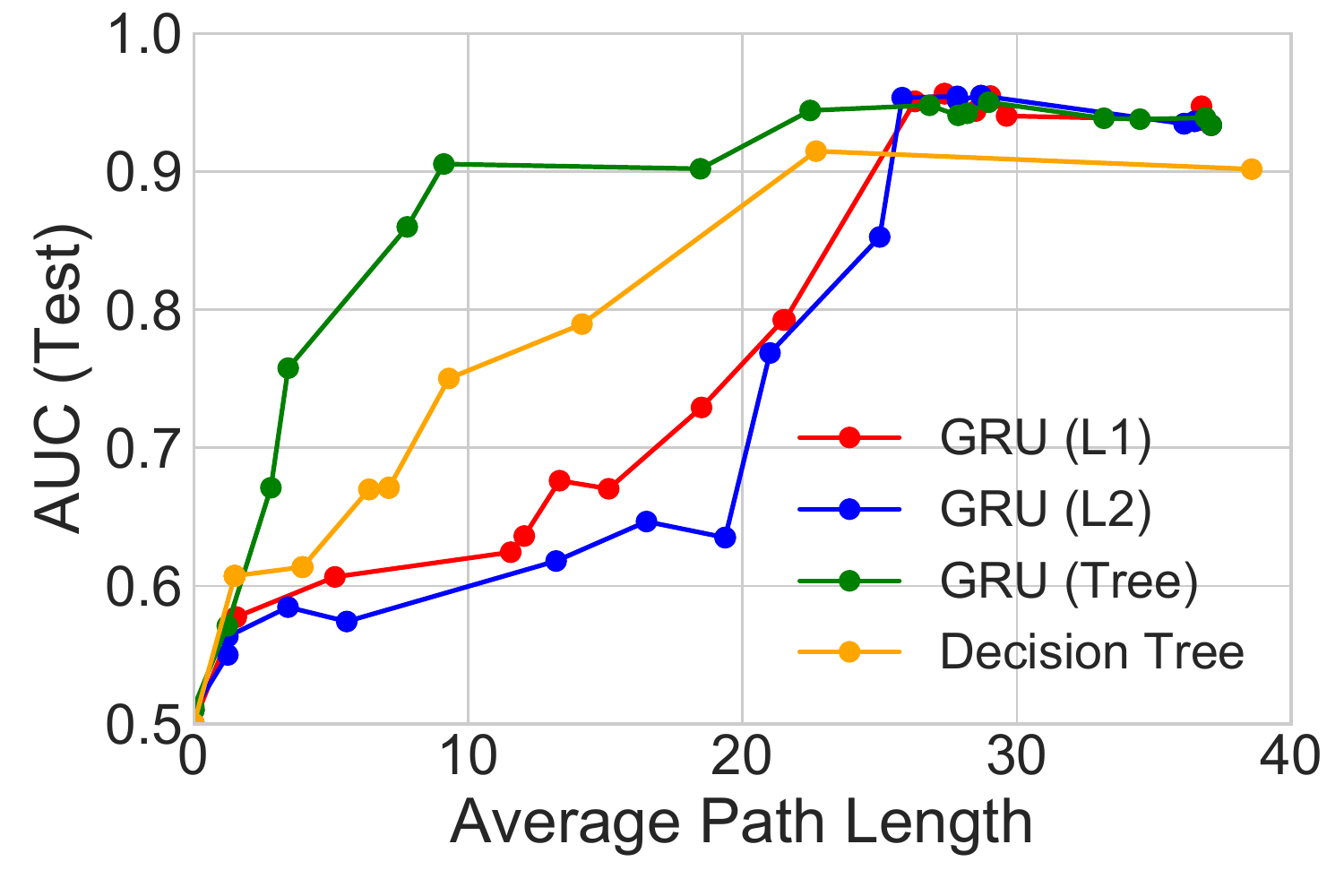}
        \caption{GRU}
        \label{fig:toy:gru:plot}
    \end{subfigure}
    \begin{subfigure}[b]{0.28\linewidth}
        \includegraphics[width=\linewidth]{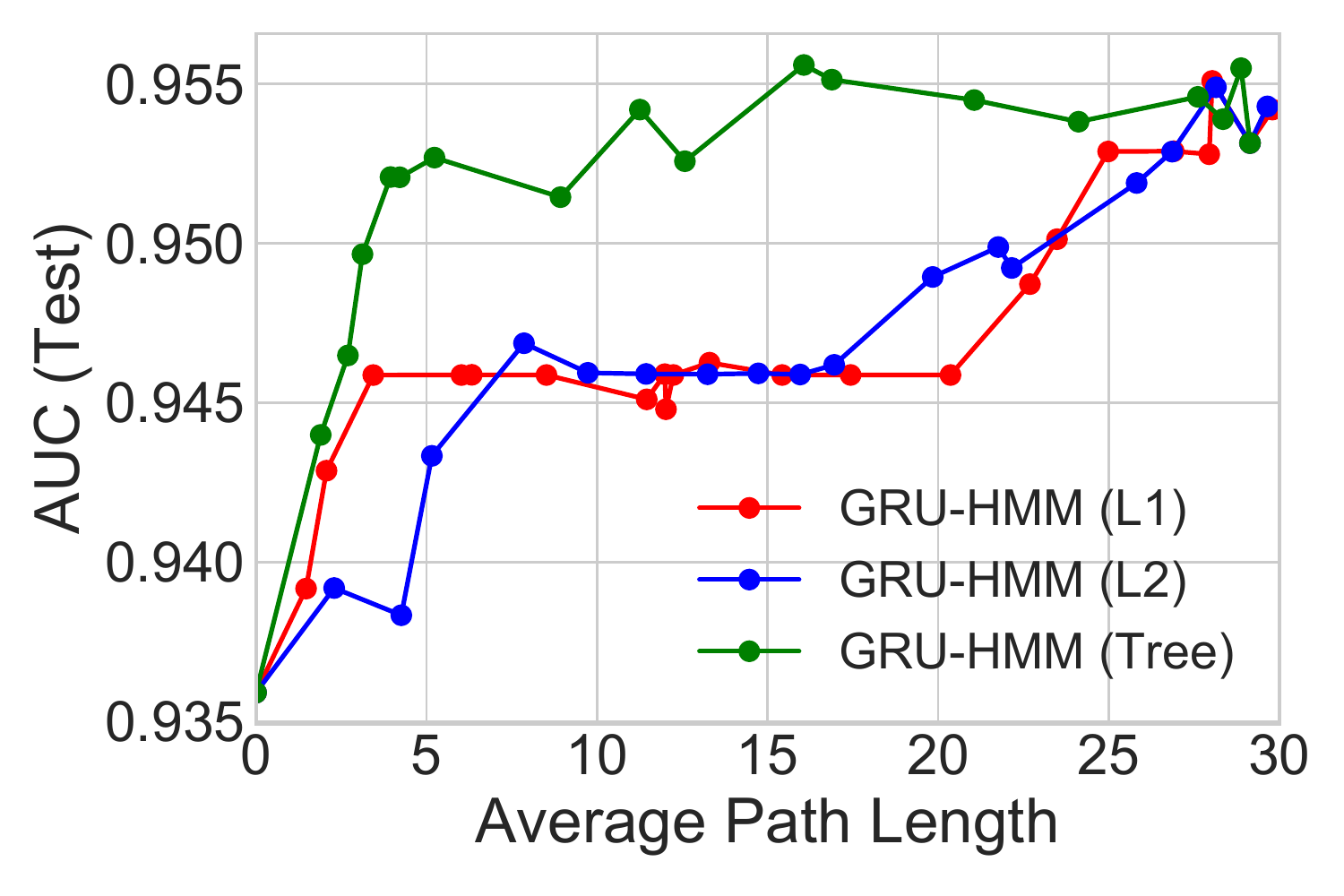}
        \caption{GRU-HMM}
        \label{fig:toy:gruhmm:plot}
    \end{subfigure}
    \caption{
      \emph{Toy Signal-and-Noise HMM Task:}
      \emph{(a)-(b)} Decision trees trained to mimic predictions of GRU models at different regularization strengths $\lambda$; as expected, increasing $\lambda$ decreases the size of the learned trees. Decision tree (b) suggests the model learns to predict positive output (blue) if and only if ``x[0] == 1 and s[3] == 1 and s[4] == 0''.
      This simple description is consistent with the true rule used to generate labels for our dataset: assign positive label only if first dimension is on (x[0] == 1) and first state is active (the emission probability vector for this state is: [.5 .5 .5 .5 0 $\ldots$]). \emph{(c,d)} Tree-regularization produces simpler models (as measured by APL) with higher prediction quality (AUC) across range of regularization strengths $\lambda$ for the GRU (c) and GRU-HMM (d).
    }
    \label{fig:results:toy-signal-and-noise-hmm}
\end{figure}

Next, we analyze the performance of tree regularization on synthetic timeseries data. We generated a toy dataset of $N=100$ sequences, each with $T=50$ timesteps. Each timestep has a data vector $\mathbf{x}_{nt}$ of 14 binary features and a single binary output label $\mathbf{y}_{nt}$.  The data comes from two separate HMM processes. First, a ``signal'' HMM generates the first 7 data dimensions from 5 well-separated states.  Second, an independent ``noise'' HMM generates the remaining 7 data dimensions from a different set of 5 states.  The transition and emission matrices for both HMMs are shown in Fig.~\ref{fig:toy-hmm}. The probabilities were chosen to make it difficult for a new HMM to learn. Each timestep's output label $\mathbf{y}_{nt}$ is produced by a rule involving \emph{both} the signal HMM's generated observations and the signal HMM's hidden state: the target is 1 at timestep $t$ only if both the first signal state is active and the first observation is turned on.  We deliberately designed the generation so that neither logistic regression on inputs $\mathbf{x}_n$ alone nor a GRU model that makes predictions from hidden states alone can perfectly separate this data.

\begin{figure}[!h]
  \centering
  \begin{subfigure}[b]{0.24\linewidth}
      \[
          \begin{psmallmatrix}
          .5 & .5 & .5 & .5 & 0 & 0 & 0 \\
          .5 & .5 & .5 & .5 & .5 & 0 & 0 \\
          .5 & .5 & .5 & 0 & .5 & 0 & 0 \\
          .5 & .5 & .5 & 0 & 0 & .5 & 0 \\
          .5 & .5 & .5 & 0 & 0 & 0 & .5
          \end{psmallmatrix}
      \]
      \caption{Signal: Emission}
      \label{fig:emission:hmm}
  \end{subfigure}
  \begin{subfigure}[b]{0.24\linewidth}
      \[
          \begin{psmallmatrix}
          .7 & .3 & 0 & 0 & 0 \\
          .5 & .25 & .25 & 0 & 0 \\
          0 & .25 & .5 & .25 & 0 \\
          0 & 0 & .25 & .25 & .5 \\
          0 & 0 & 0 & .5 & .5
          \end{psmallmatrix}
      \]
      \caption{Signal: Transition}
      \label{fig:transition:hmm}
  \end{subfigure}
  \begin{subfigure}[b]{0.24\linewidth}
      \[
          \begin{psmallmatrix}
          .5 & .5 & .5 & 0 & 0 & 0 & 0 \\
          0 & .5 & .5 & .5 & 0 & 0 & 0 \\
          0 & 0 & .5 & .5 & .5 & 0 & 0 \\
          0 & 0 & 0 & .5 & .5 & .5 & 0 \\
          0 & 0 & 0 & 0 & .5 & .5 & .5
          \end{psmallmatrix}
      \]
      \caption{Noise: Emission}
      \label{fig:emission:hmm2}
  \end{subfigure}
  \begin{subfigure}[b]{0.24\linewidth}
      \[
          \begin{psmallmatrix}
          .2 & .2 & .2 & .2 & .2 \\
          .2 & .2 & .2 & .2 & .2 \\
          .2 & .2 & .2 & .2 & .2 \\
          .2 & .2 & .2 & .2 & .2 \\
          .2 & .2 & .2 & .2 & .2
          \end{psmallmatrix}
      \]
      \caption{Noise: Transition}
      \label{fig:transition:hmm}
  \end{subfigure}
  \caption{Emission (5 states vs 7 features) and transition probabilities for the signal HMM (a, b) and noise HMM (c, d). We emphasize that to output 1, the signal HMM must be in state 1 and the first input feature must be 1.}
  \label{fig:toy-hmm}
\end{figure}

As with the MLP, each regularizer (tree, L2, L1) is applied to the output node of the GRU across a range of strength parameters $\lambda$ (see orange triangle in Figure~\ref{fig:what-we-regularize}). In training, we used 25 hidden dimensions for GRU models and 5 states for the HMM component of the GRU-HMM. All other choices are identical to the 2D Parabola setting.

Figure~\ref{fig:results:toy-signal-and-noise-hmm} compare the performance of regularized GRU and GRU-HMM models on the signal-and-noise HMM dataset. Since we can no longer easily visualize the decision boundary, we rely on plots like Figure~\ref{fig:results:toy-signal-and-noise-hmm}(c,d) to measure regularization effectiveness. Many of the same patterns from the 2D Parabola experiments emerge here: tree regularized GRU models achieve much higher (held-out) AUC at lower APL. Further, L$_1$ and L$_2$ are quite unreliable at high regularization strengths, doing worse than a decision tree at low APL. All regularized models converge to the same performance as APL approaches 0 (random choice) and infinity (unregularized). Additionally, we include results for the GRU-HMM (d) whose performance is lower bounded by the performance of a standalone HMM (notice the scale of the y-axis). However, as before, tree regularization on the ``GRU component" of the GRU-HMM quickly reaches near maximum performance with small APL (around 5). We hypothesize this is largely due to the compactly expressive nature of axis-aligned decision boundaries. Finally, Figure~\ref{fig:results:toy-signal-and-noise-hmm}(a,b) show two ``distilled" decision trees that are used to approximate the deep model in the last epoch of training. We can see that for small regularization strengths (a), the distilled tree is large and difficult to interpret. For larger strengths (b), the tree recovers the true generative process: predict positive output if and only if ``x[0] == 1 and s[3] == 1 and s[4] == 0''. The first component (x[0] == 1) represents the first observation being 1; the second component (s[3] == 1 and s[4] == 0) represents the first state being active (recall that the emission distribution for this is state is [.5 .5 .5 .5 0 $\ldots$]). A decision tree like this can be given to a human to help describe what mappings the deep model as learned. Critically, smaller decision trees are very easy to simulate.

\section{Applications: Real-World Timeseries Data}
Having explored a few synthetic environments, we now evaluate the tree regularizer on several real-world timeseries models in speech recognition and two sectors of healthcare. For each experiment below, we will compare a tree regularized GRU with an identical GRU regularized with L$_1$ or L$_2$. We will also include a decision tree baseline where a tree classifier is fit directly on the observations. Additionally, we will compare the GRU results with GRU-HMM performance to gauge any benefits of residual training. For optimization, we use Adam with a learning rate of 1e-3, a batch size of 256, decision tree hyperparameter $h=1000$, train for 300 epochs, surrogate datasets of size $J=100$, and retrain every 25 steps. Like above, we measure performance with AUC and simulability with APL for all models. Before sharing results, we briefly describe each task and domain.

\subsection{Tasks}
We tested our approach on several real-world tasks: predicting medical outcomes of hospitalized septic patients, HIV therapy outcome prediction, and predicting stop phoneme groups from a selection of English speech recordings. To normalize scales, we independently standardized input features via z-scoring. Like in the demonstrations above, we compare tree regularization to L$_1$ and L$_2$ baselines. Additionally, we compare a tree-regularized deep network to a decision tree classifier.

\begin{itemize}
\item
\emph{Sepsis Critical Care (ICU)}: We study timeseries data for 11\,786 septic ICU patients from the public MIMIC III dataset \cite{johnson2016mimiciii}. We observe at each hour (timestep) $t$ a data vector $\textbf{x}_{nt}$ of 35 vital signs and lab results as well as a label vector $\textbf{y}_{nt}$ of 5 binary outcomes. Hourly data $\textbf{x}_{nt}$ measures continuous input features such as respiration rate (RR), blood oxygen levels (paO$_{2}$), fluid levels, and more. Hourly binary labels $\textbf{y}_{nt}$ include whether the patient died in hospital, whether the patient died after 90 days, and if mechanical ventilation was applied. Models are trained to predict all 5 output dimensions concurrently from one shared embedding. The average sequence length is 15 hours. 7\,070 patients are used in training, 1\,769 for validation, and 294 for test.

\item
\emph{HIV Therapy Outcome (HIV)}: We make use of the EuResist Integrated Database \cite{euresist} for 53,236  patients diagnosed with HIV. We consider 4-6 month intervals (corresponding to hospital visits) as time steps. Each data vector $\textbf{x}_{nt}$ has 40 features, including blood counts, viral load measurements and lab results. Each output vector $\textbf{y}_{nt}$ has 15 binary labels, including whether a therapy was successful in reducing viral load to below detection limits, if therapy caused CD4 blood cell counts to drop to dangerous levels (indicating AIDS), or if the patient suffered adherence issues to medication. The average sequence length is 14 steps. 37\,618 patients are used for training; 7\,986 for testing, and 7\,632 for validation.

\item
\textit{Phonetic Speech (TIMIT)}: Timeseries data containing broadband recordings of 630 speakers of eight major dialects of American English reading ten phonetically rich sentences \cite{garofolo1993timit}. Each sentence contains time-aligned phonetic transcriptions of 60 phonemes. We focus on the problem of distinguishing stop phonemes (those that stop the flow of air, such as ``b'', ``d'', or ``g'') from non-stops. Each timestep has one binary output $\textbf{y}_{nt}$ indicating whether a stop phoneme occurs or not. There are 26 continuous features for each input vector $\textbf{x}_{nt}$ representing the Mel-frequency cepstral coefficients and derivatives of the acoustic signal.  There are 6\,303 sequences: which we split into 3\,697 for training, 925 for validation, and 1\,681 for testing. The average length is 614 tokens in a sequence.
\end{itemize}

\subsection{Results and Analysis}

The results on ICU, HIV, and TIMIT share many consistent characteristics. We summarize the many experiments with analysis on common patterns and provide a few takeaways.

\begin{figure}[h!]
    \centering
    \begin{subfigure}[b]{0.23\linewidth}
        \includegraphics[width=\linewidth]{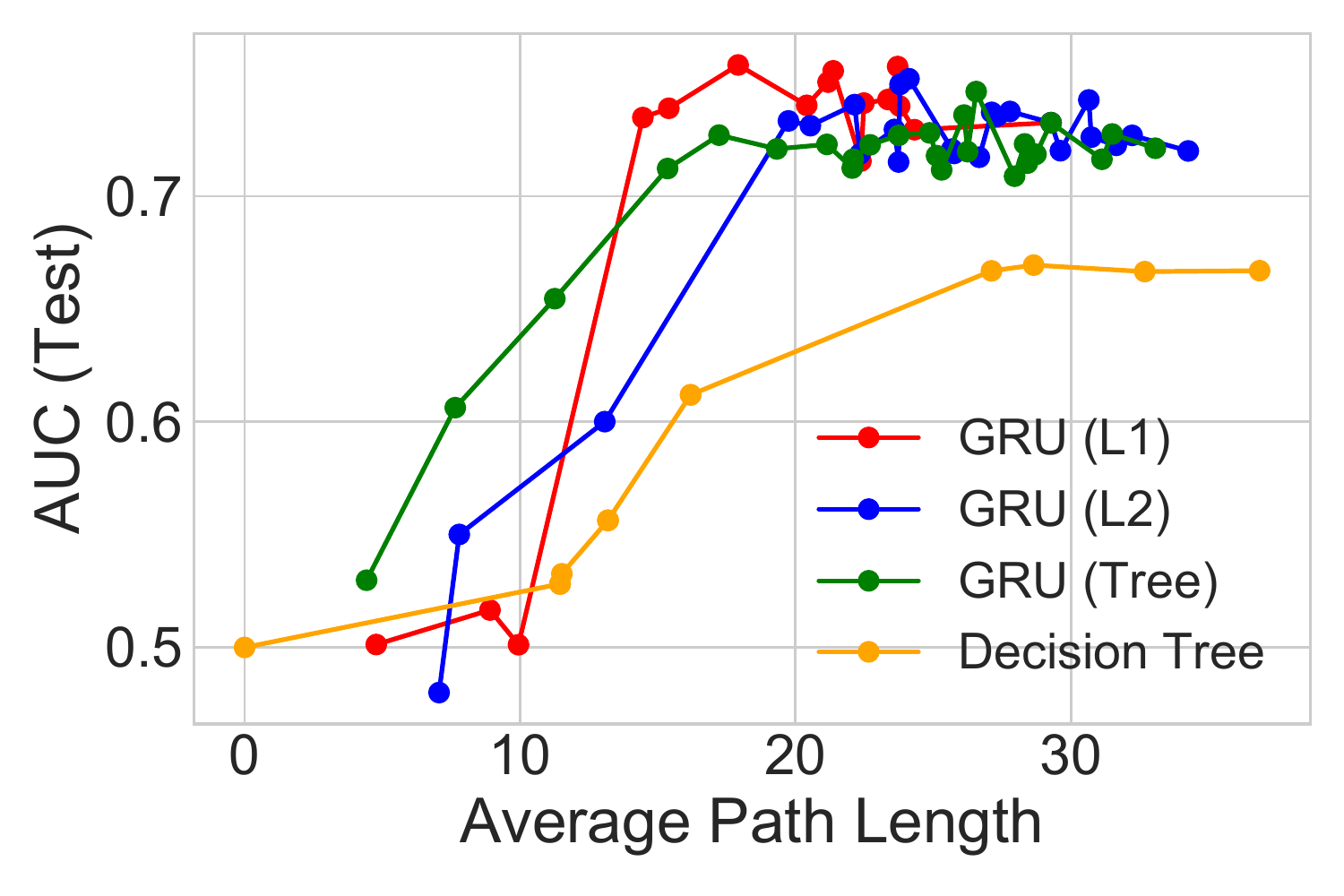}
        \caption{Hospital Mortality}
        \label{fig:results:sepsis:gru:mortality:trace_plots}
    \end{subfigure}
    \begin{subfigure}[b]{0.25\linewidth}
        \includegraphics[width=0.8\linewidth,height=3cm]{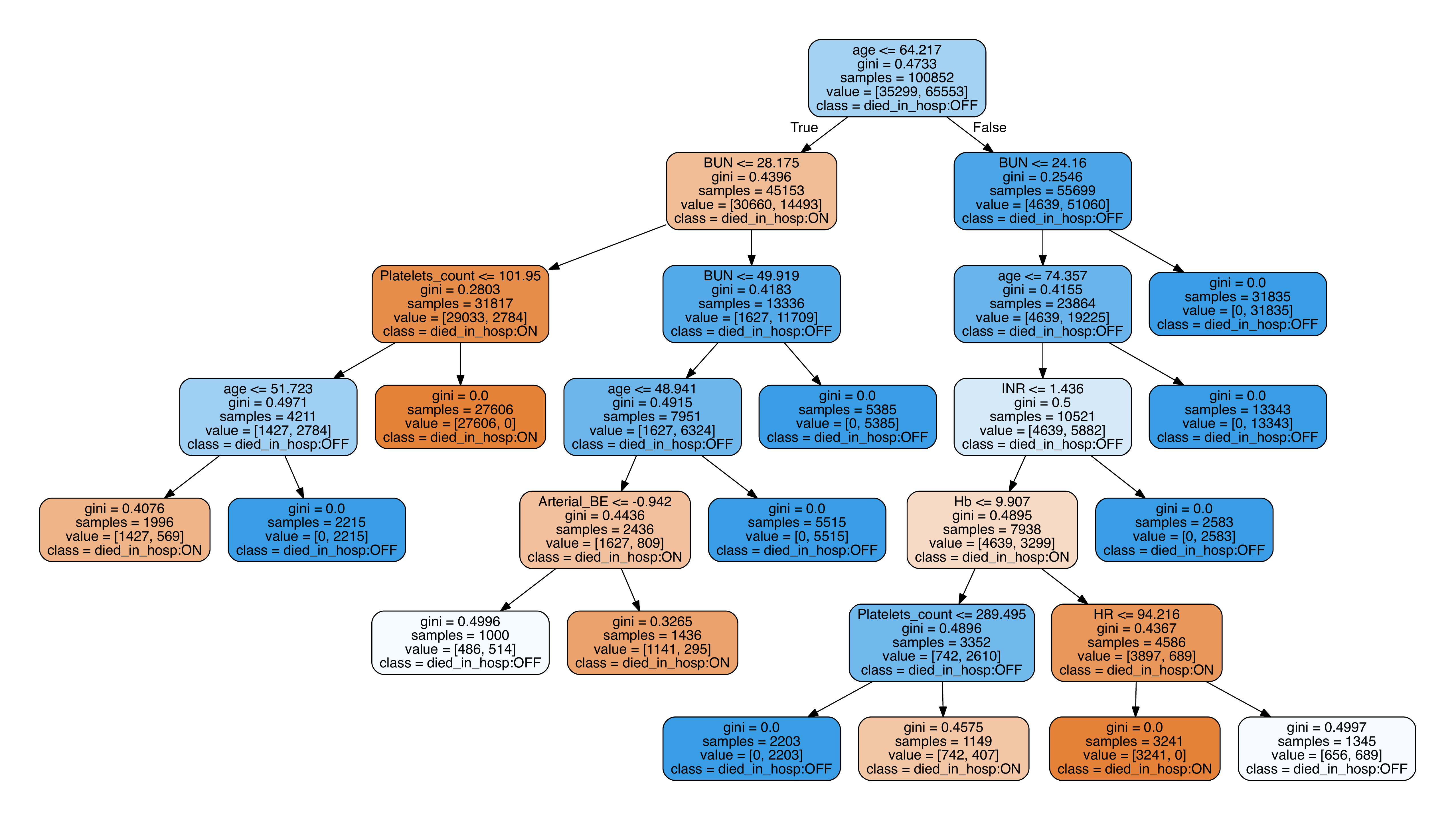}
        \caption{Hospital Mortality}
        \label{fig:results:sepsis:gru:mortality:tree}
    \end{subfigure}
    \begin{subfigure}[b]{0.23\linewidth}
        \includegraphics[width=\linewidth]{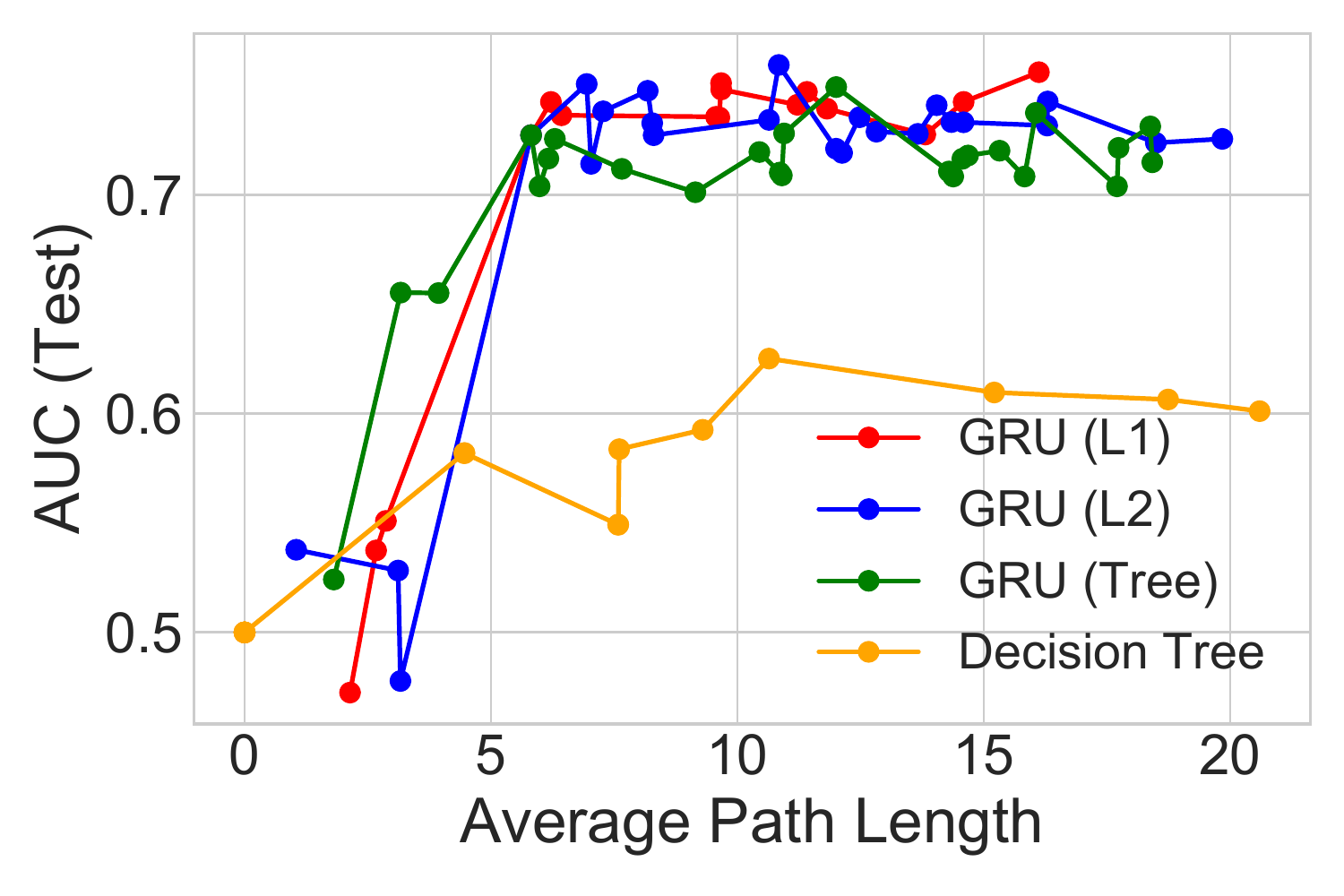}
        \caption{90-day Mortality}
        \label{fig:results:sepsis:gru:90mortality:trace_plots}
    \end{subfigure}
    \begin{subfigure}[b]{0.25\linewidth}
        \includegraphics[width=0.8\linewidth,height=3cm]{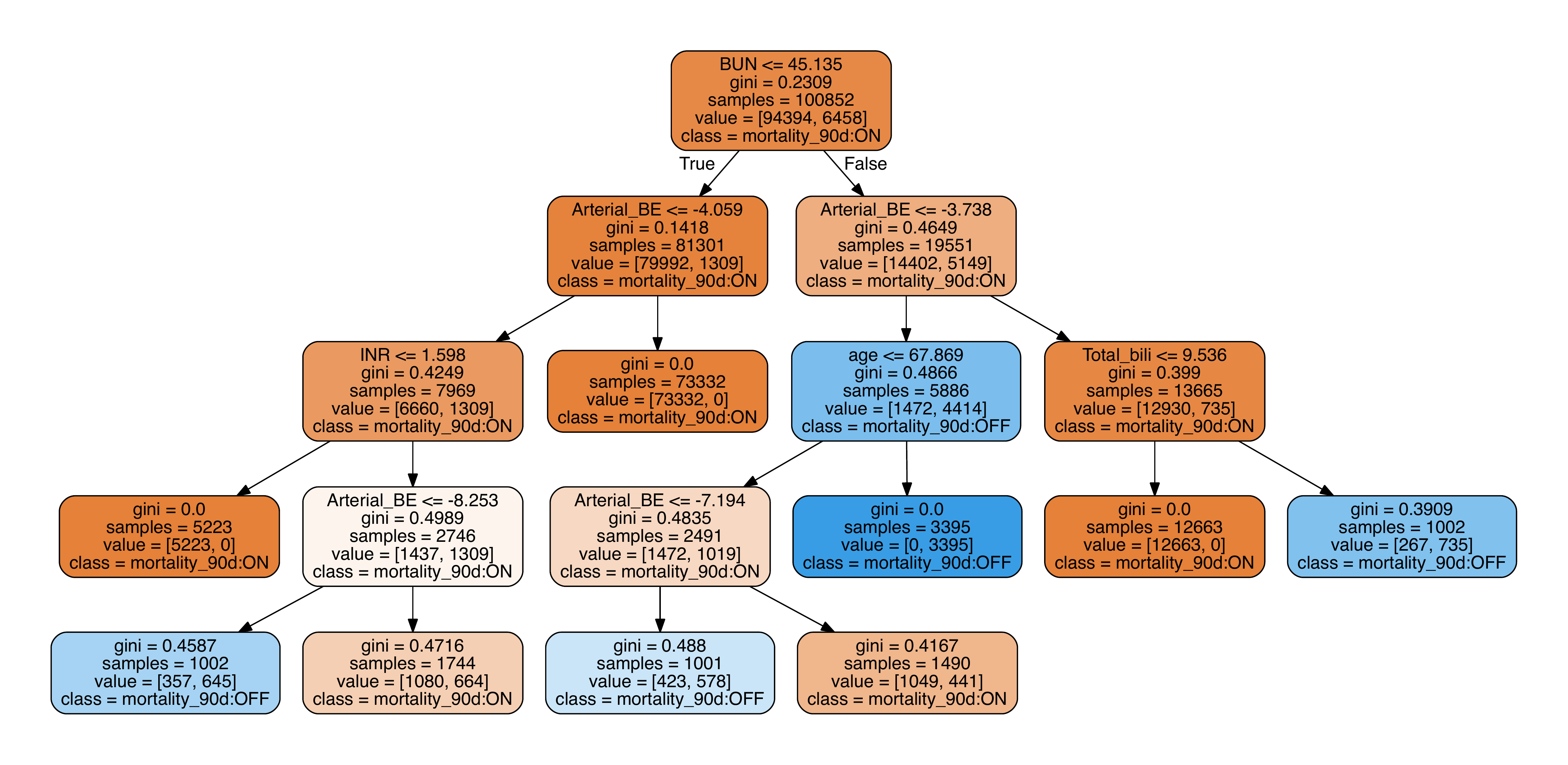}
        \caption{90-day Mortality}
        \label{fig:results:sepsis:gru:90mortality:tree}
    \end{subfigure}
    \begin{subfigure}[b]{0.23\linewidth}
        \includegraphics[width=\linewidth]{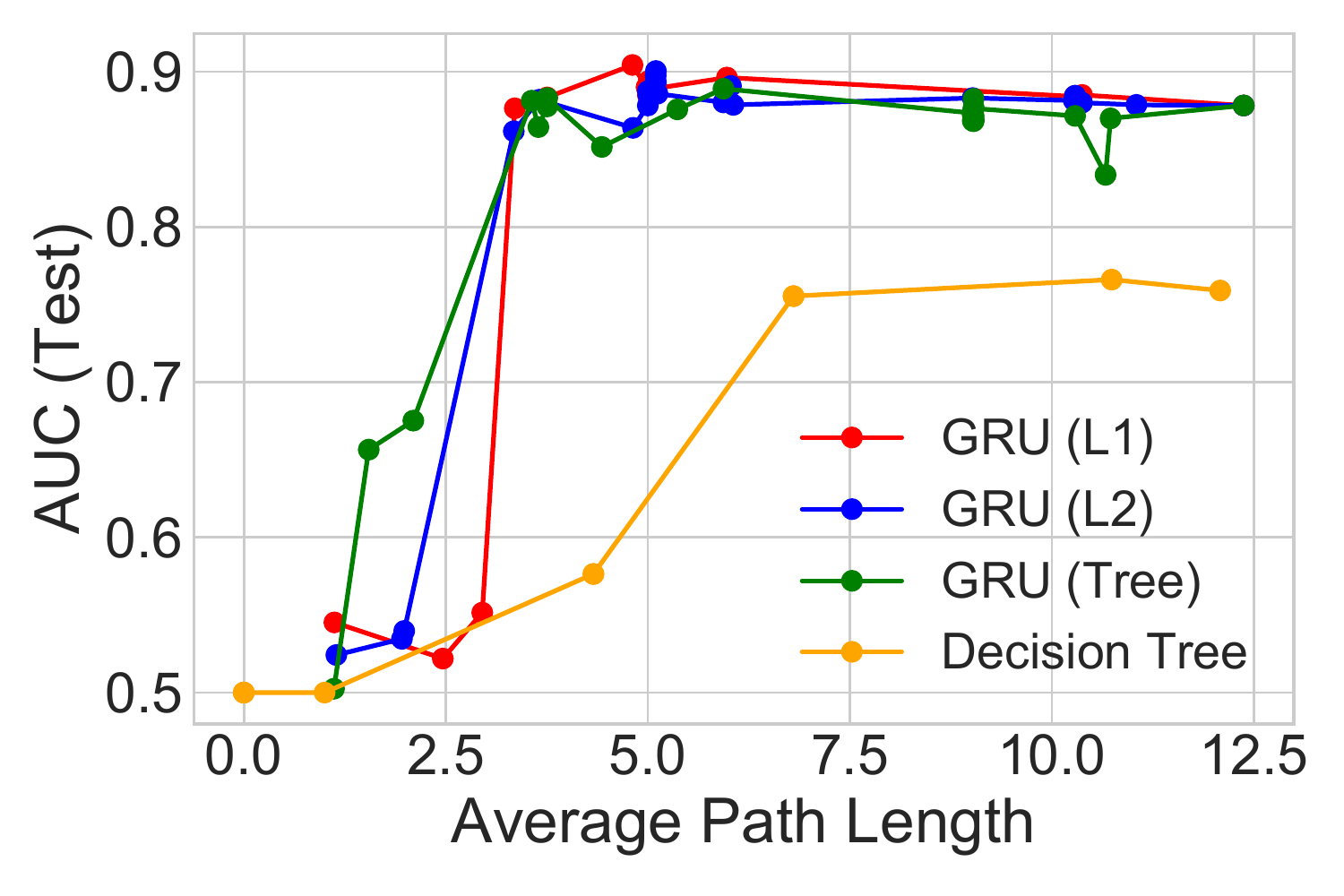}
        \caption{Mech. Vent.}
        \label{fig:results:sepsis:gru:mechvent:trace_plots}
    \end{subfigure}
    \begin{subfigure}[b]{0.22\linewidth}
        \includegraphics[width=0.8\linewidth]{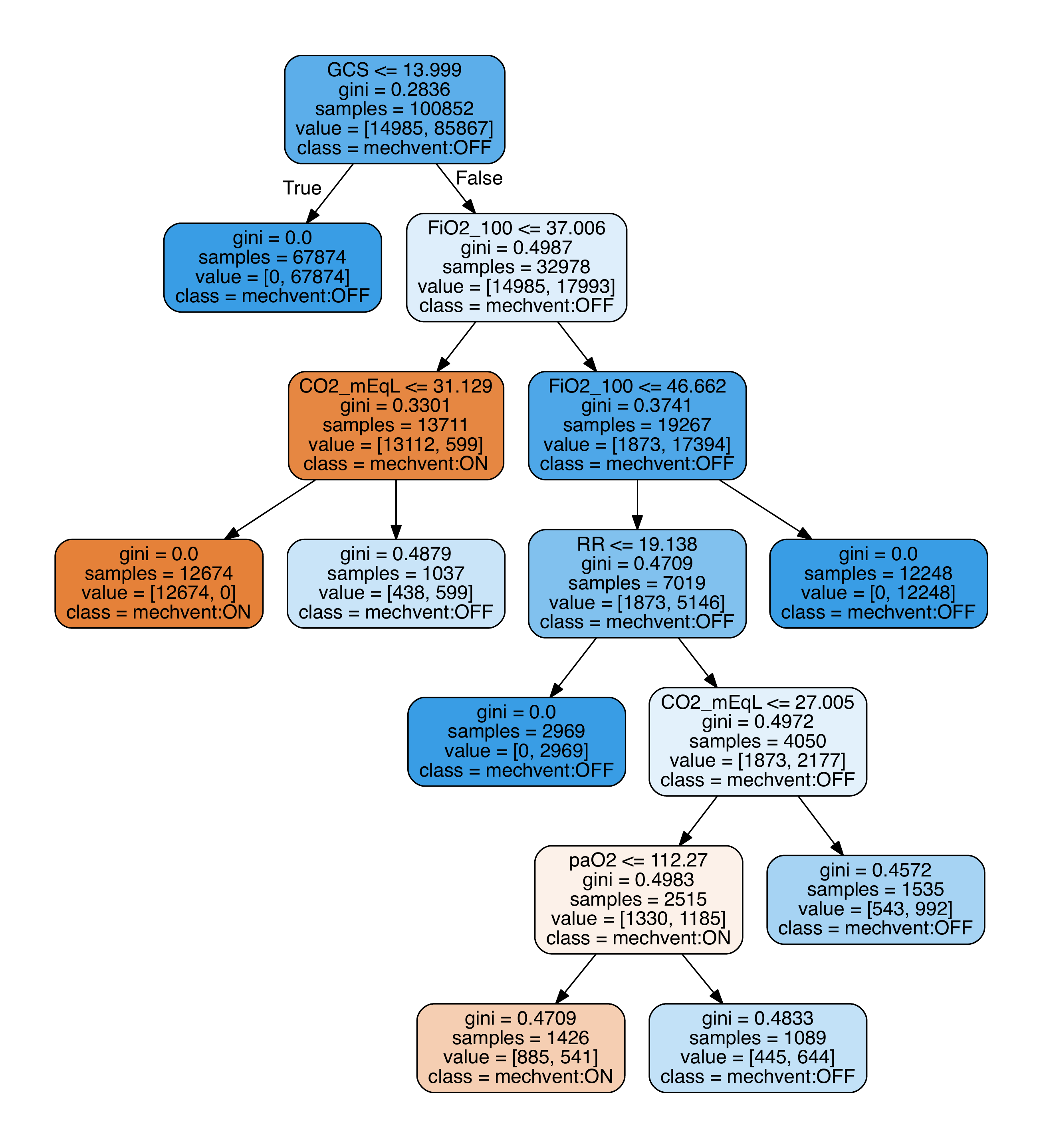}
        \caption{Mech. Vent.}
        \label{fig:results:sepsis:gru:mechvent:tree}
    \end{subfigure}
    \begin{subfigure}[b]{0.23\linewidth}
        \includegraphics[width=\linewidth]{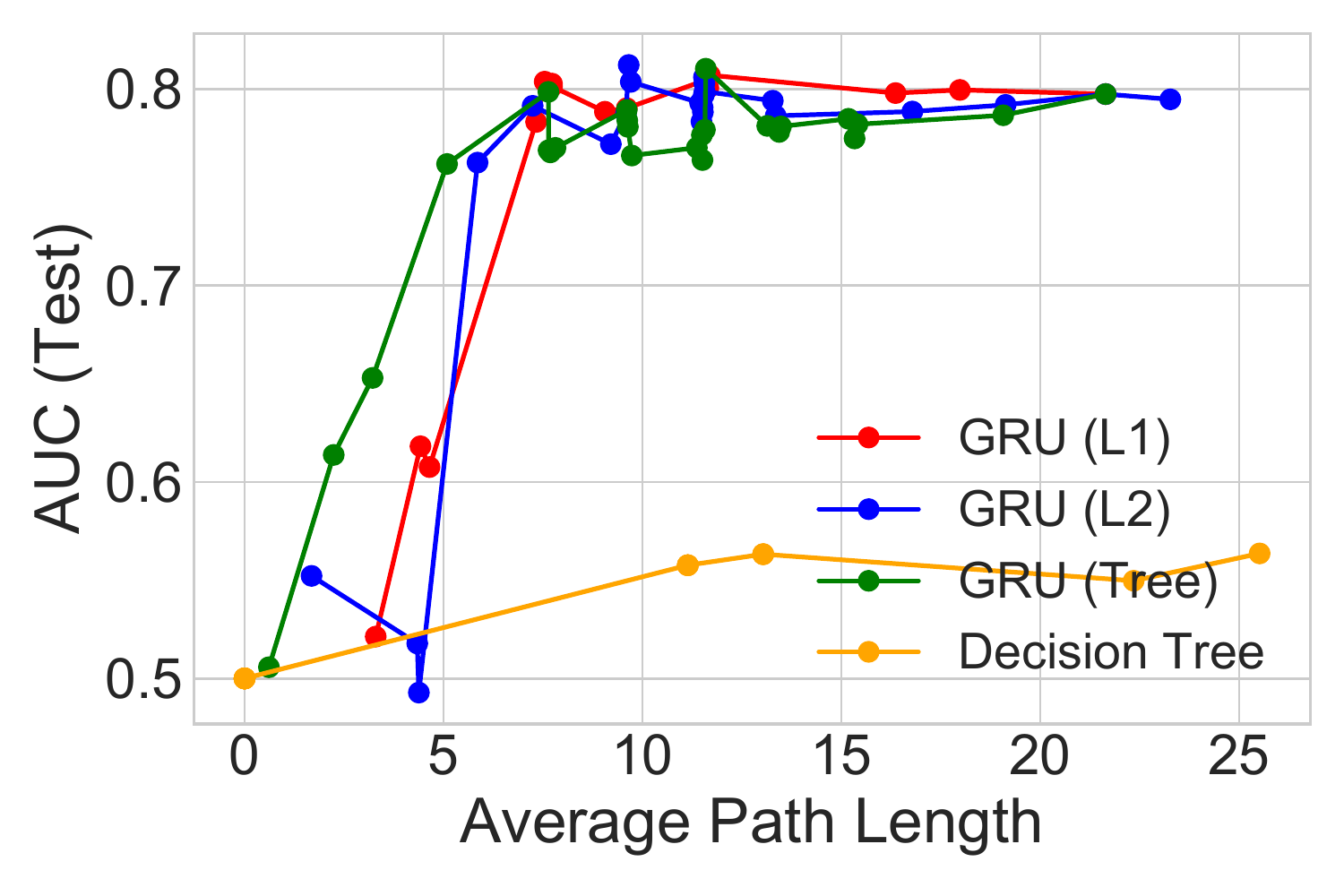}
        \caption{Max Vaso.}
        \label{fig:results:sepsis:gru:vaso:trace_plots}
    \end{subfigure}
    \begin{subfigure}[b]{0.22\linewidth}
        \includegraphics[width=0.8\linewidth]{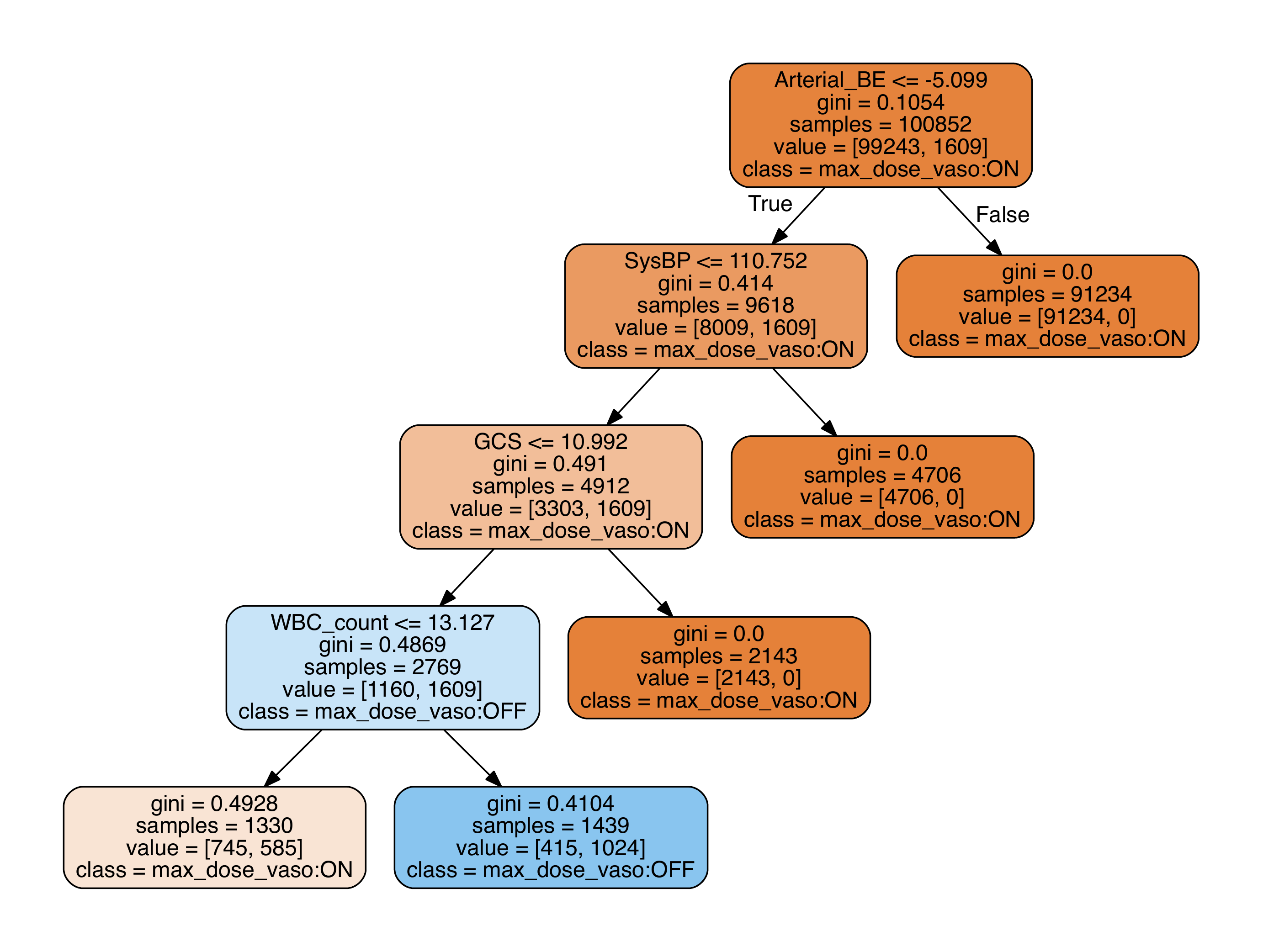}
        \caption{Max Vaso.}
        \label{fig:results:sepsis:gru:vaso:tree}
    \end{subfigure}
    \caption{
\emph{SEPSIS task} -- Study of different regularizers for a GRU model with 100 states, trained to jointly predict 5 binary outcomes.  Panels \emph{(a,c,e,g)} show AUC vs. APL for 4 of the 5 outcomes; in all cases, tree regularization provides higher accuracy in the target regime of low-complexity decision trees. Panels \emph{(b,d,f,h)} show the associated decision trees for $\lambda = 2\,000$; these were found by clinically interpretable by an ICU clinician.
}
\label{fig:results:sepsis}
\end{figure}

\begin{figure}[h!]
    \centering{
    \begin{subfigure}[b]{0.23\linewidth}
        \includegraphics[width=\linewidth]{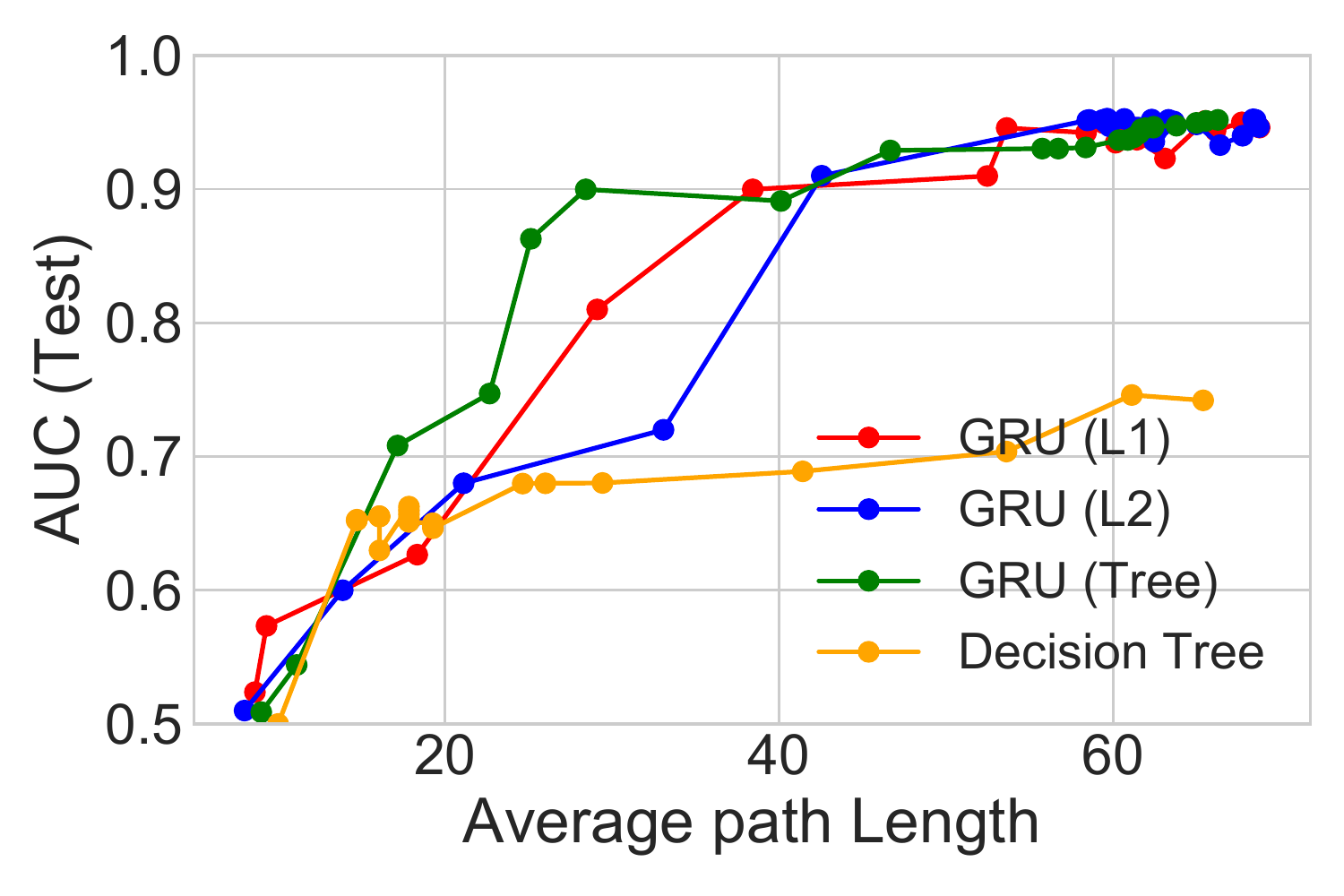}
        \caption{TIMIT ``Stop"}
        \label{fig:timit:gru:trace_plots}
    \end{subfigure}
    \begin{subfigure}[b]{0.23\linewidth}
        \includegraphics[width=\linewidth]{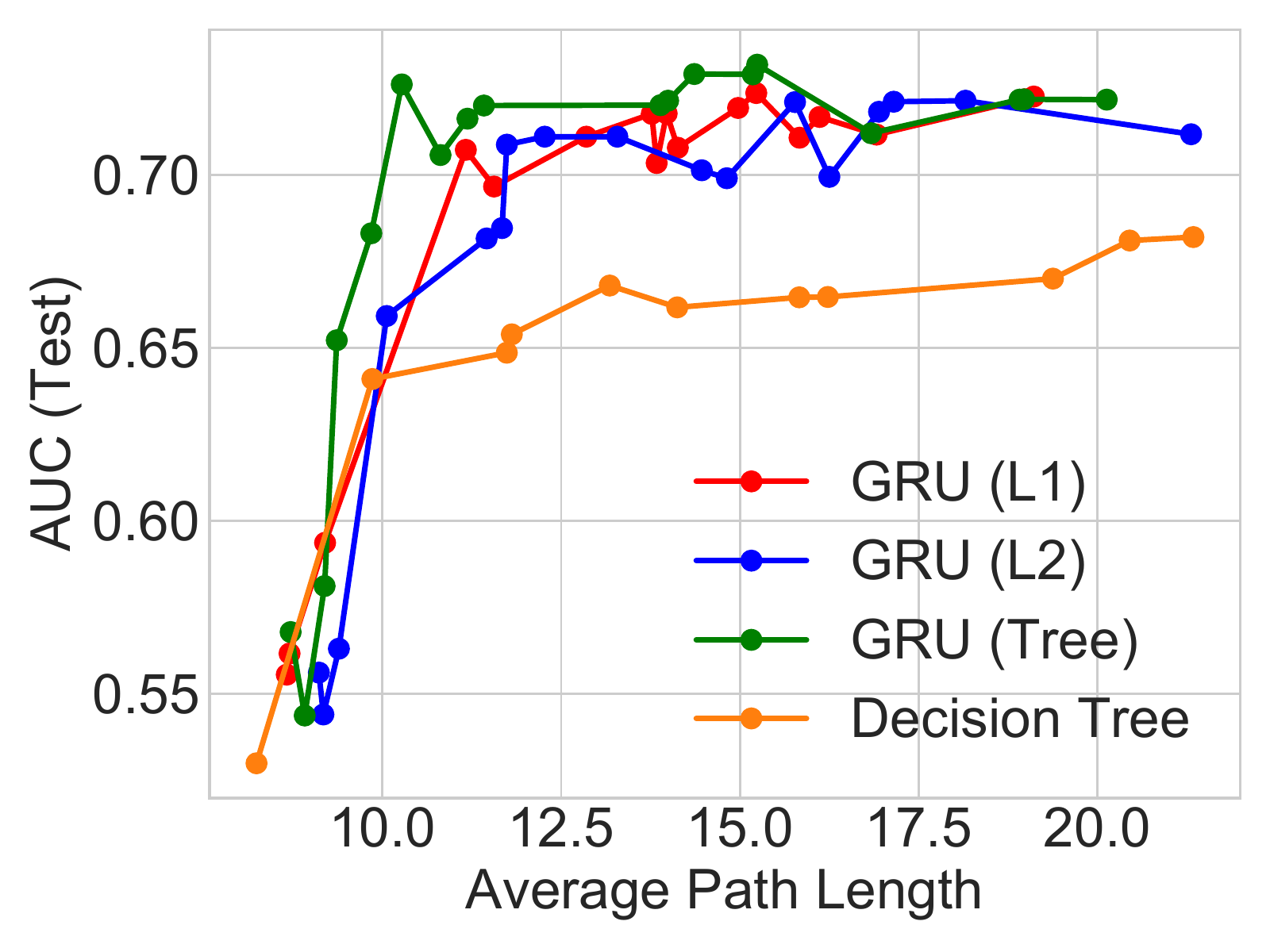}
        \caption{HIV: CD4$^+$}
        \label{fig:hiv:cd4}
    \end{subfigure}
       \begin{subfigure}[b]{0.23\linewidth}
        \includegraphics[width=\linewidth]{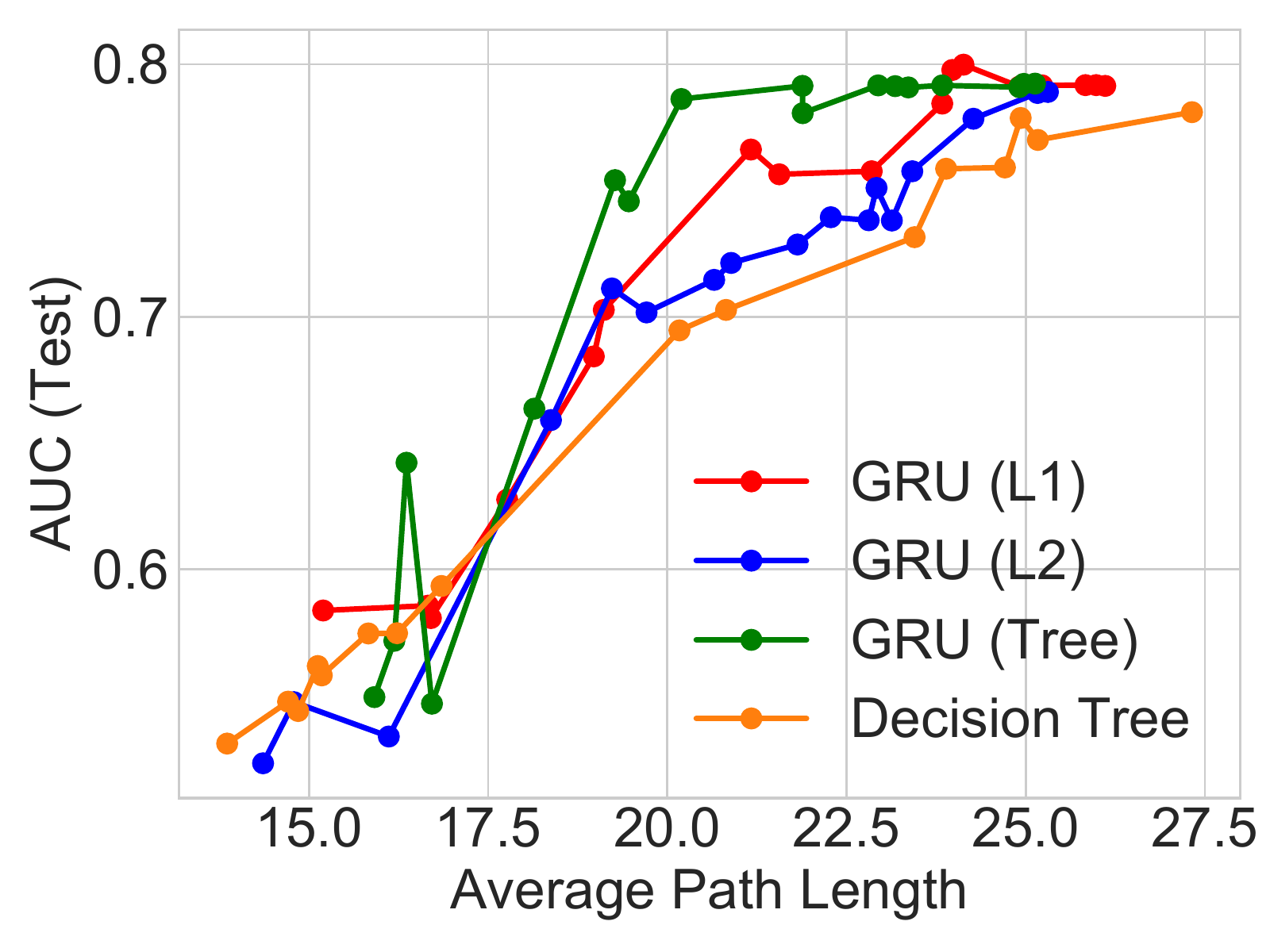}
        \caption{HIV Therapy}
        \label{fig:hiv:therapy}
    \end{subfigure}
    \begin{subfigure}[b]{0.23\linewidth}
        \includegraphics[width=\linewidth]{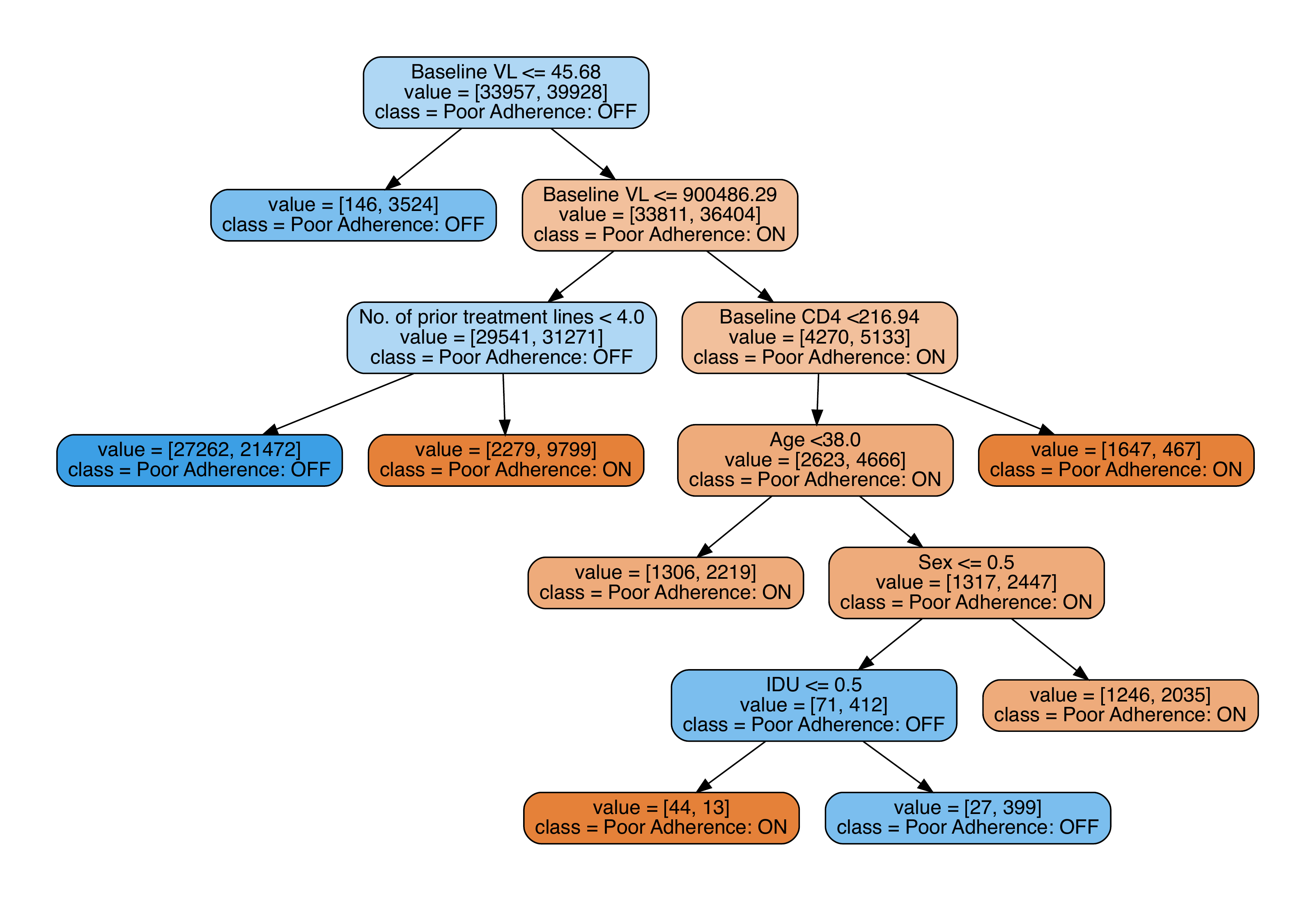}
        \caption{HIV Therapy}
        \label{fig:hiv:adherence}
    \end{subfigure}}
\caption{\emph{TIMIT and HIV tasks:} -- Study of different regularizers for a GRU model with 75 states. Panels \emph{(a)-(c)} are tradeoff curves showing how predictive power and decision-tree complexity evolve with increasing strength of L$_1$, L$_2$ or tree regularization in both TIMIT (stop phoneme prediction) and HIV (CD4$^+ \leq 200$ cells/ml and therapy adherence prediction). The TIMIT task has only one binary outcome. However, for the HIV task, the GRU is trained to jointly predict 15 binary outcomes, of which 2 are shown here in Panels \emph{(b)-(c)}. The decision tree associated with HIV adherence is shown in \emph{(d)}.}
\label{fig:results:timit}
\end{figure}

\paragraph{Tree-regularized models have fewer nodes than other forms of regularization.}
Across tasks, we see that in the target regime of small decision trees (low APLs), our proposed regularization achieves higher prediction quality (higher AUCs).  In the signal-and-noise HMM task, tree regularization (green line in Figure~\ref{fig:results:toy-signal-and-noise-hmm}(d)) achieves AUC
values near 0.9 when its trees have an average path length of 10. Similar models with L$_1$ or L$_2$ regularization reach this AUC only with trees that are nearly double in complexity (APL over 25). On both the SEPSIS (Figure~\ref{fig:results:sepsis}) and TIMIT (Figure~\ref{fig:timit:gru:trace_plots}), we see considerable gains in accuracy over other regularizers---AUC differences of 0.05 to 0.15---for path lengths of 20-30. On the HIV task in Figure~\ref{fig:hiv:cd4}, we see AUC differences of between 0.03 and 0.15 for path lengths of 10-15. Similarly, on the other HIV outcomes in Figures~\ref{fig:hiv:therapy}-\ref{fig:hiv:adherence}, we see AUC differences of between 0.03 and 0.09 for path lengths of 20-30. These gains are particularly useful in determining how to administer subsequent therapies. More specifically, in domains where human-simulability is required, these increases in accuracy in the small-complexity regime can mean the difference between models that provide value on
a task and models that are unusable, either because their performance is too poor or they are uninterpretable.
We emphasize that across all tasks, standalone decision trees (marked by yellow dots in line plots) cannot reach this high-accuracy, low-complexity sweet spot, suggesting that tree regularization still enables neural networks to be nonlinear.

\begin{table}
  \parbox{.45\linewidth}{
    \begin{tabular}{ l | c}
        \toprule
        Dataset & Fidelity \\
        \midrule
        signal-and-noise HMM & 0.8762 \\
        SEPSIS (In-Hospital Mortality) & 0.8144\\
        SEPSIS (90-Day Mortality) & 0.8845\\
        SEPSIS (Mech. Vent.) & 0.9008\\
        SEPSIS (Median Vaso.) & 0.9166\\
        SEPSIS (Max Vaso.) & 0.9260\\
        HIV (CD4$^{+}$ below 200) & 0.8426 \\
        HIV (Therapy Success) & 0.8761 \\
        HIV (Mortality) & 0.9318\\
        HIV (Poor Adherence) & 0.9014 \\
        HIV (AIDS Onset) & 0.9344\\
        TIMIT & 0.8477\\
        \bottomrule
    \end{tabular}
    \caption{Fidelity of predictions from our trained deep GRU and its corresponding decision tree. Fidelity is defined as the percentage of test examples on which the prediction made by a tree agrees with the deep model \cite{craven1996extracting}.}
    \label{table:fidelity}
  }
  \hfill
  \parbox{.45\linewidth}{
    \begin{tabular}{ l | l | c}
        \toprule
        Dataset & Model & Epoch Time\\
        \midrule
        SEPSIS & HMM & $589.8 \pm 24.1$ \\
        SEPSIS & GRU & $822.3 \pm 11.2$ \\
        SEPSIS & GRU-HMM & $1666.9 \pm 147.0$ \\
        SEPSIS & GRU$^\ddagger$ & $2015.1 \pm 388.1$ \\
        SEPSIS & GRU-HMM$^\ddagger$ & $2443.7 \pm 351.2$ \\
        TIMIT & HMM & $1668.9 \pm 126.9$ \\
        TIMIT & GRU & $2116.8 \pm 438.8$ \\
        TIMIT & GRU-HMM & $3207.2 \pm 651.9$ \\
        TIMIT & GRU$^\ddagger$ & $3977.0 \pm 812.1$ \\
        TIMIT & GRU-HMM$^\ddagger$ & $4601.4 \pm 805.9$ \\
        \bottomrule
    \end{tabular}
    \caption{Training time for a single epoch in seconds on a single Intel Core i5 CPU. The ($\ddagger$) symbol represents using tree regularization. The times for tree regularized models include surrogate training expenses. If we retrain sparsely, then the cost is amortized to close to negligible.}
    \label{table:runtime}
  }
\end{table}

\paragraph{Our learned decision-tree-like boundaries are interpretable.} Recall that a consequence of tree regularization is a distillation of the deep model as a decision tree. Across all tasks, these trees which mimic the predictions of tree-regularized deep models are small enough to simulate by hand and help users grasp the model's nonlinear prediction logic. We have already seen this to be the case for the signal-and-noise HMM task. Similarly, in Figure~\ref{fig:results:sepsis}, we show
decision trees for two sepsis prediction tasks.  We consulted a clinical expert on sepsis treatment, who noted that the trees helped him understand what the models might be doing and thus determine if he would trust the deep model. For example, he said that using FiO$_{2}$, RR, CO$_{2}$ and paO$_{2}$ to predict need for mechanical ventilation (Figure~\ref{fig:results:sepsis:gru:mechvent:tree}) was sensible, as these all measure breathing quality.
In contrast, the in-hospital mortality tree (Figure~\ref{fig:results:sepsis:gru:mortality:tree}) predicts that some young patients with no organ failure have high mortality rates while other young patients with organ failure have low mortality.  These counter-intuitive results led to hypotheses about how uncaptured variables impact the
training process. Such reasoning would not be possible from
simple sensitivity analyses of the deep model. Moreover, our distilled trees for HIV such as those in Figure \ref{fig:hiv:adherence}, are also interpretable. We observe that the baseline viral load and number of prior treatment lines are crucial factors in predicting whether a patient will suffer adherence issues. This is consistent with several medical studies which show that patients with higher viral loads at baseline tend to have faster disease progression, and hence have to take several drug cocktails to potentially combat resistance. This typically makes it more difficult for these patients to adhere to the medication.

\paragraph{Practical runtimes for tree regularization are less than twice that of simpler L2.}
While our tree-regularized GRU with 10 states takes 3977 seconds per epoch on TIMIT, an equivalent L$_2$-regularized GRU takes 2116 seconds per epoch. Thus, our new method has cost less than twice the baseline \emph{even when the path-length surrogate is serially computed}. Because the surrogate $\hat{\Omega}$ will in general be a much smaller model than the target neural model, we expect one could get much smaller per-epoch times by parallelizing the creation of $(\theta,\Omega(\theta))$ training pairs and the surrogate training. Additionally, 3\,977 seconds includes the time needed to train the surrogate. In practice, we do this sparingly, only once every 25 epochs, yielding an amortized per-epoch cost of 2\,191 seconds. More exhaustive runtime results with standard deviations over 10 epochs are in Table~\ref{table:runtime}.

\begin{figure}[!h]
    \centering
    \begin{subfigure}[b]{0.16\linewidth}
        \centering
        \includegraphics[width=\linewidth]{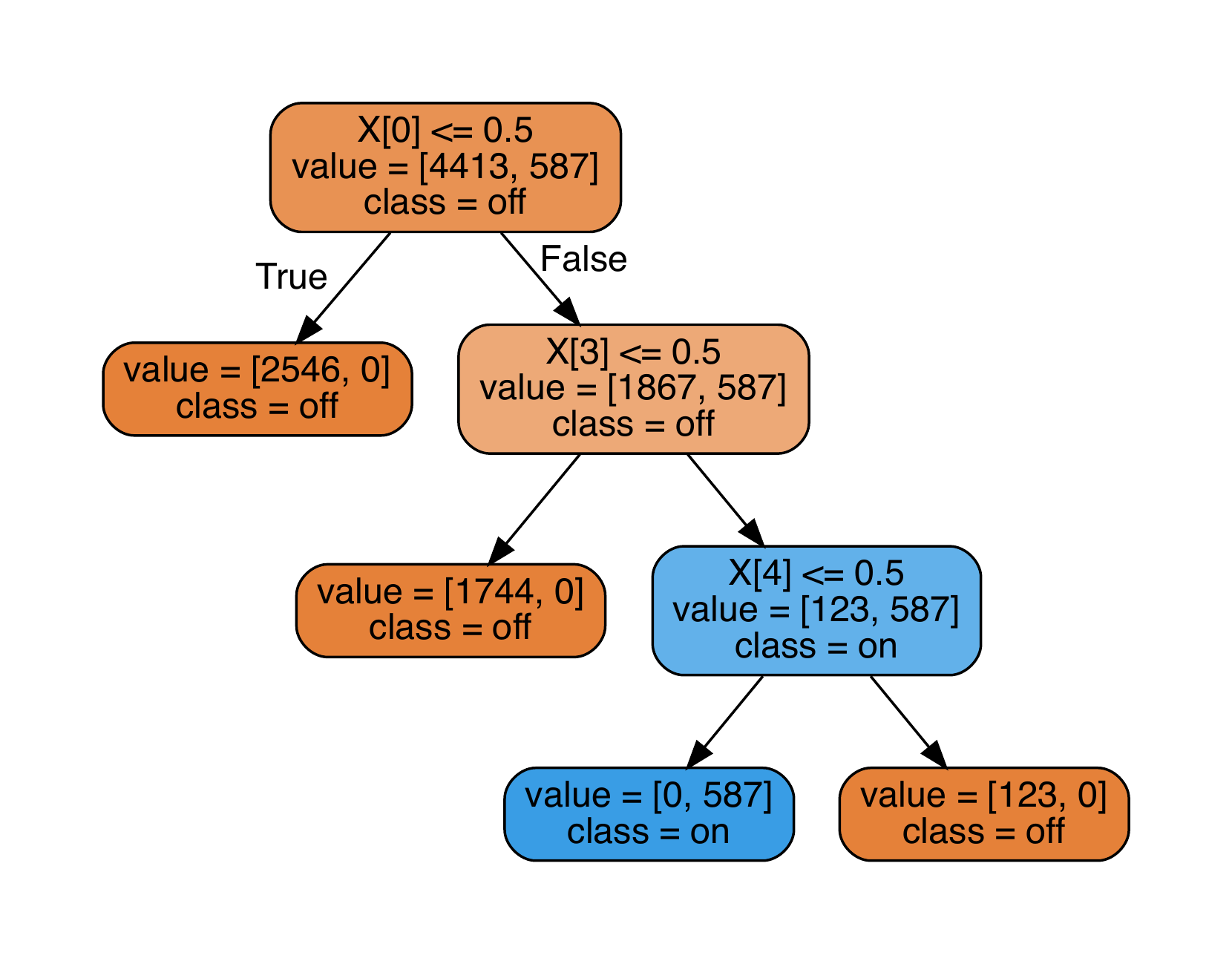}
        \caption{$\frac{7}{10}$ Runs}
        \label{}
    \end{subfigure}
    \begin{subfigure}[b]{0.16\linewidth}
        \centering
        \includegraphics[width=\linewidth]{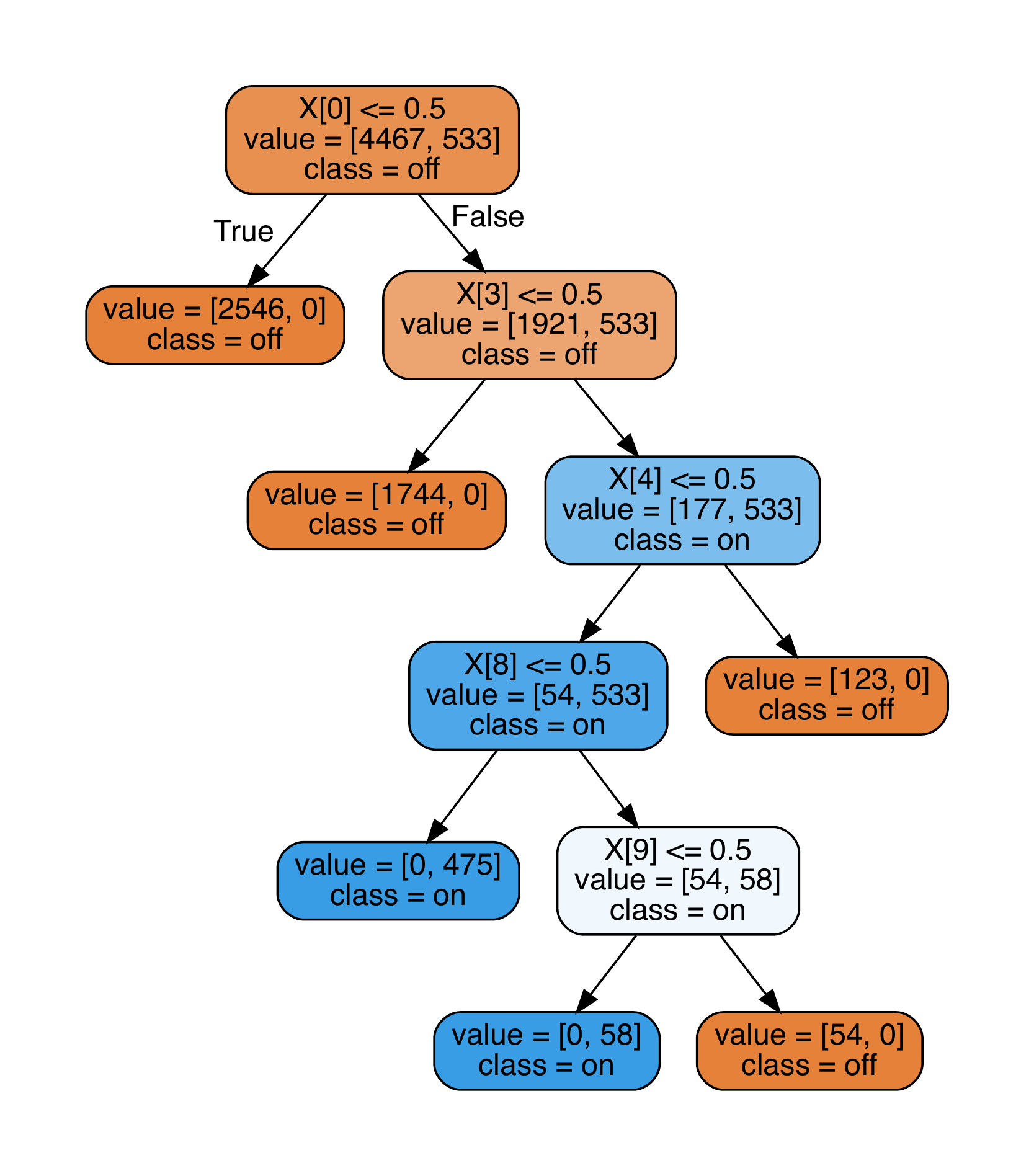}
        \caption{$\frac{2}{10}$ Runs}
        \label{}
    \end{subfigure}
    \begin{subfigure}[b]{0.16\linewidth}
        \centering
        \includegraphics[width=\linewidth]{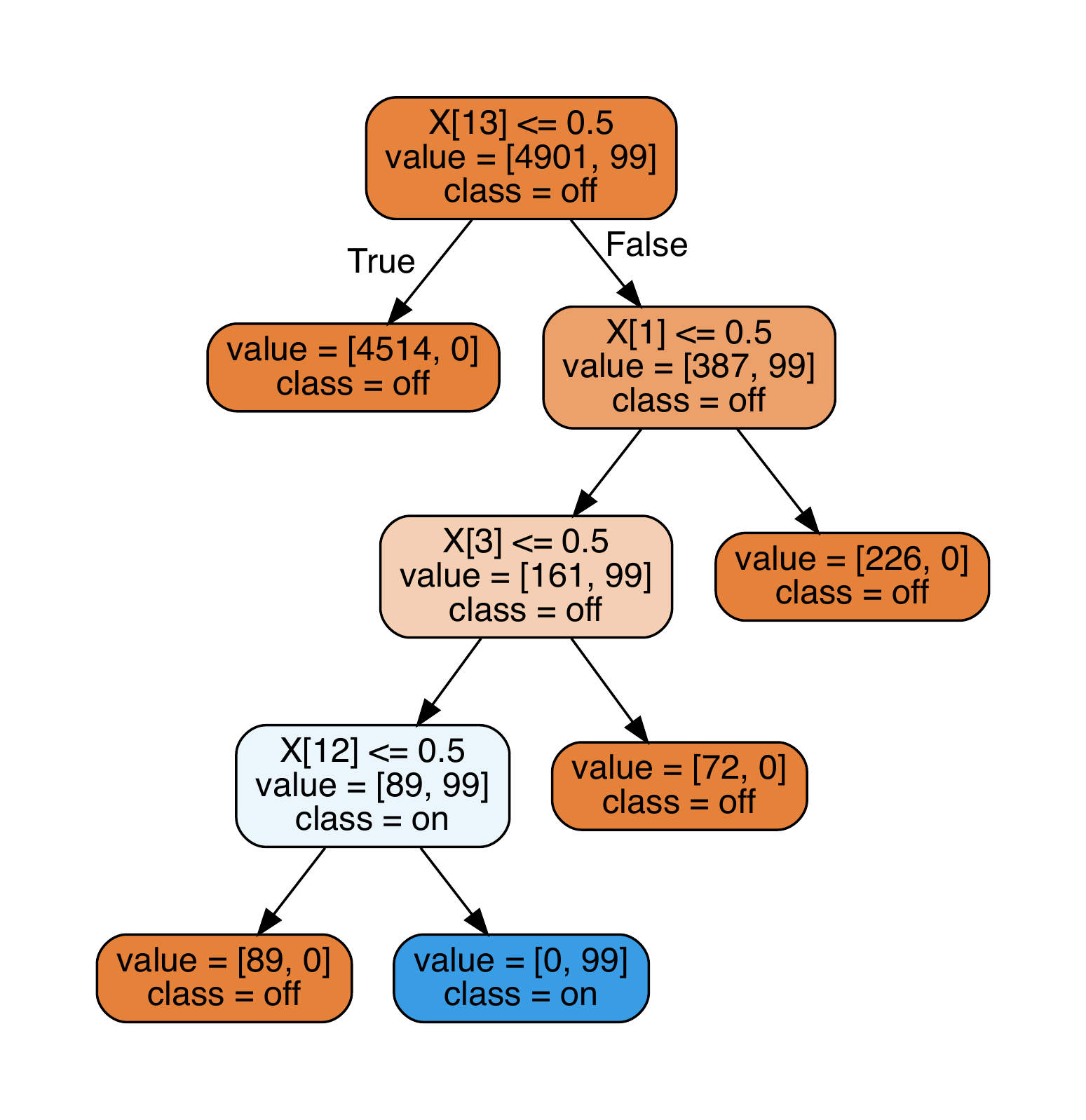}
        \caption{$\frac{1}{10}$ Runs}
        \label{}
    \end{subfigure}
    \begin{subfigure}[b]{0.16\linewidth}
        \centering
        \includegraphics[width=\linewidth]{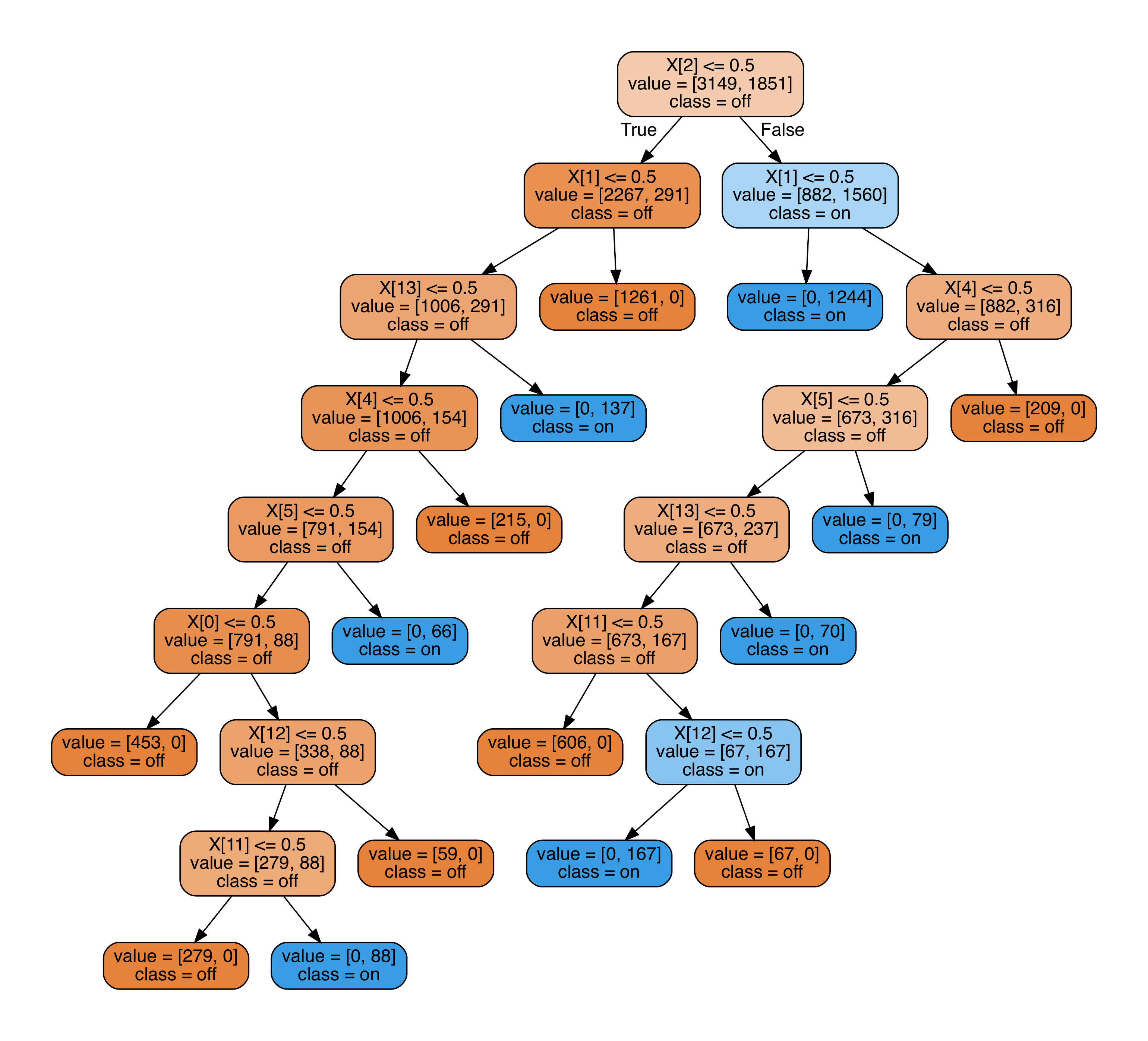}
        \caption{}
        \label{}
    \end{subfigure}
    \begin{subfigure}[b]{0.16\linewidth}
        \centering
        \includegraphics[width=\linewidth]{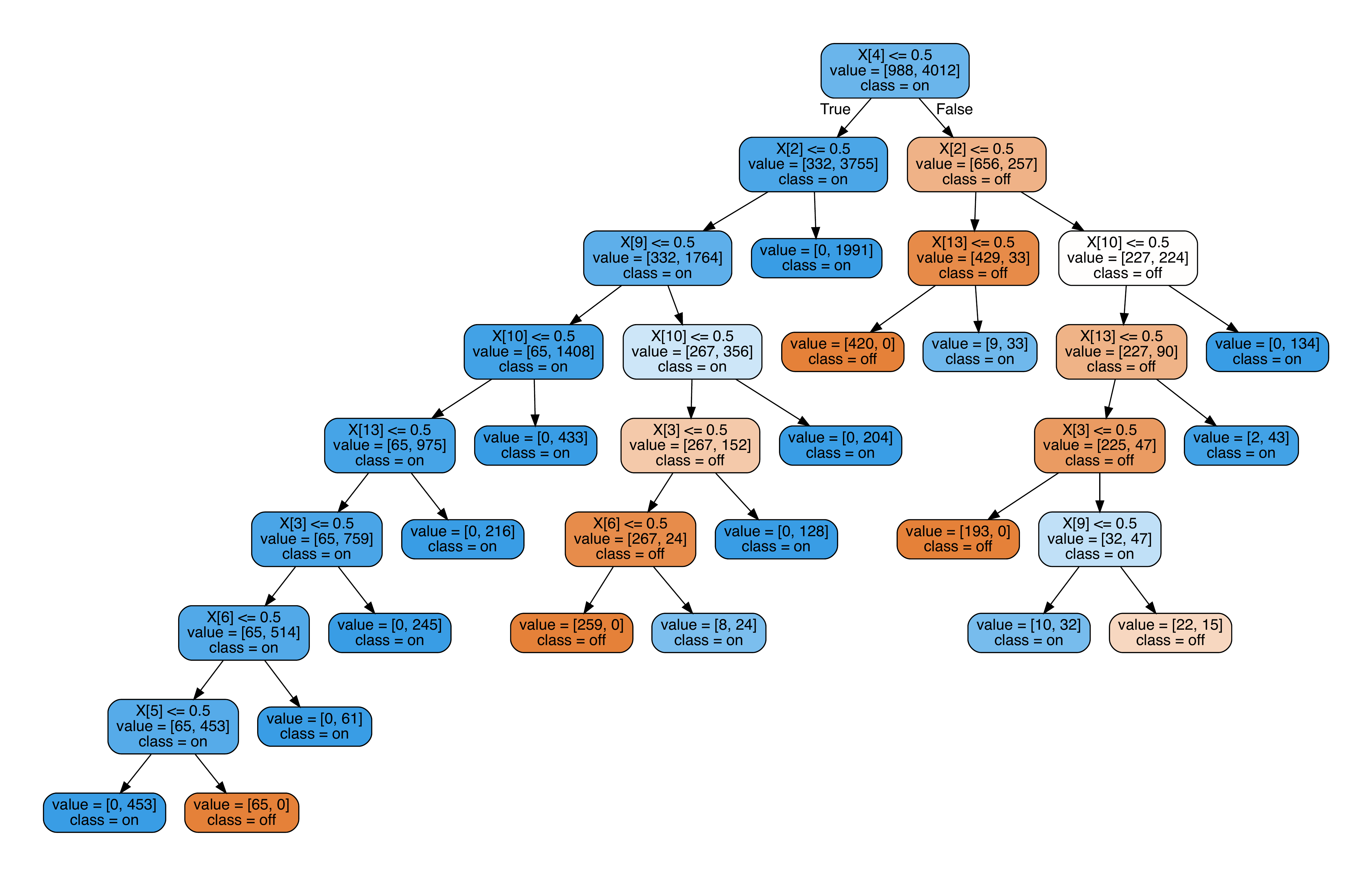}
        \caption{}
        \label{}
    \end{subfigure}
    \begin{subfigure}[b]{0.16\linewidth}
        \centering
        \includegraphics[width=\linewidth]{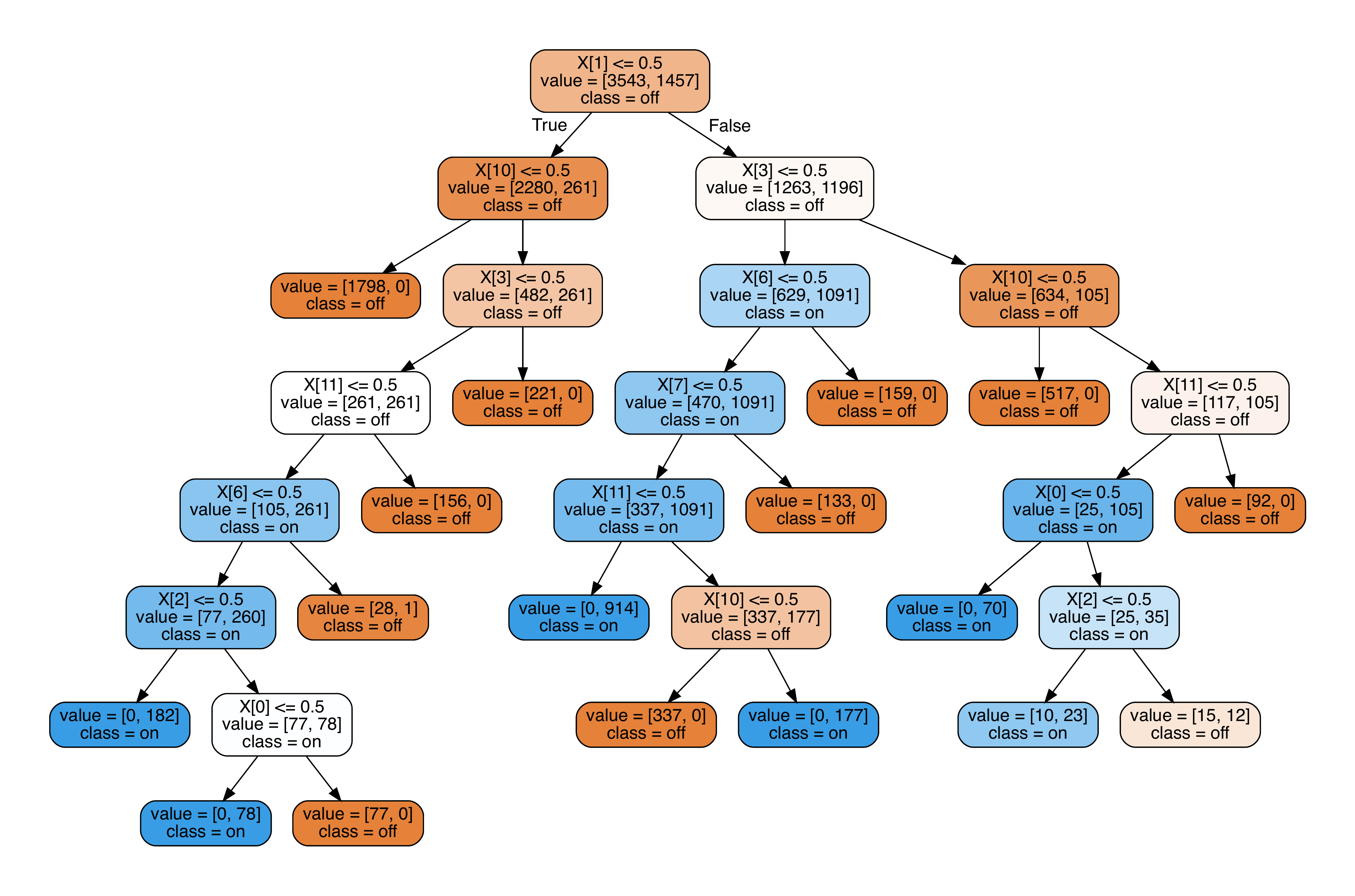}
        \caption{}
        \label{}
    \end{subfigure}
    \caption{\emph{(a-c)} Decision trees from 10 independent runs on the signal-and-noise HMM dataset with $\lambda = 1000.0$. Seven of the ten runs resulted in a tree of the same structure. The other three are similar, having additional subtrees but sharing the same splits and features. \emph{(d-f)} Similar experiment with $\lambda = 0.01$. Low regularization causes high variance in tree size and shape. Sub-figures (d-f) show three of many variations.}
    \label{fig:results:stable:test}
\end{figure}

\paragraph{Decision trees are stable over multiple optimization runs.}
When tree regularization is strong (high $\lambda$), the decision trees trained to match the predictions of deep models are stable. For both signal-and-noise and Sepsis tasks, multiple runs from different random restarts have nearly identical tree shape and size, perhaps differing by a few nodes. This stability is crucial to building trust in our method. On the signal-and-noise task ($\lambda = 7000$), 7 of 10 independent runs with random initializations resulted in trees of exactly the same structure, and the others closely resembled those sharing the same subtrees and features. On the other hand, with weak regularization (small $\lambda$), variability in the distilled decision trees is high. See Figure~\ref{fig:results:stable:test} example trees under strong (a-c) and weak (d-f) regularization.

\paragraph{Target neural models are faithful to decision trees.} \textit{Fidelity} is defined by \cite{craven1996extracting} as the percentage of examples where the prediction of the target network and the decision tree agree. Thus, fidelity is a measurement of how faithful the deep network is to the distilled tree. A fidelity of 1 would indicate perfect agreement, in which the neural network has learned exactly the axis-aligned boundaries of a tree. In some sense, a fidelity of 1 is undesirable as we hope the deep network can make use of nonlinearity on the examples that a simulable tree would struggle with. Table~\ref{table:fidelity} shows that the fidelity is high but not perfect, ranging from 0.80 to 0.94 across datasets.

\begin{figure}[h!]
    \begin{subfigure}[b]{0.24\linewidth}
        \centering
        \includegraphics[width=\linewidth]{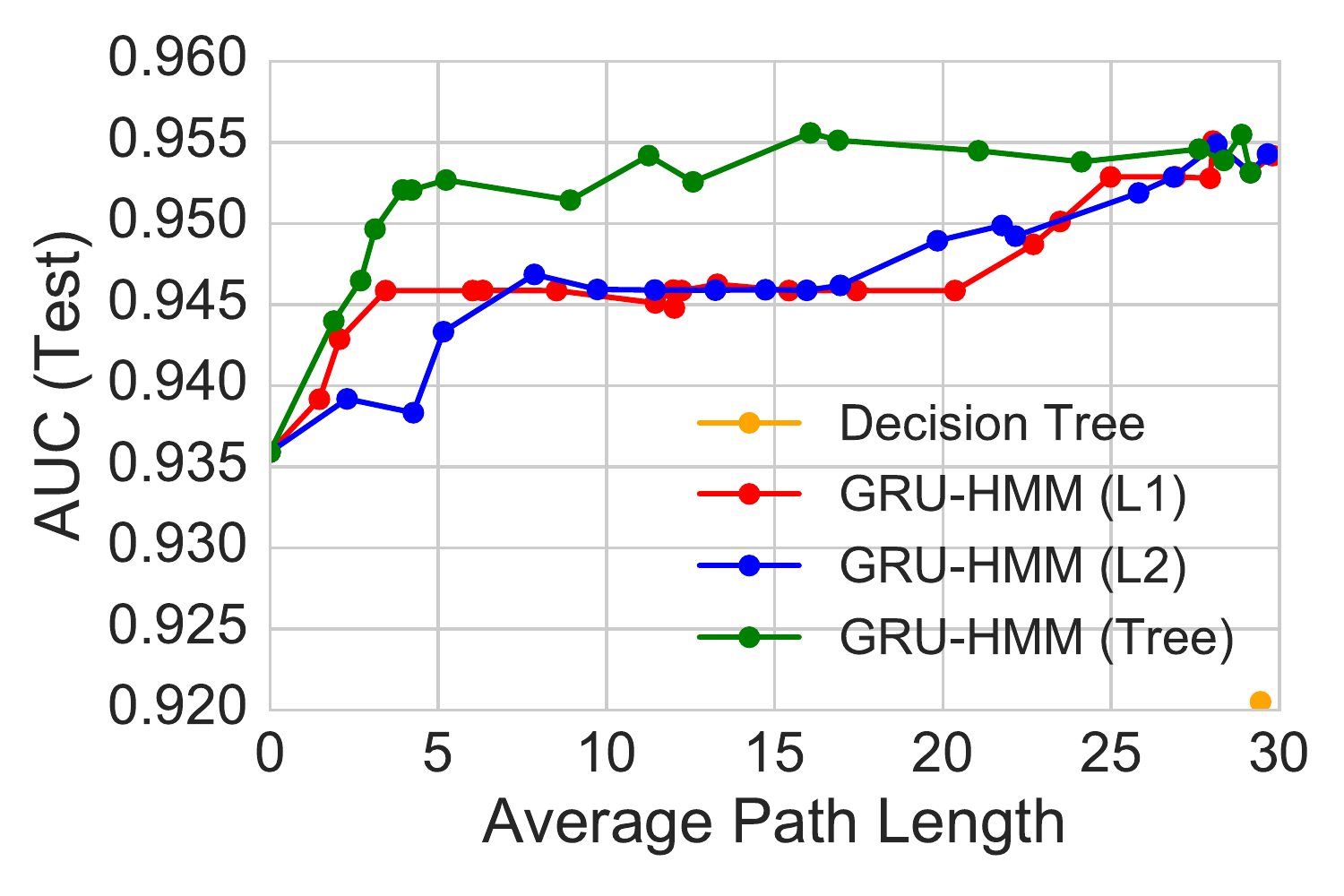}
        \caption{SNR {\tiny 20+5 states}}
        \label{fig:2hmm:gruhmm:plot}
    \end{subfigure}
    \begin{subfigure}[b]{0.24\linewidth}
        \centering
        \includegraphics[width=\linewidth]{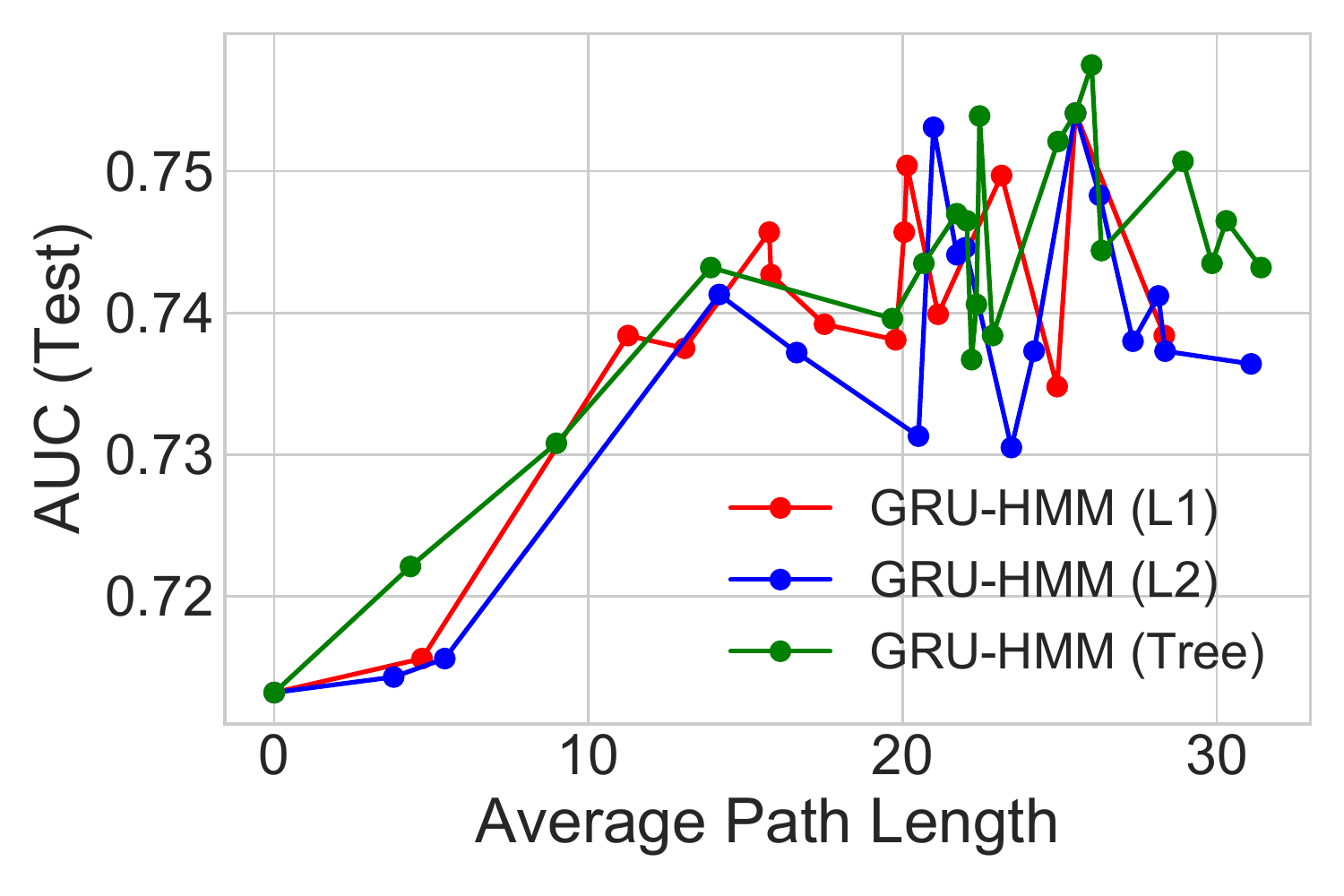}
        \caption{Mortality {\tiny 50+50}}
        \label{fig:sepsis:mortality:gruhmm:plot}
    \end{subfigure}
    \begin{subfigure}[b]{0.24\linewidth}
        \centering
        \includegraphics[width=\linewidth]{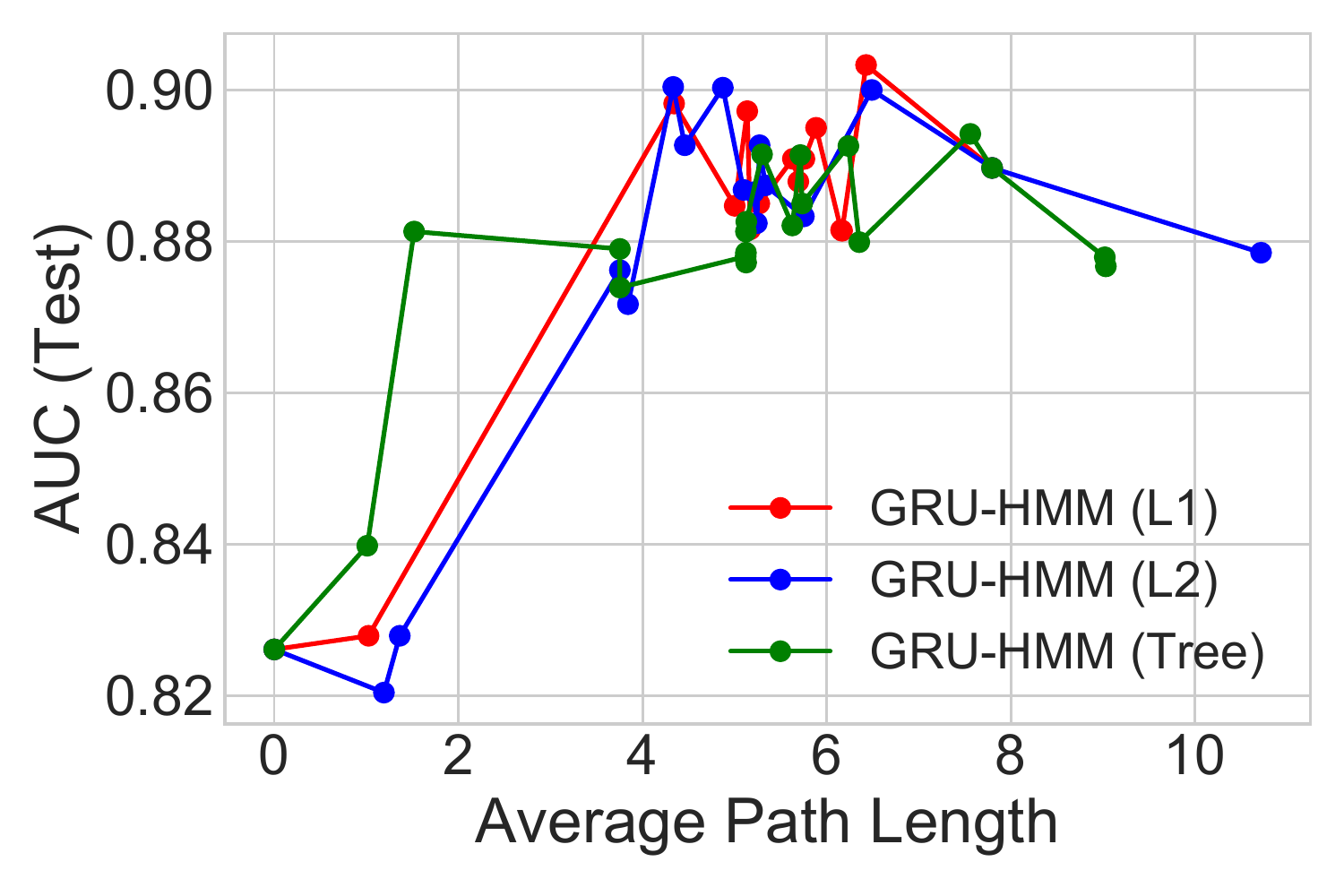}
        \caption{Mech. Vent. {\tiny 50+50}}
        \label{fig:sepsis:vent:gruhmm:plot}
    \end{subfigure}
    \begin{subfigure}[b]{0.24\linewidth}
        \centering
        \includegraphics[width=\linewidth]{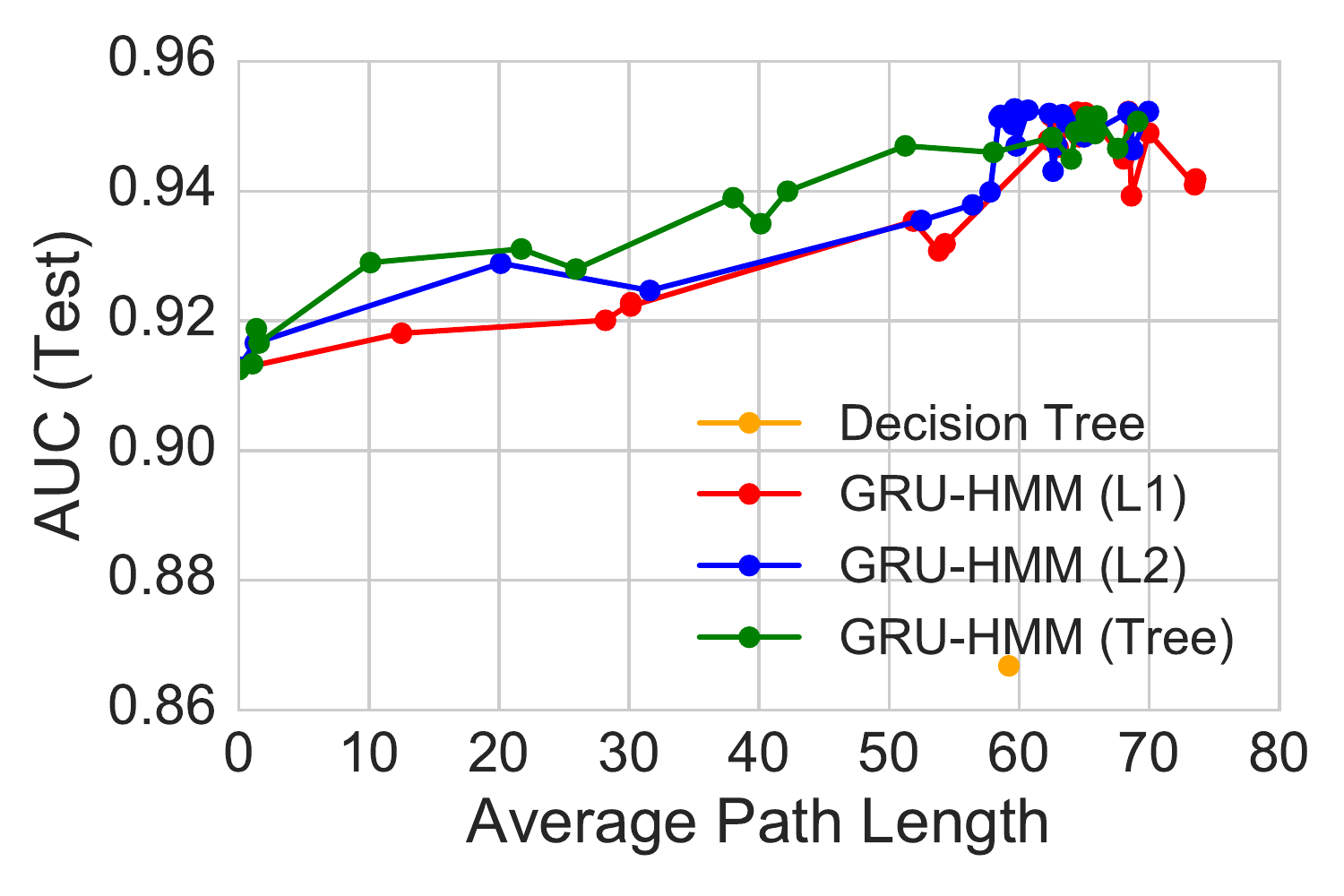}
        \caption{TIMIT ``Stop" {\tiny 50+25}}
        \label{fig:timit:gruhmm:plot}
    \end{subfigure}
    \caption{
    Fitness curves for the GRU-HMM, showing prediction quality (AUC) vs. complexity (APL) across range of regularization strengths $\lambda$. Captions show the number of HMM states plus the number of GRU states. See Figures~\ref{fig:results:sepsis} and \ref{fig:results:timit} to compare these GRU-HMM numbers to simpler GRU and decision tree baselines.
}
\label{fig:results:gruhmm}
\end{figure}

\paragraph{The deep residual GRU-HMM can achieve high AUC with less complexity.}

In Figure~\ref{fig:results:gruhmm}, we show the performance of jointly training the residual model, GRU-HMM, which combines an HMM with a tree-regularized GRU to improve its predictions. Here, the ideal APL is zero, indicating only the HMM makes predictions (only the GRU output node is regularized). For small APLs, the GRU-HMM substantially improves the original HMM's predictions \emph{and} has simulability gains over earlier GRUs. On the mechanical ventilation task, the GRU-HMM requires an APL of only 28 to reach AUC of 0.88, while the GRU alone with the same number of states requires a path length of 60 to reach the same AUC. This suggests that jointly-trained deep residual models may provide even better interpretability.

\section{Regionally Faithful Explanations with Expert Priors}
\label{sec:treereg}
Global summaries such as L$_1$, L$_2$, or even tree regularization as presented above face a tough trade-off between human-simulability and being faithful to the underlying model. For instance, if we require a minimum fidelity of 0.95, it simply may not be possible to fit a faithful  decision tree that is also human-simulable. In our experiments so far, we have been fortunate but there is little guarantee that such a tree must exist. More generally, for a complex enough domain (or for particularly difficult examples), it is again unreasonable to assume that there a decision tree can be small, bushy and performant. In such a case, tree regularization of a deep network may not be able to find a good compromise between accuracy and complexity. To get the best of both worlds, we will need a \textit{finer-grained} definition of interpretability. Doing so might help find a new wealth of minima with high AUC and low APL (aka powerful yet simulable).

In this extension, we take advantage of the fact that domain experts may already have notions about how regions of the input space operate differently.  For example, a clinical intensivist may already cognitively consider patients in the surgical intensive care unit (ICU) as different from patients in the cardiac ICU.  Analogously, biologists may be happy with different models for classifying diseases in deciduous versus in coniferous plants. In fact, this way of partitioning thinking into independent compartments is a very general phenomena. Cognitive science literature tells us that people build context-dependent models of the world; they do not expect the same rule to apply in all circumstances \cite{miller2018explanation}.

Using this intuition, we divide the input space into exclusive regions. We assume that this division is available \emph{a priori} via domain knowledge. In fact, this is a good opportunity to inject human beliefs into training the model. Formally, this translates into $R$ exclusive regions $\mathcal{X}_1, \ldots \mathcal{X}_R$, where $\cup_{r=1}^R \mathcal{X}_r \subseteq \mathcal{X}^P$. We denote the observed dataset belonging to region $r$ as $X_r \triangleq \{\textbf{x}_n : \textbf{x}_n \in \mathcal{X}_r\}$. Thus, we shall apply a \textit{regionally-faithful regularization} that encourages the target neural model to be ``simple'' in \emph{every} region (where a region corresponds to a human context). This partitioning  of the input space into regions allows a regularized neural model to approximate very complex decision boundaries with simple components (in each region) still, thereby remaining simulable.  We emphasize that our regional explanations are distinct from local explanations (e.g.~\cite{ribeiro2016should}): the latter concerns itself with behavior within an $\epsilon$-ball around a single data point, $\mathbf{x}_n$ and makes no claims about general behavior across data points. In contrast, \emph{regional} explanations are faithful over an entire region $\mathcal{X}_r$.
\begin{figure}[h]
  \centering
  \begin{subfigure}[b]{0.24\linewidth}
    \centering
    \includegraphics[width=0.7\linewidth]{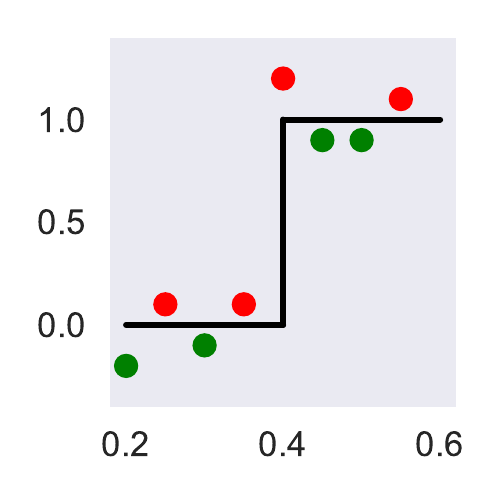}
    \caption{True}
  \end{subfigure}
  \begin{subfigure}[b]{0.24\linewidth}
    \centering
    \includegraphics[width=0.7\linewidth]{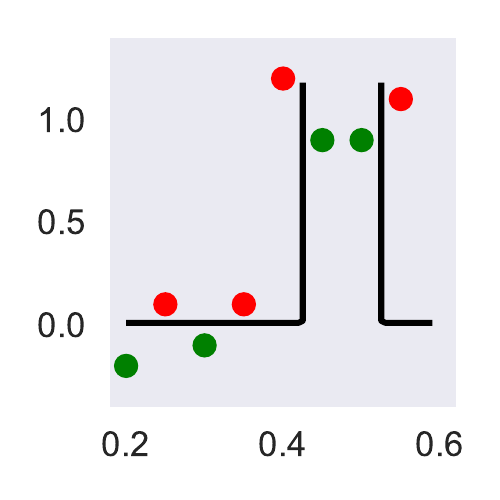}
    \caption{Global}
  \end{subfigure}
  \begin{subfigure}[b]{0.24\linewidth}
    \centering
    \includegraphics[width=0.7\linewidth]{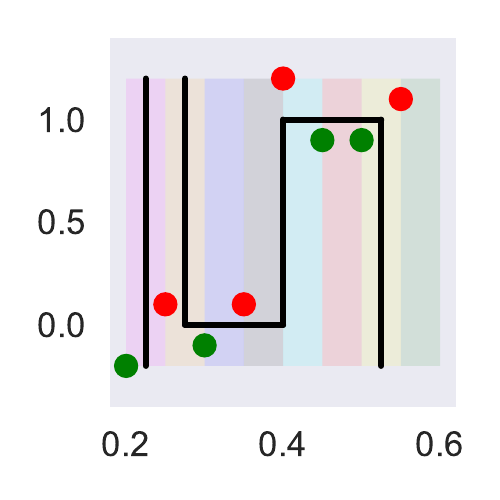}
    \caption{Local}
  \end{subfigure}
  \begin{subfigure}[b]{0.24\linewidth}
    \centering
    \includegraphics[width=0.7\linewidth]{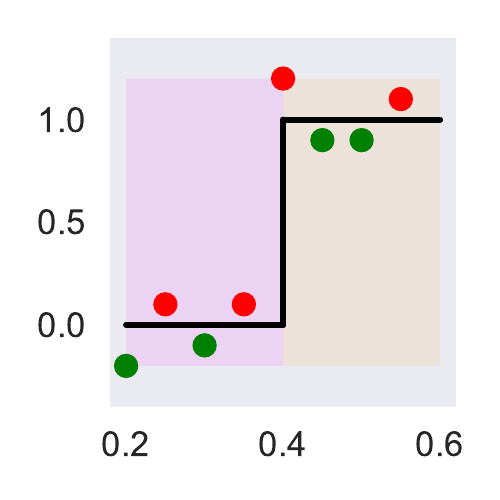}
    \caption{Regional}
  \end{subfigure}
  \caption{We show the differences between global (b), local (c), and
    regional (d) tree regularization using a synthetic classification
    task. (a) shows the true decision boundary. Red and green points
    represent the training dataset. Lightly colored areas represent
    regions. In (b), the model is over-regularized and ignores underlying structure.
    In (c), regions are made as small as possible to simulate
    locality---resulting in highly variable rules for nearby points.
    Regional tree regularization (d) provides an interpretable middle
    ground.}
  \label{fig:toy2}
\end{figure}
As a preview, Figure~\ref{fig:toy2} highlights the distinctions between global, local, and regional tree regularization on a two-dimensional toy dataset where the true decision boundary is divided in half at $x=0.4$. We see that global explanations (b) lack information about the input space and have to choose from a large set of possible solutions, converging to a different boundary. On the other hand, local explanations (c) produce simple boundaries around each data point but fail to capture global relationships, resulting in a complex overall decision function. Finally, regional explanations (d) over two regions divided at 0.4 share the benefits of (b) and (c), converging to the true boundary.

\begin{algorithm}[!t]
\caption{Pruned Average-Path-Length (APL) Cost Function}
\begin{algorithmic}[1]
\Require{
  \Statex $f(\cdot; \theta)$: discrete prediction function, with parameters $\theta$
  \Statex $\{ \mathbf{x}_i \}_{i=1}^{N}$: a set of $N$ input examples
  \Statex $N_{\text{train}}$: number of examples to use for training
  \Statex $h$: minimum number of samples required to be a leaf node
}
\Function{APL}{$\{ \mathbf{x}_i \}_{i=1}^{N}, f(\cdot; \theta), h$}
  \State $\hat{\mathbf{y}}_i = f(\mathbf{x}_{i}, \theta)$, $\forall i \in \{1, 2, \ldots N\}$
  \State $T = \textsc{TrainTree}( \{ \mathbf{x}_{i}, \hat{\mathbf{y}}_i \}_{i=1}^{N_{\text{train}}})$
  \State $T = \textsc{PruneTree}(T, \{ \mathbf{x}_{i}, \hat{\mathbf{y}}_i \}_{i=N_{\text{train}}}^N)$
  \State \Return $\text{mean}( \{ \textsc{GetDepth}(T, \mathbf{x}_i) \}_{i=1}^N )$
\EndFunction
\end{algorithmic}
\label{algorithm:2}
\end{algorithm}

\subsection{Regional Tree Regularization Objective}
\begin{figure}[t]
  \centering
  \begin{subfigure}[b]{0.19\linewidth}
    \includegraphics[width=0.8\linewidth]{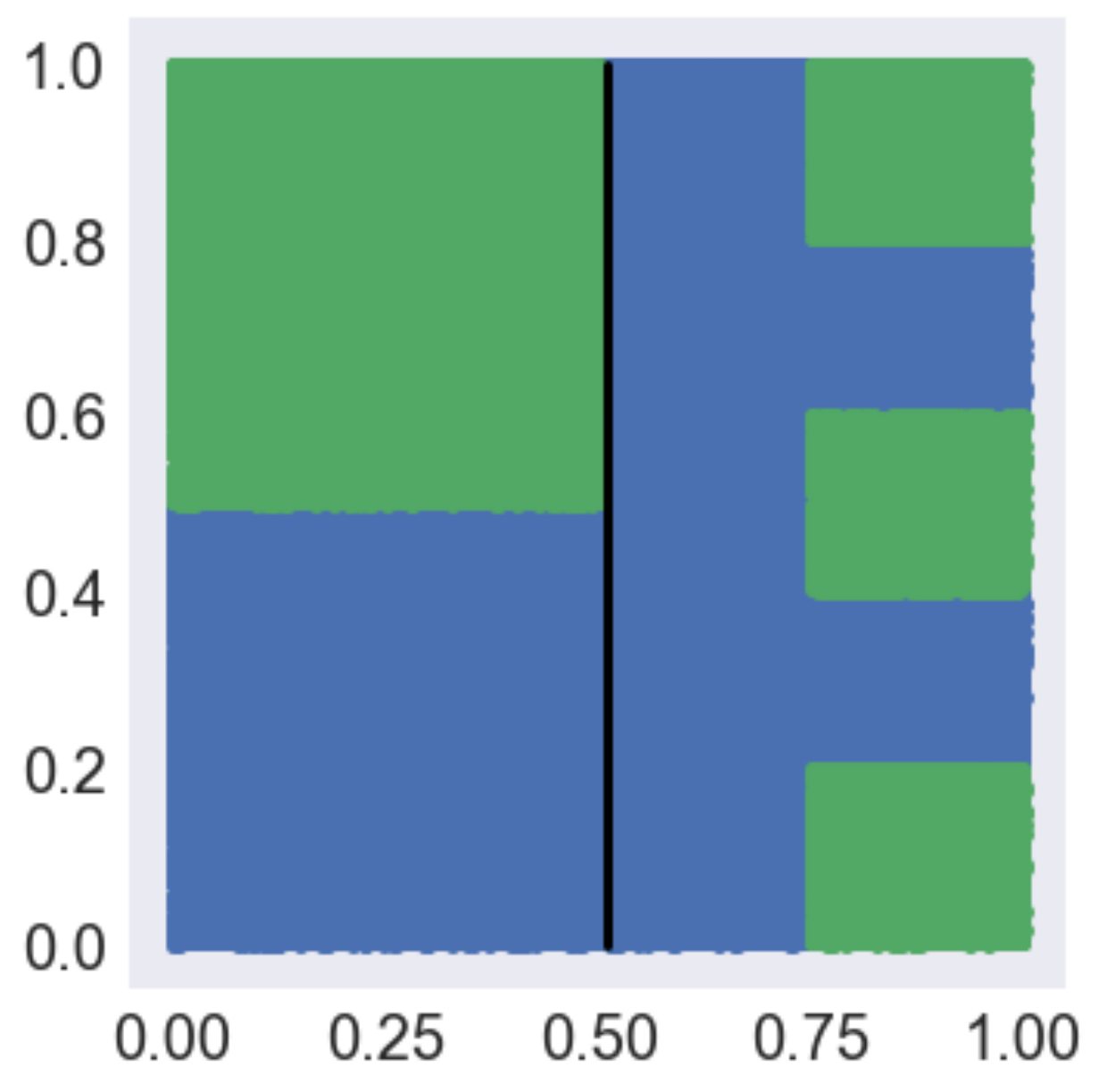}
    \caption{True}
  \end{subfigure}
  \begin{subfigure}[b]{0.19\linewidth}
    \includegraphics[width=0.8\linewidth]{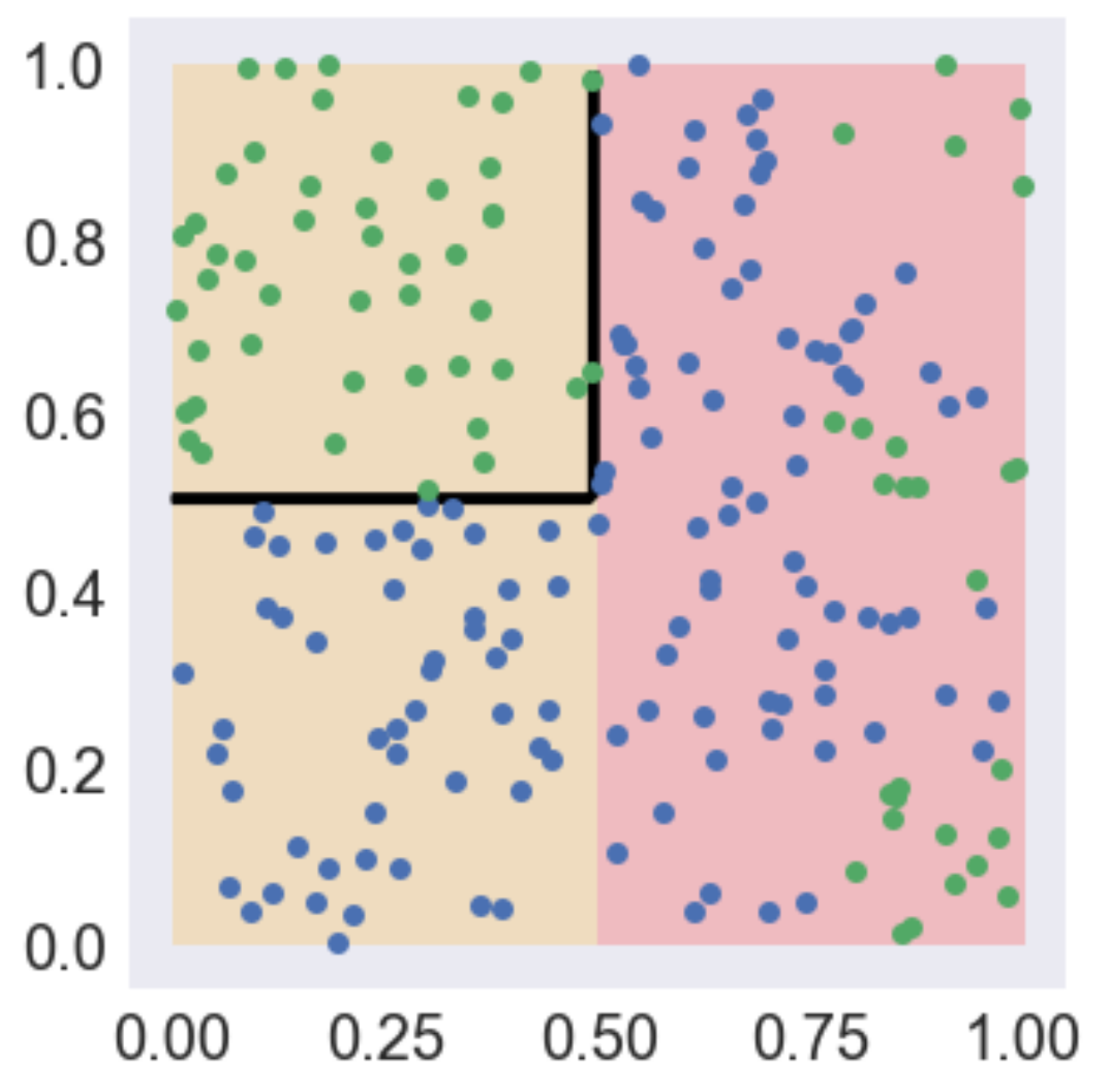}
    \caption{L$_1$}
  \end{subfigure}
  \begin{subfigure}[b]{0.19\linewidth}
    \includegraphics[width=0.8\linewidth]{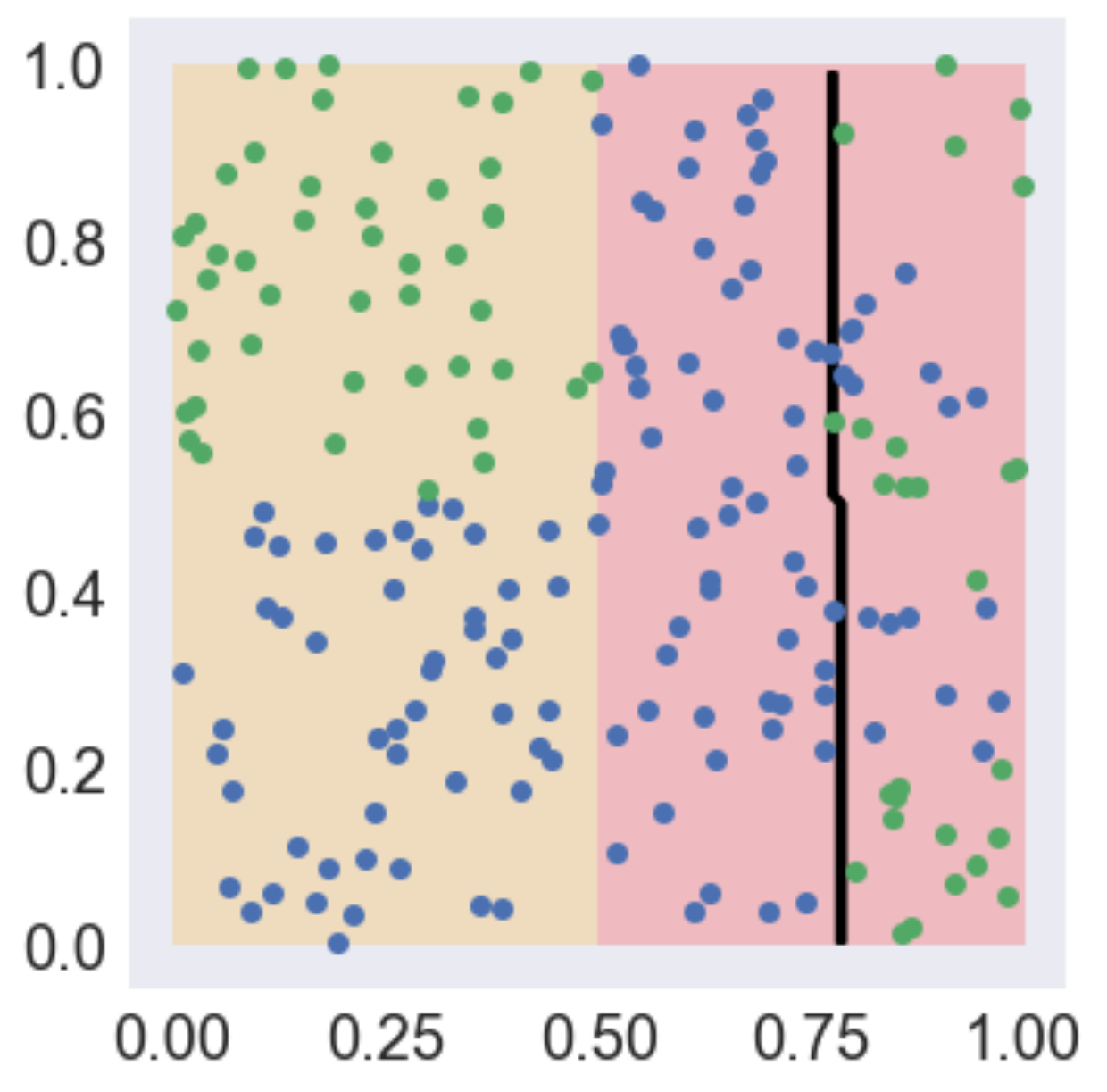}
    \caption{L$_1$}
  \end{subfigure}
  \begin{subfigure}[b]{0.19\linewidth}
    \includegraphics[width=0.8\linewidth]{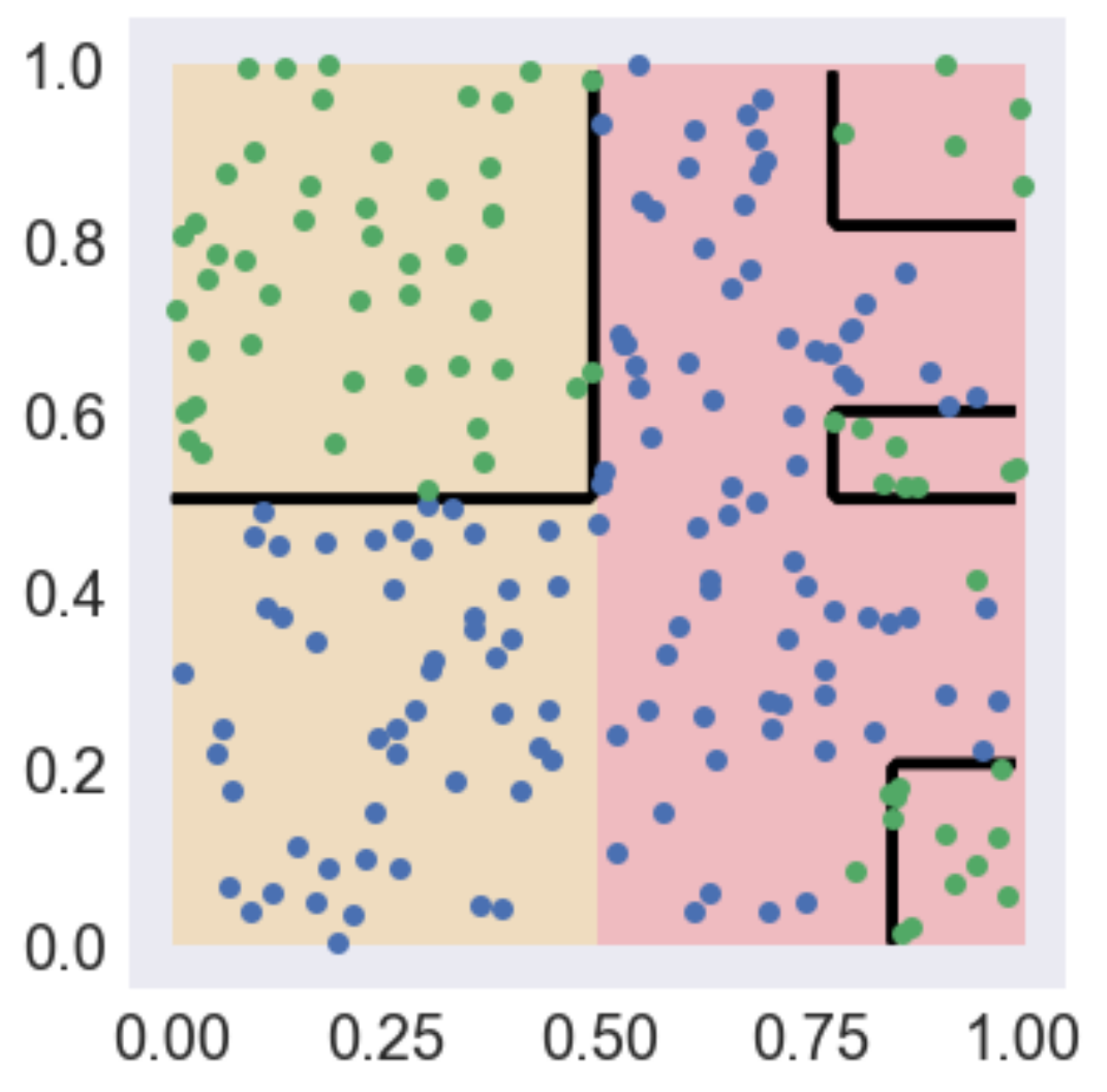}
    \caption{L$_0$}
  \end{subfigure}
  \begin{subfigure}[b]{0.19\linewidth}
    \includegraphics[width=0.8\linewidth]{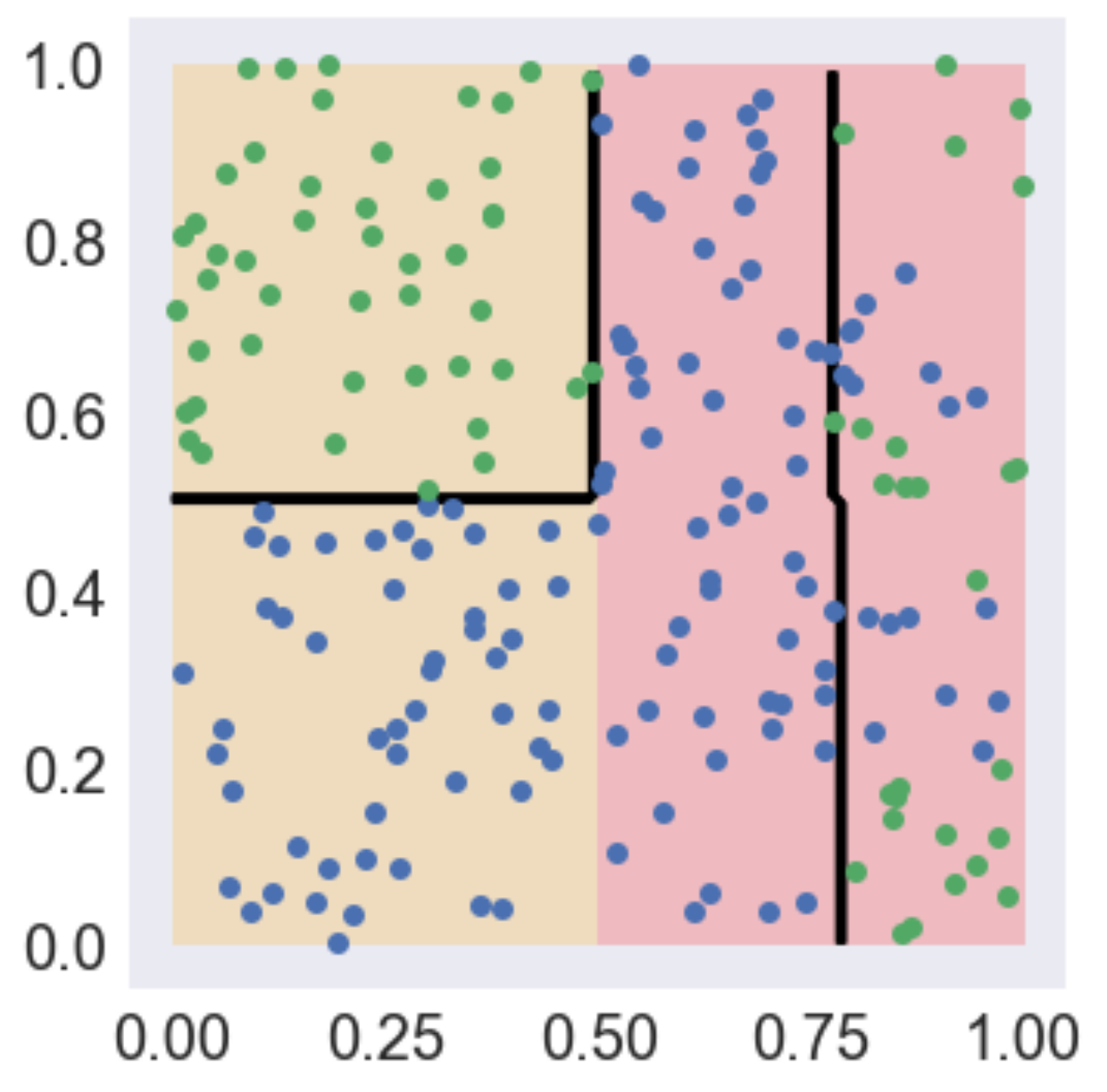}
    \caption{L$_0$}
  \end{subfigure}
  \caption{An L$_1$ penalty on per-region APLs can over-penalize, resulting in an entire region with far too simple predictions. Subplots (b) and (c) show results from two different initializations using the L$_1$ norm, while (d) and (e) show the same using the L$_0$ norm.}
  \label{fig:toyexp3}
\end{figure}

We now formally introduce regional tree regularization, which will require that the target neural model $f(\cdot; \theta)$ is well-approximated by a separate compact decision tree in \emph{every} region. In contrast, we will rename the tree regularizer presented above as \emph{global tree regularization}. Regionally simple decision boundaries are particularly hard to achieve with global tree regularization as the global APL metric may allow some human-relevant regions to be complex as long as most are simple. In particular, global tree regularization has an incentive to ``ignore" simpler regions in order to minimize the regularization term (i.e. trivially prediction a single label). In many contexts, this behavior is undesirable. For example, if a clinician splits his/her patients by severity of illness, regularizing for simple global explanations can completely ignore a group of patients, rendering the machine learning system useless. To address this, we define our regional tree regularization as follows. First, let the APL for region $r$ be:
\begin{align}
  \Omega^{\texttt{regional}}_{r}(\theta)
  &\triangleq \textup{APL}(\mathcal{X}_r, f(\cdot; \theta))
  \\
  \Omega^{\texttt{global}}(\theta)
  &\triangleq \textup{APL}(\mathcal{X}^P, f(\cdot; \theta))
\label{eqn:region-apl}
\end{align}
where the average path length, $\textup{APL}$ can be computed with Algorithm~\ref{algorithm:2} (note that the target network and its parameters $\theta$ are the same for all regions $r$, meaning a strong sharing of parameters across regions). For all future instances of computing APL, we use Algorithm~\ref{algorithm:2}, not Algorithm~\ref{alg:true_tree_regularization}. We will elaborate on this distinction later. Note that $\Omega^{\texttt{global}}(\theta)$ is equivalent to global tree regularization as presented above. Next, to ensure that some regions cannot be made simple at the expense of others, we penalize only the most complex region:
\begin{equation}
  \Omega^{\texttt{regional}}(\theta)
  \triangleq \texttt{max}_r (\{ \Omega^{\texttt{regional}}_{r}(\theta) \}_{r=1}^R)
\label{eqn:argmax-apl}
\end{equation}
in other words, a L$_0$ norm over $\{\Omega_r\}$.  The choice of L$_0$ norm produces significantly different (and desirable) behavior than if we had simply used, for example, the L$_1$ norm (or sum) over $\{\Omega_r\}$.  Regularizing the sum of $\Omega_r$ is equivalent to simply regularizing APL in a global tree that first branches by region.  In contrast, as a nonlinear regularizer, L$_0$ keeps \emph{all} regions simple (aka low APL), while not penalizing regions that are already simple. We show an example of this effect in Figure~\ref{fig:toyexp3}: (a) shows a toy dataset with two regions (split by the black line): the left has a simple decision boundary dividing the region in half; the right has a more complex boundary. (b) and (c) then show two minima using L$_1$ regional tree regularization.  In both cases, one of the regions collapses to a trivial decision boundary (predicting all one label) to minimize the overall sum of APLs. On the other hand, since L$_0$ is sparse, simple regions are not included in the objective, resulting in a more ``balanced" regularization between regions (see d and e).

However, gradient descent with Equation~\ref{eqn:argmax-apl} has several challenges. For example, both $\Omega_r$ and the $\max$ functions are non-differentiable.  In the following, we describe how we address these challenges as well as concerns over optimization stability.

\begin{algorithm}[b]
\caption{\textsc{SparseMax For Regional Tree Reg.}}
\begin{algorithmic}[1]
\Require{
  \Statex $\mathbf{\hat{\Omega}} = \{\hat{\Omega}^{\texttt{regional}}_r\}_{r=1}^R$: APL for each of $R$ regions
}
\Function{$\textsc{SparseMax}$}{$\mathbf{\hat{\Omega}}$}
   \State Sort $\mathbf{\hat{\Omega}}$ such that $\mathbf{\hat{\Omega}}[i] \geq \mathbf{\hat{\Omega}}[j]$ if $i \geq j$
   \State $k = \max \{ r \in [1, R] | (1 + r\mathbf{\hat{\Omega}}[r]) > \sum_{i \leq r} \mathbf{\hat{\Omega}}[i] \}$
   \State $\tau = k^{-1}(-1 + \sum_{i \leq k} \mathbf{\hat{\Omega}}[i])$
   \State $\mathbf{p} = \{p_r\}_{r=1}^R$ where $p_r = \max\{\mathbf{\hat{\Omega}}_r - \tau, 0 \}$
   \State \Return $\mathbf{p}$
\EndFunction
\end{algorithmic}
\label{algorithm:3}
\end{algorithm}

\subsection{Gradient-based optimization with SparseMax}
Gradient-based optimization of our proposed regularizer in Equation~\ref{eqn:argmax-apl} is challenging because the $\texttt{max}$ operator is not differentiable.  Further, common differentiable approximations like $\texttt{softmax}$ are dense (include non-zero contributions from all regions), which makes it difficult to focus on the most complex regions as $\texttt{max}$ does (using a dense approximation of $\texttt{max}$ would suffer from the same problems as using a L$_1$ norm). Instead, we use the recently-proposed $\textsc{SparseMax}$ transformation~\cite{martins2016softmax},
which can focus on the most problematic regions (setting others to zero contribution) while remaining smooth and differentiable almost everywhere. Intuitively, $\textsc{SparseMax}$ corresponds to a Euclidean projection of an input vector $\mathbf{\hat{\Omega}}$ with $R$ entries (one APL per region) to an $R$-length vector $\mathbf{p}$ of non-negative entries that sums to one (i.e. the ($R-1$)-dimensional probability simplex).
When the projection lands on a boundary in the simplex (which is likely), then the resulting vector will be sparse.
Efficient implementations of this projection are well-known~\cite{duchi2008efficient} (see Algorithm~\ref{algorithm:3}), as are Jacobians for automatic differentiation~\cite{martins2016softmax}.
We refer to using $\textsc{SparseMax}$ as L$_0$ regional tree regularization (we call using the sum of the APLs L$_1$ regional tree regularization).

\begin{figure*}[t!]
    \centering
    \includegraphics[width=\textwidth]{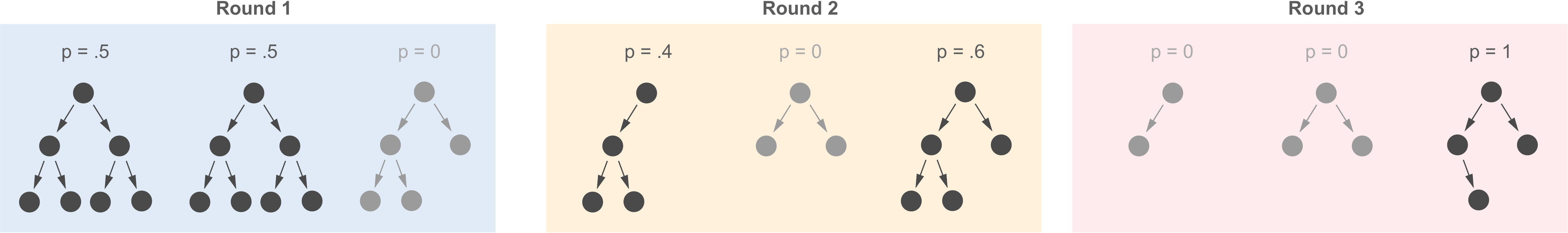}
    \caption{Illustratoin of L$_0$ regional tree regularization. Each round contains three trees representing regions. Light gray color indicates regions given 0 probability by \texttt{sparsemax}. Over the three rounds, different regions are given priority while other regions are given no weight. The ability to disregard regions of low complexity makes for smoother learning.}
\end{figure*}

\subsection{Differentiable Regional Tree Regularization Loss $\hat{\Omega}_r$}
The regional APL $\Omega^{\texttt{regional}}_r(\theta)$ is not differentiable as derivatives cannot flow through CART (the common method for training decision trees). To circumvent this, we again employ surrogate loss functions $\hat{\Omega}^{\texttt{regional}}_{r}:$ that map a parameter vector $\theta \in \Theta$ to an \textit{estimate} of $\Omega^{\texttt{regional}}_{r}(\cdot)$, the APL in region $r$. This process is identical to global tree regularization but only for observations lying in region $r$. Each surrogate $\hat{\Omega}^{\texttt{regional}}_r$ has its own parameters $\phi_r$. Specifically, we fit each $\hat{\Omega}^{\texttt{regional}}_{r}(\theta)$ by minimizing a mean squared error loss,
\begin{equation}
  \min_{\phi_r} \sum_{j=1}^{J} (\Omega^{\texttt{regional}}_{r}(\theta_{j}) - \hat{\Omega}^{\texttt{regional}}_{r}(\theta_{j}, \phi_r))^{2}
  \label{eqn:opt:local}
\end{equation}
for all $r=1, ..., R$ where $\theta_j$ is sampled from a dataset of $J$ known parameter vectors and their true APLs: $\mathcal{D}^{\theta}_{r} = \{\theta_j, \Omega^{\texttt{regional}}_{r}(\theta_{j})\}_{j=1}^{J}$. This dataset can be assembled using the candidate $\theta$ vectors obtained over $J$ gradient steps while training the target model $f(\cdot, \theta)$. For $R$ regions, we curate one such dataset for each surrogate model.

The ability of each surrogate to stay faithful is a function of many factors. For global tree regularization (above), we used a fairly simple strategy for training a surrogate and found it sufficient; we find that especially when there are multiple surrogates to be maintained, sophistication is needed to keep the gradients accurate and the variances low. We describe these innovations in the next section.

\subsection{Innovations for Optimization Stability}
\label{sec:global:improve}

\begin{figure}[t]
  \centering
  \begin{subfigure}[b]{0.16\linewidth}
    \includegraphics[width=\linewidth]{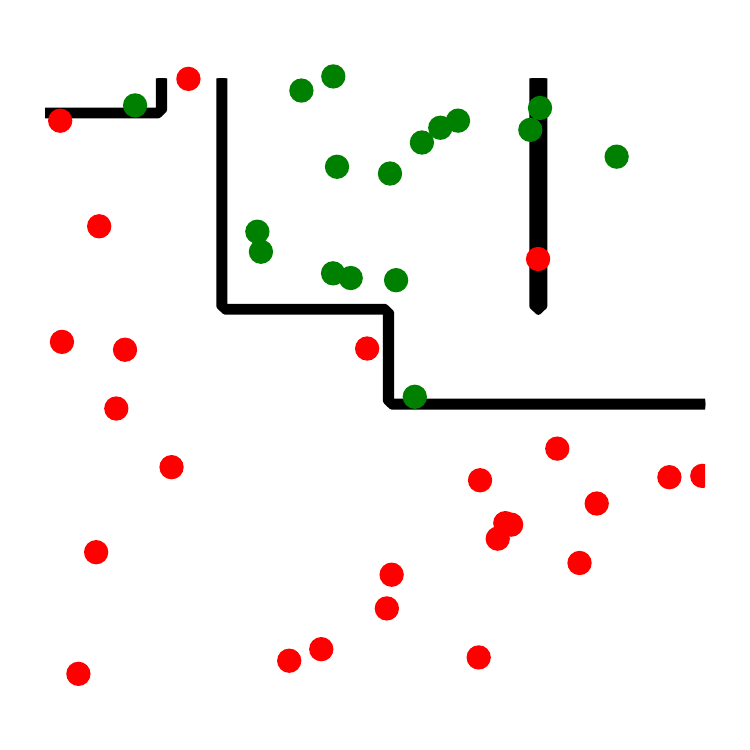}
    \caption{Random:1}
  \end{subfigure}
  \begin{subfigure}[b]{0.16\linewidth}
    \includegraphics[width=\linewidth]{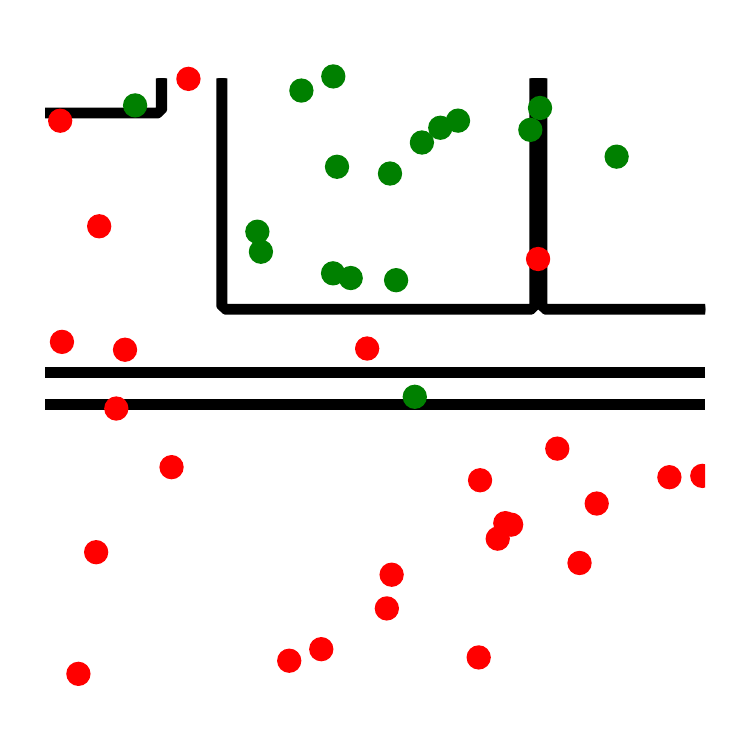}
    \caption{Random:2}
  \end{subfigure}
  \begin{subfigure}[b]{0.16\linewidth}
    \includegraphics[width=\linewidth]{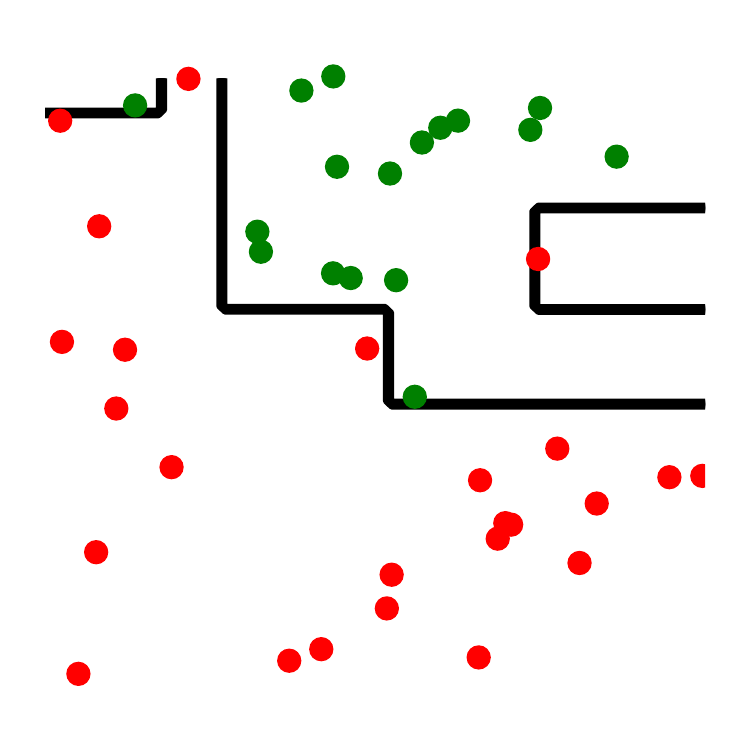}
    \caption{Random:3}
  \end{subfigure}
  \begin{subfigure}[b]{0.16\linewidth}
    \includegraphics[width=\linewidth]{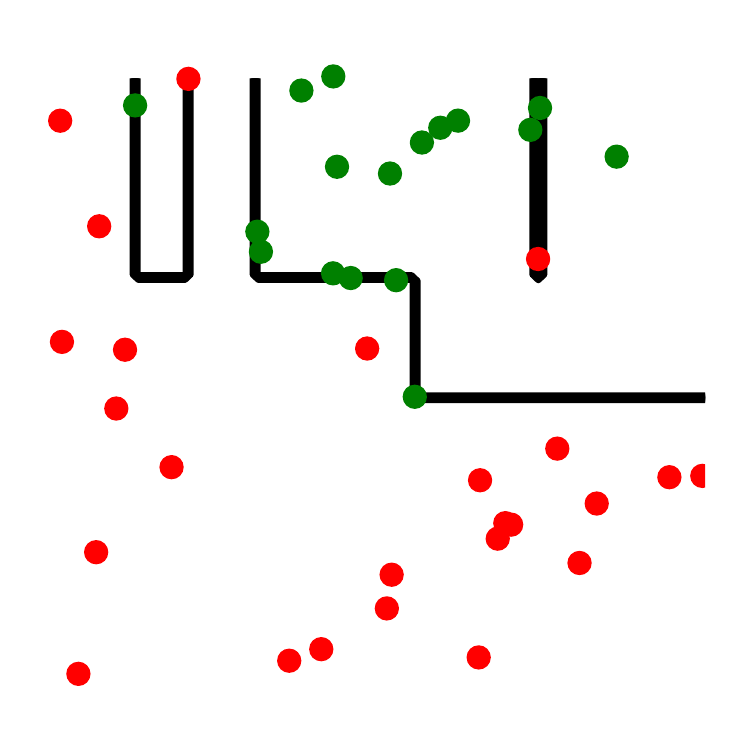}
    \caption{Fixed:1}
  \end{subfigure}
  \begin{subfigure}[b]{0.16\linewidth}
    \includegraphics[width=\linewidth]{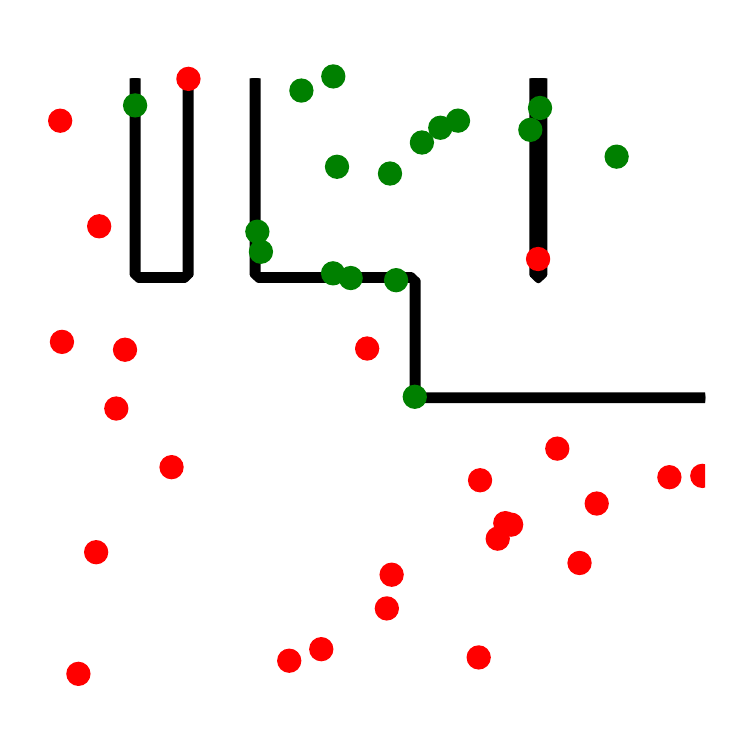}
    \caption{Fixed:2}
  \end{subfigure}
  \begin{subfigure}[b]{0.16\linewidth}
    \includegraphics[width=\linewidth]{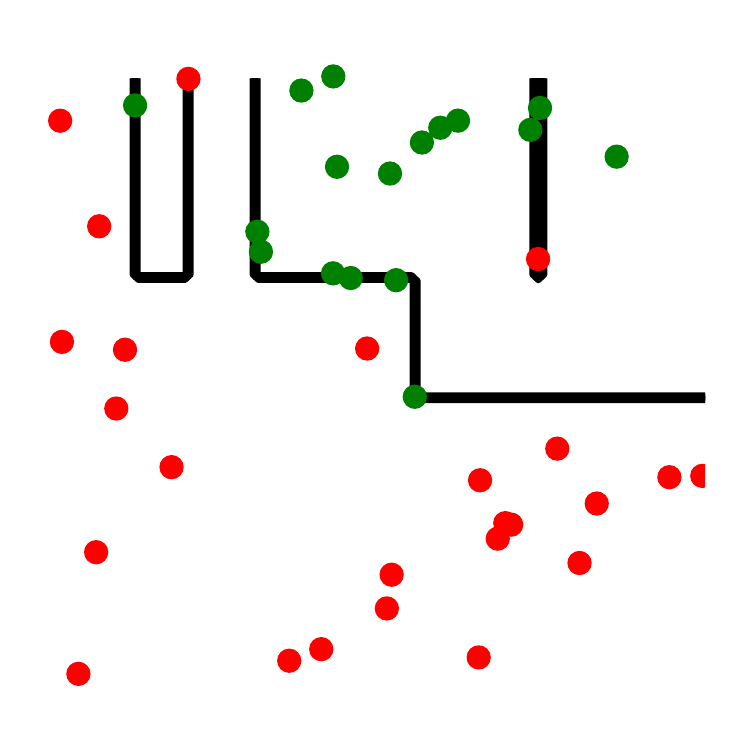}
    \caption{Fixed:3}
  \end{subfigure}
  \caption{\emph{(a-d)} Decision trees using randomized training; \emph{(e-h)} Decision trees using deterministic training. Note that randomized training leads to very different optima.}
  \label{fig:deterministic}
\end{figure}

Optimizing multiple surrogate networks is a delicate operation. We found that depending on hyperparameters, the regional surrogates were unable to accurately predict the APL, causing regularization to fail. Further, repeated runs also often found different minima, making regional tree regularization feel unreliable. In short, it presents a much more difficult technical challenge than training a single surrogate as in global tree regularization. Below, we list optimization innovations that are essential to stabilize training, identify consistent minima, and get good APL prediction---all of which enabled robust regional tree regularization.

\begin{wrapfigure}{r}{0.5\textwidth}
\centering
\begin{tabular}{l c c}
\toprule
Experiment & Mean MSE & Max MSE \\
\midrule
No data aug. & 0.069 & 0.987 \\
With data aug. & 0.015 & 0.298 \\
\hline
Randomized & 0.116 & 1.731 \\
Deterministic & 0.024 & 0.371 \\
\bottomrule
\end{tabular}
\caption{Comparison of the average and max mean squared error (MSE) between surrogate predictions and true average path lengths over 500 epochs. Non-deterministic training and lack of data introduces large errors.}
\label{table:tricks}
\end{wrapfigure}

\paragraph{Data augmentation makes for a robust surrogate.}
Especially for regional explanations, relatively small changes in the underlying model can mean large changes for the pattern in a specific region.  As such, the surrogates need to be retrained frequently (e.g. every 50 gradient steps).  The practice used in global tree regularization of computing the true APL for a dataset $\mathcal{D}^\theta$ of the most recent $\theta$ is insufficient to learn the mapping from a thousand-dimensional weight vector to the APL.  Using stale (very old) $\theta$ from previous epochs, however, would result in a poor surrogate model given outdated information. Previous heuristics as in random restarts or arbitrarily sampling random weights introduced more noise than signal. Thus, we supplement the dataset with randomly sampled weight vectors \textit{from the convex hull defined by the recent weights}.  Specifically, to generate a new $\theta$, we sample from a Dirichlet distribution with $J$ categories and form a new parameter as a convex combination of the elements in $\mathcal{D}^{\theta}$. For each of these samples, we compute its true APL to train the surrogate.  Table~\ref{table:tricks} shows this to reduce noise.

\paragraph{Decision trees should be pruned.}
Given a dataset, $\mathcal{D}$, even with a fixed seed, there are many decision trees that can fit $\mathcal{D}$. One can always add additional subtrees that predict the same label as the parent node, thereby not effecting performance. This invariance again introduces difficulty in learning a surrogate model. To remedy this, we use \textit{reduced error pruning}, which removes any subtree that does not effect performance as measured on a portion of $\mathcal{D}$ not used in $\textsc{TrainTree}$. Note that line 4 in Algorithm~\ref{algorithm:2} is not in the original tree regularization algorithm. Intuitively, pruning collapses the set of possible trees describing a single classifier to a singleton.

\paragraph{Decision trees should be trained deterministically.}
CART is a common algorithm to train a decision tree. However, it has poor complexity in the number of features as it enumerates over all unique values per dimension. To scale efficiently, many open-source implementations (e.g. Scikit-Learn \cite{pedregosa2011scikit}) randomly sample a small subset of features. As such, independent training instances can lead to different decision trees of varying APL. For tree regularization, unexplained variance in APL means difficulty in training the surrogate model, since the function from model parameters to APL is no longer many-to-one.  The error is compounded when there are many surrogates.  To remedy this, we fix the random seed that governs the choice of features.  As an example, Figure~\ref{fig:deterministic} shows the high variance of decision boundaries from a randomized treatment of fitting decision trees (a-d) on a very sparsely sampled data set, leading to higher error in surrogate predictions (Table~\ref{table:tricks}).  Setting the seed removes this variance.

\paragraph{A large learning rate will lead to thrashing.}
As mentioned before, with many regions, small changes in the deep model can already have large effects on a region.  If the learning rate is fast, each gradient step can lead to a dramatically different decision boundary than the previous. Thus, the function that each surrogate must learn is no longer continuous. Empirically, we found large learning rates to lead to \textit{thrashing}, or oscillating between high and low APL where the surrogate is effectively memorizing the APL from the last epoch (with poor generalization to new $\theta$).

\begin{figure}[h!]
  \centering
  \begin{subfigure}[b]{0.15\linewidth}
    \includegraphics[width=\linewidth]{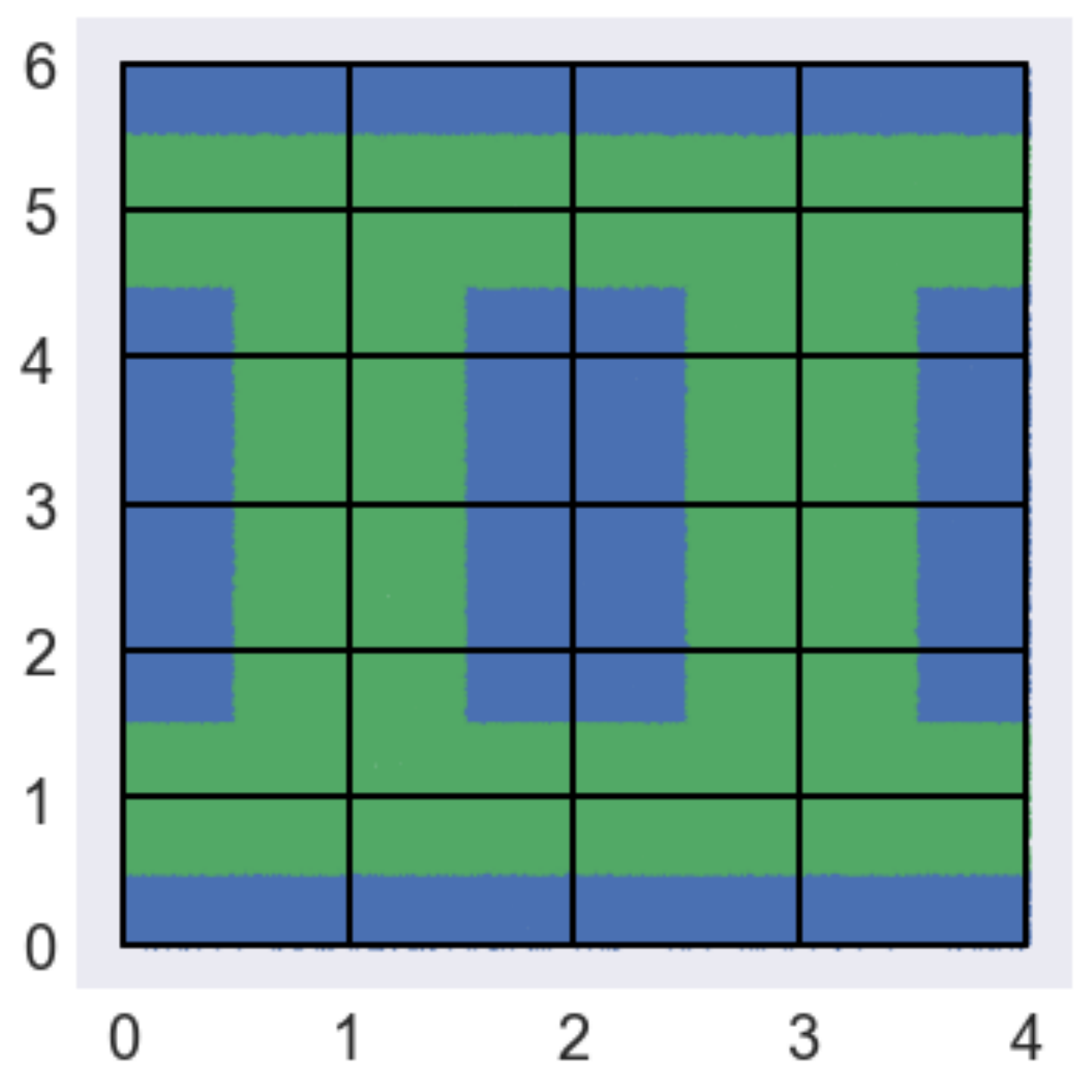}
    \caption{}
  \end{subfigure}
  \begin{subfigure}[b]{0.15\linewidth}
    \includegraphics[width=\linewidth]{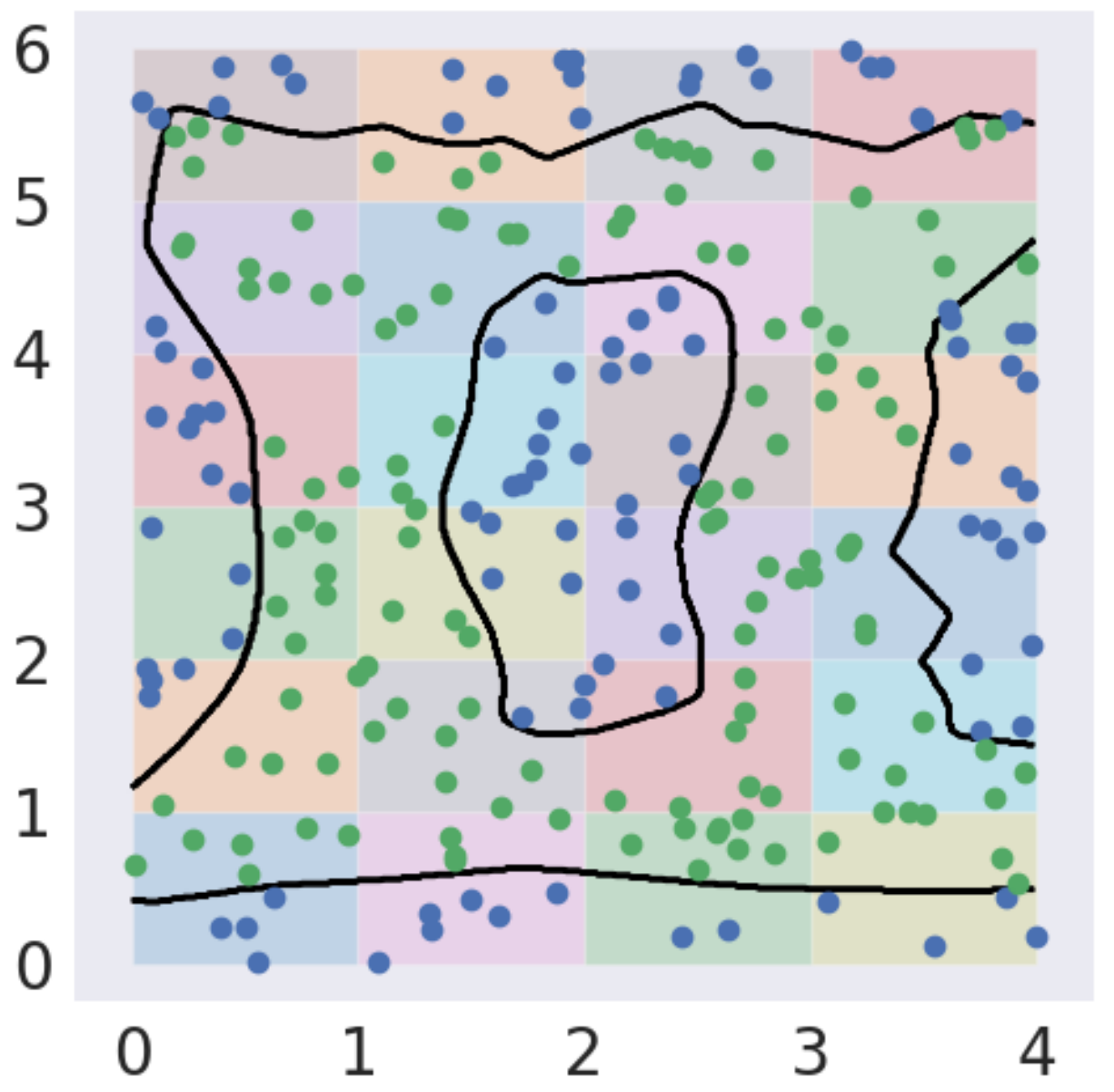}
    \caption{}
  \end{subfigure}
  \begin{subfigure}[b]{0.15\linewidth}
    \includegraphics[width=\linewidth]{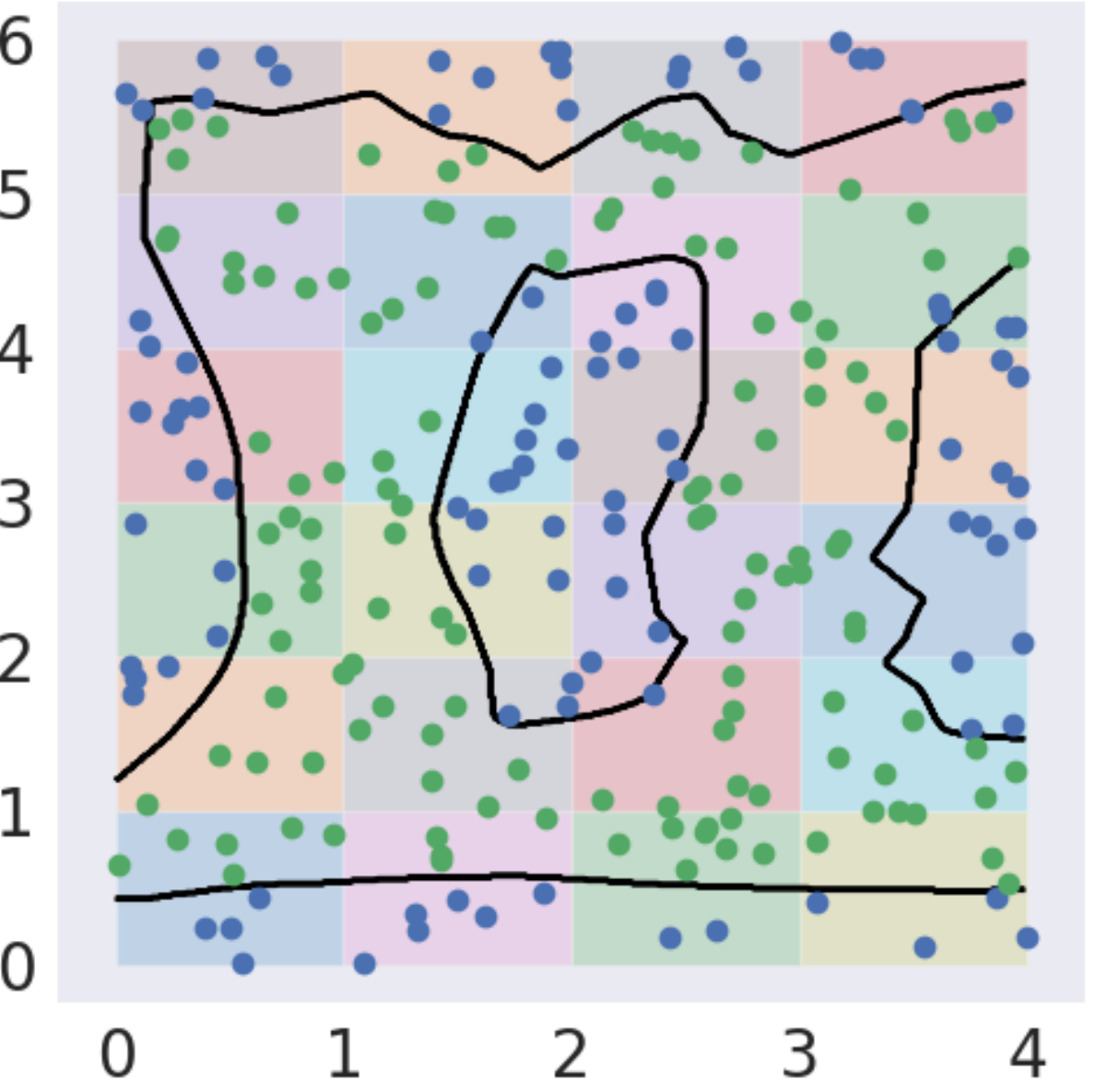}
    \caption{}
  \end{subfigure}
  \begin{subfigure}[b]{0.15\linewidth}
    \includegraphics[width=\linewidth]{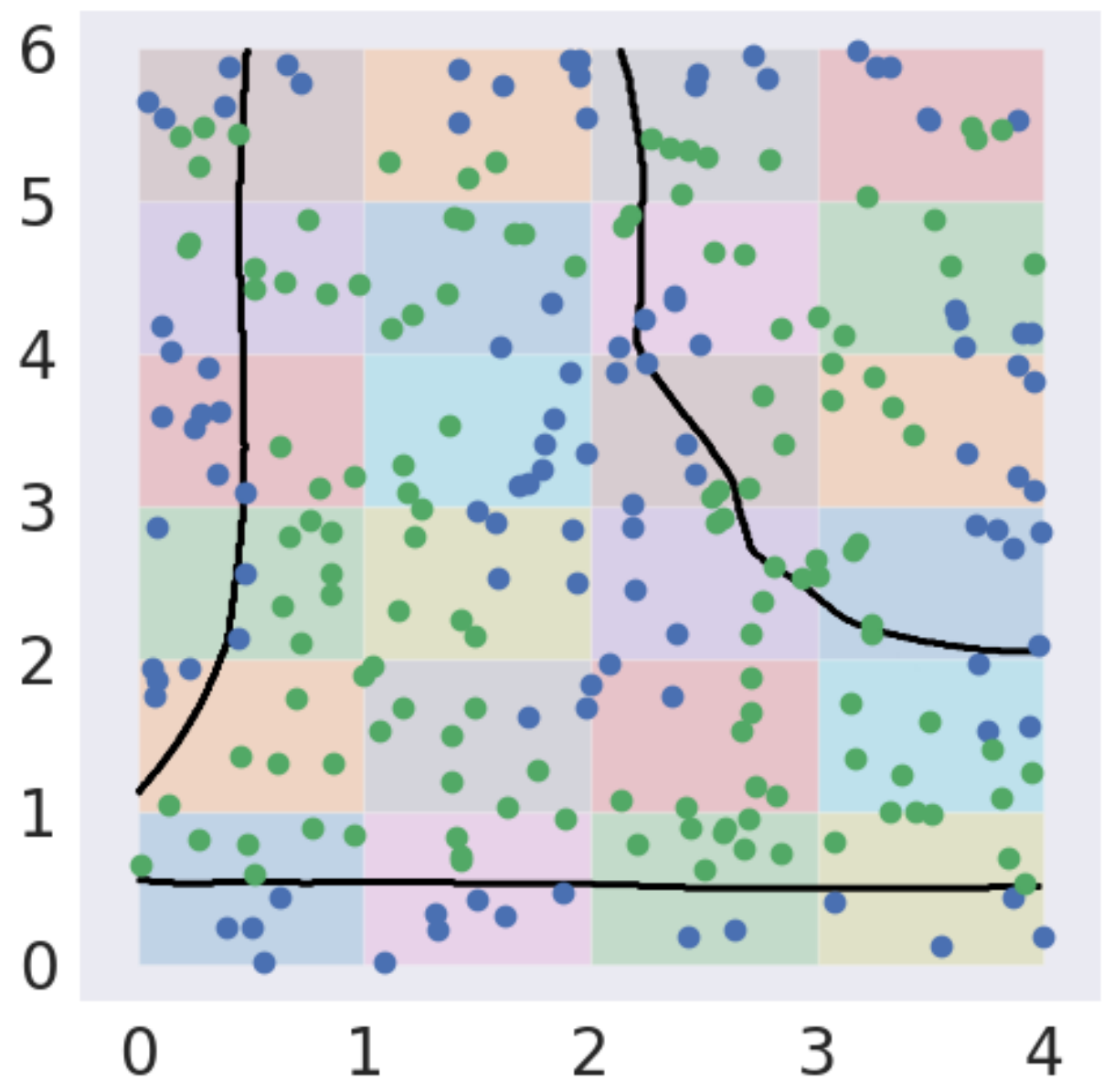}
    \caption{}
  \end{subfigure}
  \begin{subfigure}[b]{0.15\linewidth}
    \includegraphics[width=\linewidth]{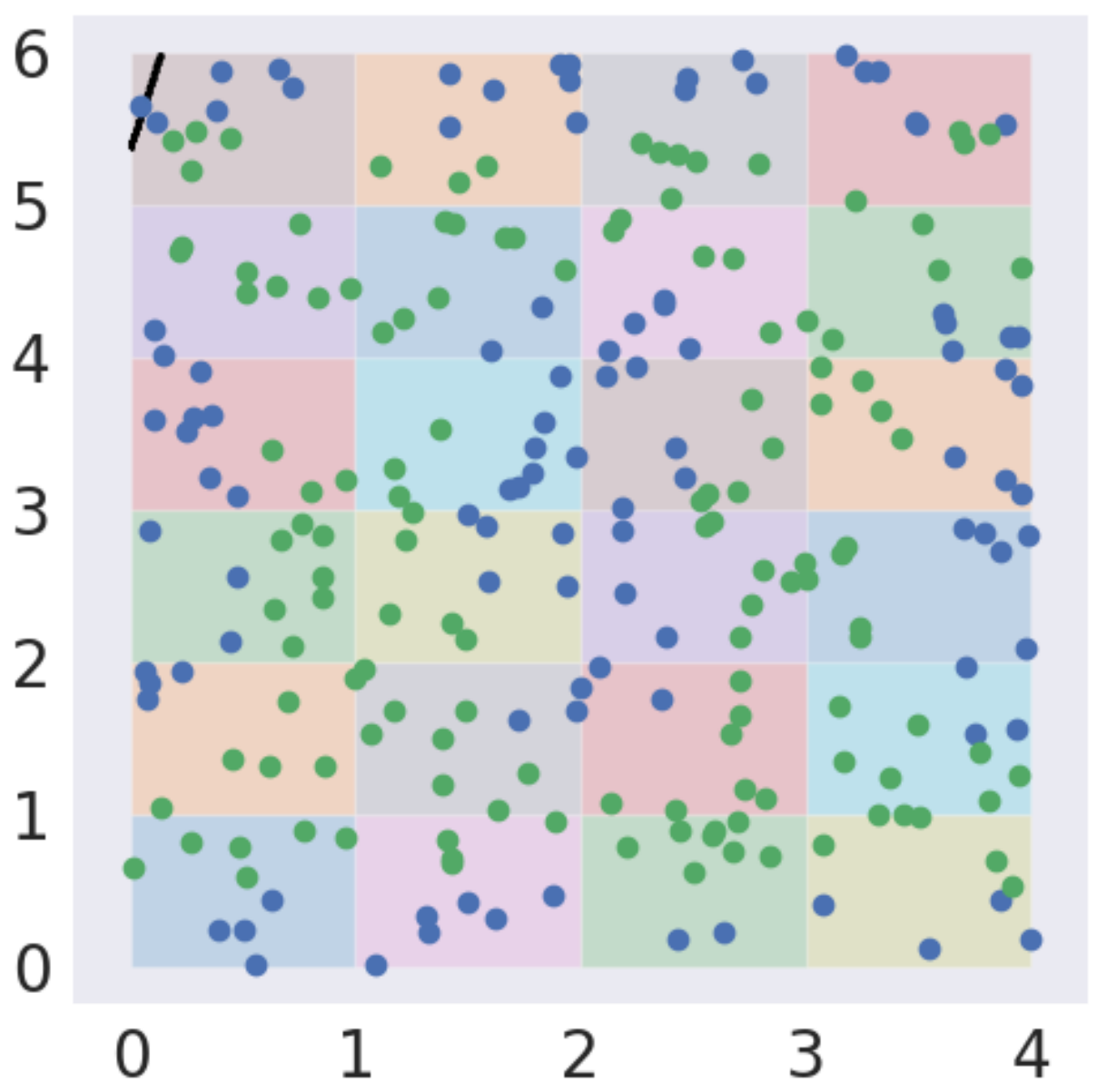}
    \caption{}
  \end{subfigure}
  \begin{subfigure}[b]{0.15\linewidth}
    \includegraphics[width=\linewidth]{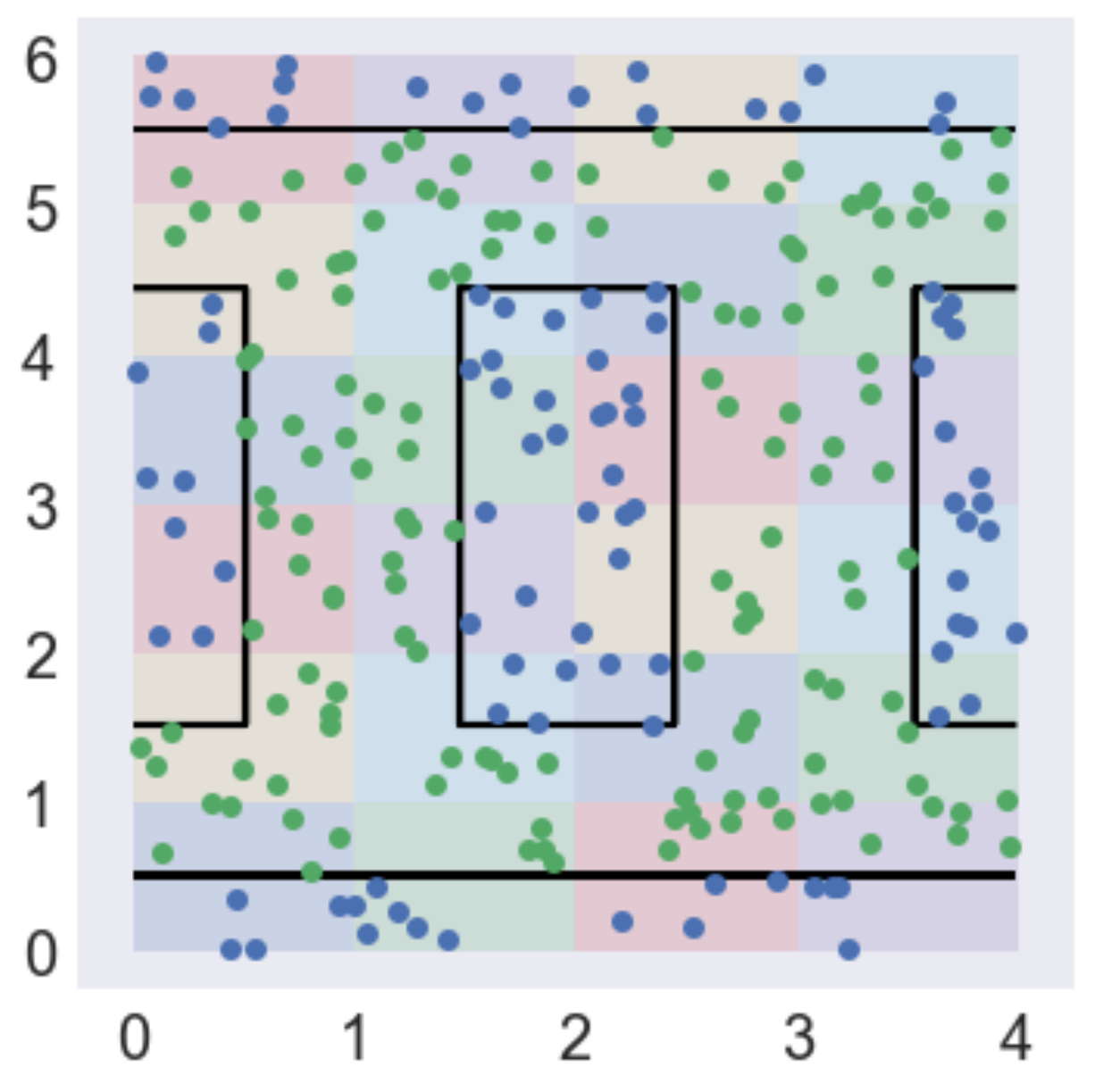}
    \caption{}
  \end{subfigure}
  \caption{(a) Ground truth decision boundary with 25 regions; green represents positive labels. \emph{(b)} Minima with no regularization. \emph{(c)} Minima with no data augmentation. \emph{(d)} Minima with no pruning or determinism  in training trees. \emph{(e)} Minima with bad learning rate. \emph{(f)} Minima using optimization innovations. Colored patches represent regions.}
  \label{fig:toyexp4}
\end{figure}

These optimization innovations are crucial for learning with regional tree regularization. Without them, optimization is very unstable, resulting in undesirable minima. Figure~\ref{fig:toyexp4} shows a few examples in a synthetic dataset: without data augmentation (c), there are not enough examples to fully train each surrogate, resulting in poor estimates of $\Omega^{\texttt{regional}}$ in which we converge to the same minima as no regularization (b); without pruning and fixing seeds, the path lengths vary due to randomness in fitting a decision tree, which can lead to over- or under- estimating the true APL. As shown in (d), this leads to strange decision boundaries. Finally, (e) shows the effect of large learning rates that leads to thrashing, resulting in a trivial decision boundary in efforts to minimize the loss. Only with the optimization innovations (f), do we converge to a properly regularized decision boundary.

\begin{figure*}[t!]
  \centering
  \begin{subfigure}[b]{0.16\textwidth}
    \includegraphics[width=\textwidth]{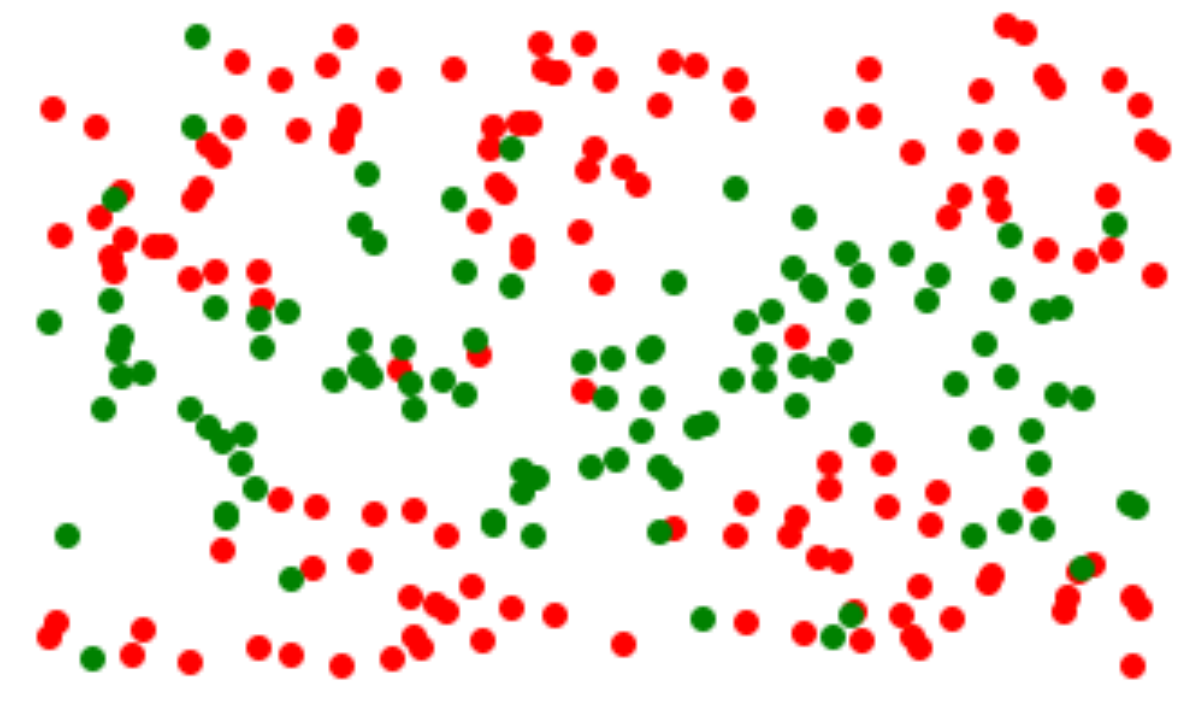}
    \caption{$\mathcal{D}^{\text{train}}$}
  \end{subfigure}
  \begin{subfigure}[b]{0.16\textwidth}
    \includegraphics[width=\textwidth]{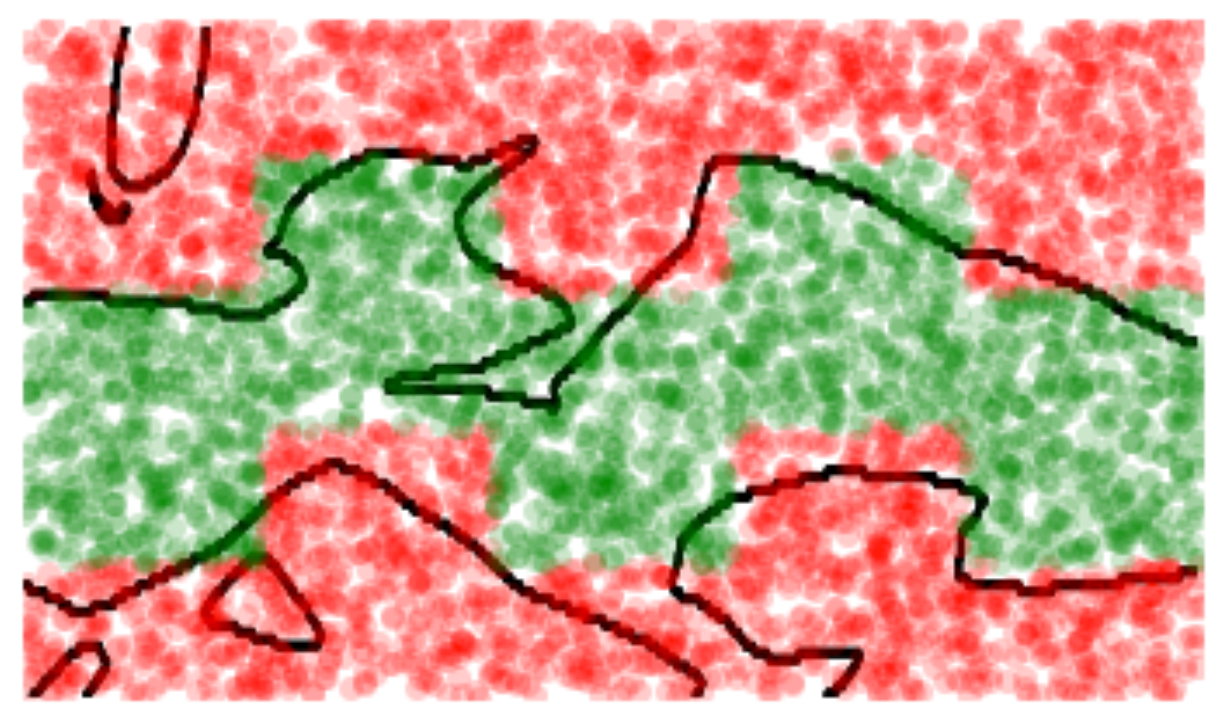}
    \caption{No Reg.}
  \end{subfigure}
  \begin{subfigure}[b]{0.16\textwidth}
    \includegraphics[width=\textwidth]{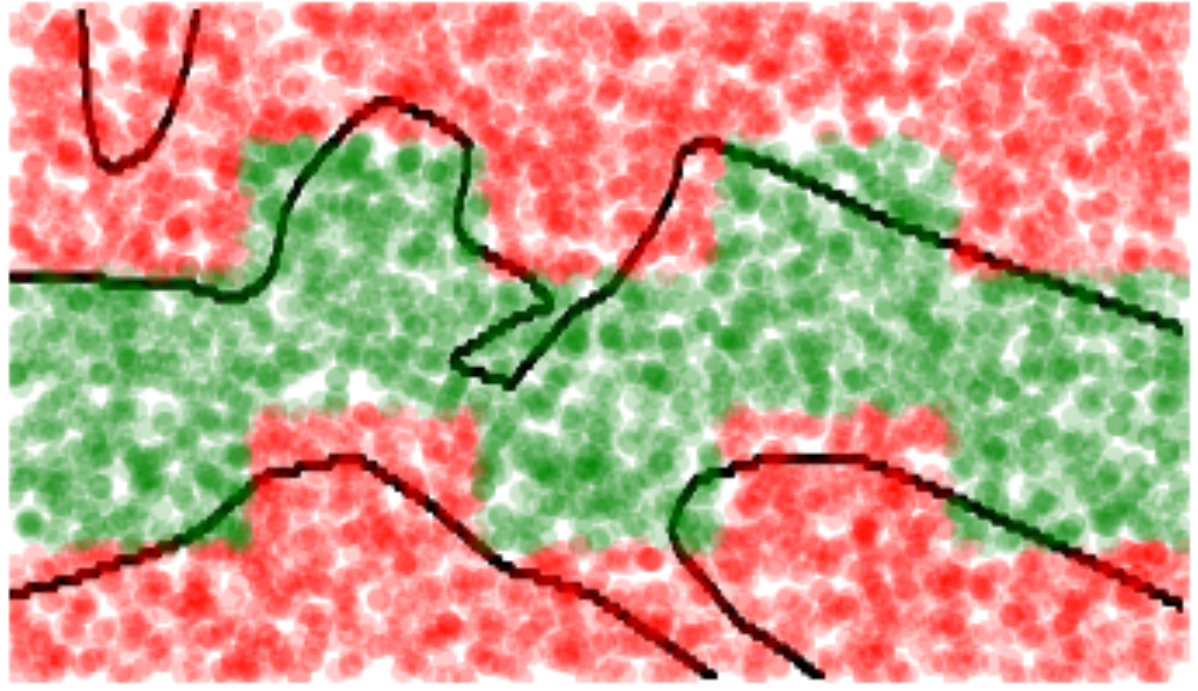}
    \caption{L2}
  \end{subfigure}
  \begin{subfigure}[b]{0.16\textwidth}
    \includegraphics[width=\textwidth]{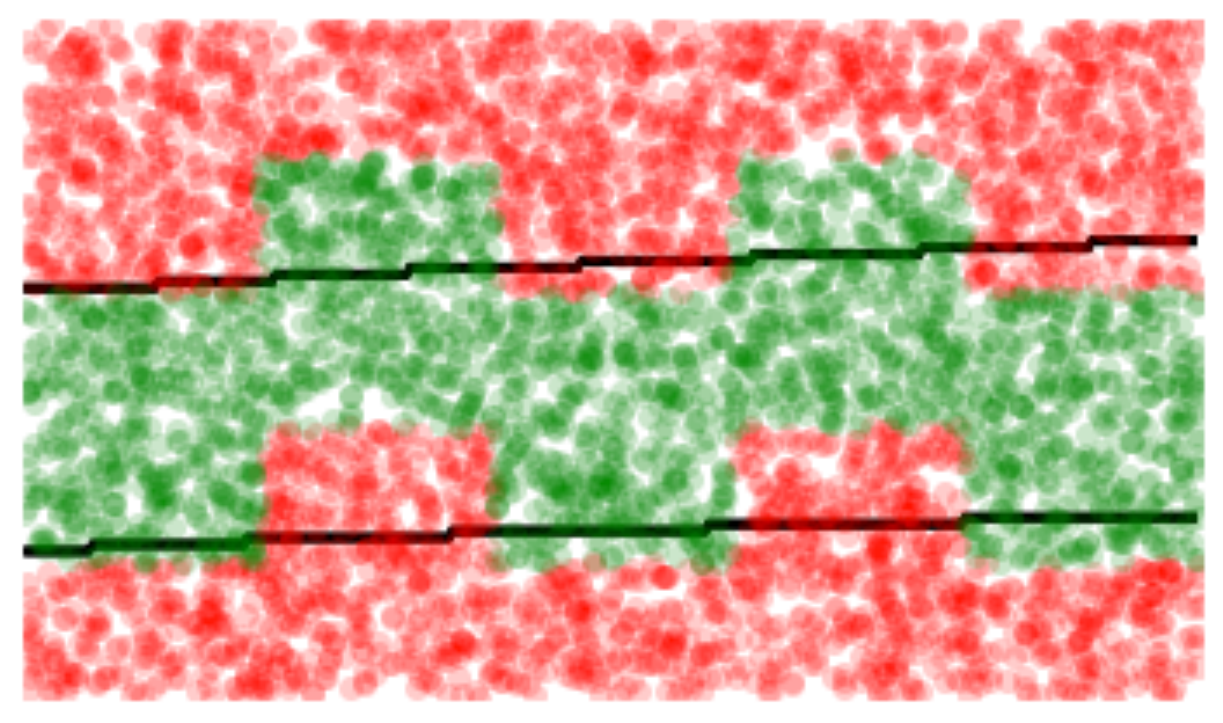}
    \caption{Global Tree}
  \end{subfigure}
  \begin{subfigure}[b]{0.16\textwidth}
    \includegraphics[width=\textwidth]{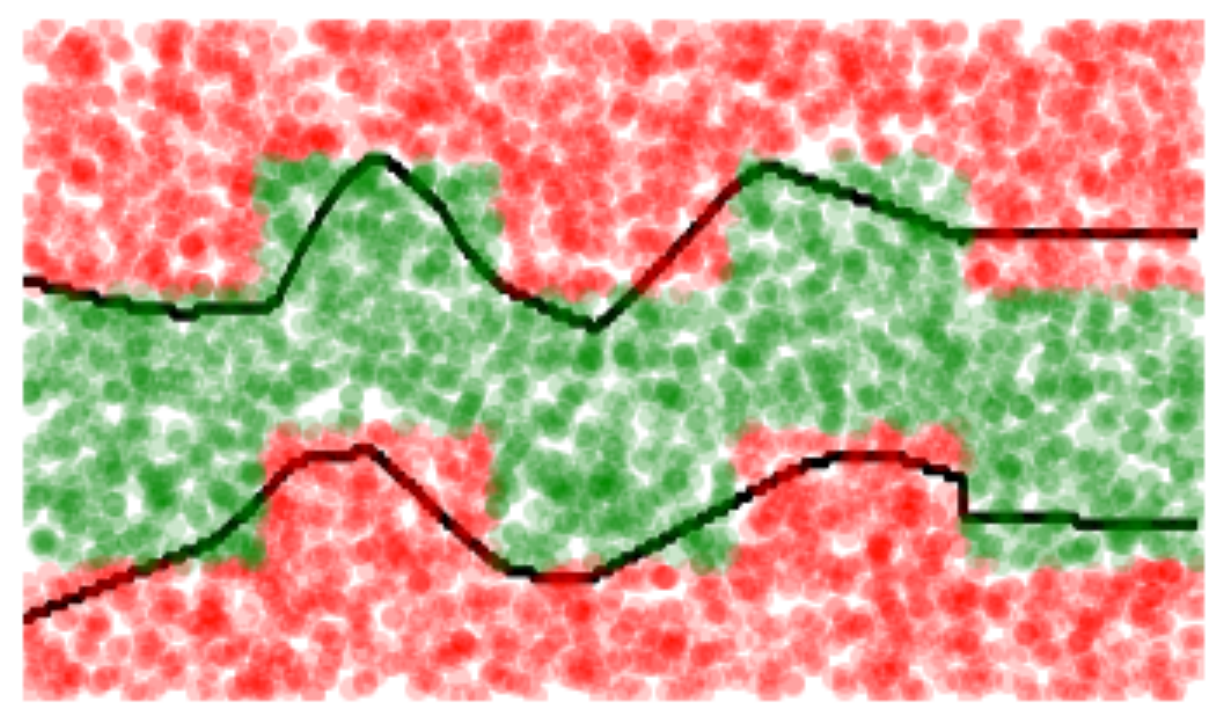}
    \caption{L$_1$ Reg. Tree}
  \end{subfigure}
  \begin{subfigure}[b]{0.16\textwidth}
    \includegraphics[width=\textwidth]{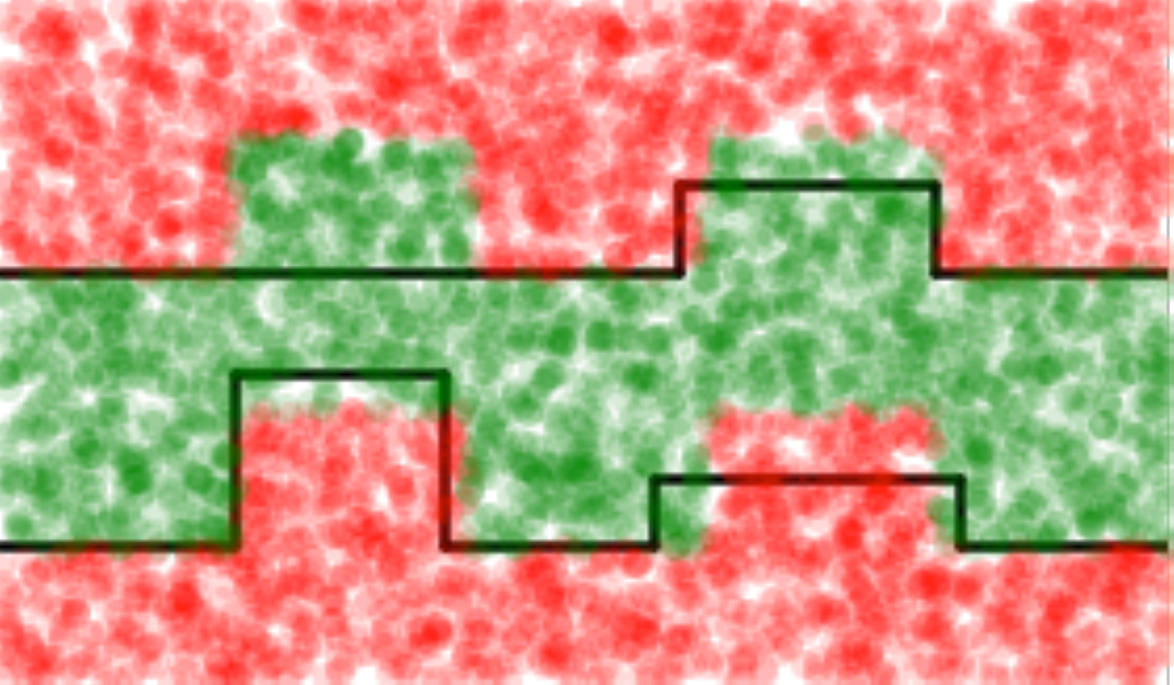}
    \caption{L$_0$ Reg. Tree}
  \end{subfigure}
  \caption{Synthetic data with a sparse training set \emph{(a)} and a dense test set \emph{(b)}. Due to sparsity, the division of five rectangles is not trivial to uncover from \emph{(a)}. \emph{(c-g)} show contours of decision functions learned with varying regularizations. Only the regional tree regularized model captures the vertical structure of the five regions, leading to high accuracy.}
  \label{fig:toy}
\end{figure*}

\section{Demonstration: Five Rectangles Dataset}
\label{sec:toy}
To build intuition, we present experiments in a toy setting: We define a ground-truth classification function composed of five rectangles (height of 0.5 and width of 1) in $\mathbb{R}^2$ concatenated along the x-axis to span the domain of $[0, 5]$. The first three rectangles are centered at $y=0.4$ (shifted slightly downwards) while the remaining two rectangles are centered at $y=0.6$ (shifted slightly upwards). The training dataset is intended to be sparse, containing only 250 points with the labels of 5\% of points randomly flipped to introduce noise and encourage overfitting. In contrast, the test dataset is densely sampled without noise. This is intended to model real-world settings where regional structure is only partially observable from an empirical dataset. It is exactly in these contexts that prior knowledge can be helpful.
\begin{table}[h!]
\centering
\begin{tabular}{c c c c}
\toprule
Model & Test Acc. & Test APL \\
\midrule
Unregularized & 0.8296 & 17.9490 \\
L2 ($\lambda=0.001$) & 0.8550 & 16.1130 \\
Global Tree ($\lambda=1$) & 0.8454 & 6.3398 \\
L$_1$ Regional Tree ($\lambda=0.1$) & 0.9168 & 10.1223 \\
L$_0$ Regional Tree ($\lambda=0.1$) & 0.9308 & 8.1962 \\
\bottomrule
\end{tabular}
\caption{Classification performance on a toy demonstration with varying regularizations. The reported test APL is averaged over APLs in each of the five regions.}
\label{table:toy}
\end{table}
Figure~\ref{fig:toy} show the learned decision boundary with (b) no regularization, (c) L2 regularization, (d) global tree regularization, and (e,f) regional tree regularization. As global regularization is restricted to penalizing all data points evenly, it fails to find the happy medium between being too complex or too simple. In other words, increasing the regularization strength quickly causes the target neural model to collapse from a complex nonlinear decision boundary to a single axis-aligned boundary. As shown in (d), this fails to capture any structure imposed by the five rectangles\footnote{It might be possible to capture the true structure (in a simple domain such as this) with very careful tuning of the hyperparameters in global tree regularization. However, this is difficult to do consistently and regional tree regularization presents a much easier solution.}. Similarly, if we increase the strength of L2 regularization even slightly from (c), the model collapses to the trivial solution of predicting entirely one label. Only regional tree regularization (e,f) is able to model the up-and-down curvature of the true decision function. With high $\lambda$, L$_0$ regional tree regularization produces a more axis-aligned decision boundary than its L$_1$ equivalent, primarily because we can regularize complex regions more harshly without collapsing simpler regions. Knowledge of the region divisions provides a model with prior information about underlying structure in the data; we should expect that with such information, a regionally regularized model can better prevent itself from over- or underfitting. We train for 500 epochs with a learning rate of 4e-3, a minibatch size of 32, retrain the surrogate function every epoch (a loop over the full training dataset) and sample 1000 weights from the convex hull each time. Decision trees were trained with $h=1$. Table~\ref{table:toy} compares metrics between the different regularizations: although the regional tree regularization is slightly more complex than global tree regularization, it comes with a large increase in accuracy.

\begin{figure*}[t]
  \centering
  \begin{subfigure}[b]{0.9\textwidth}
    \includegraphics[width=\textwidth]{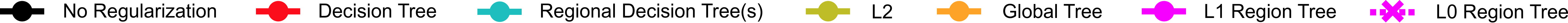}
  \end{subfigure}
  \begin{subfigure}[b]{0.24\textwidth}
    \includegraphics[width=\textwidth]{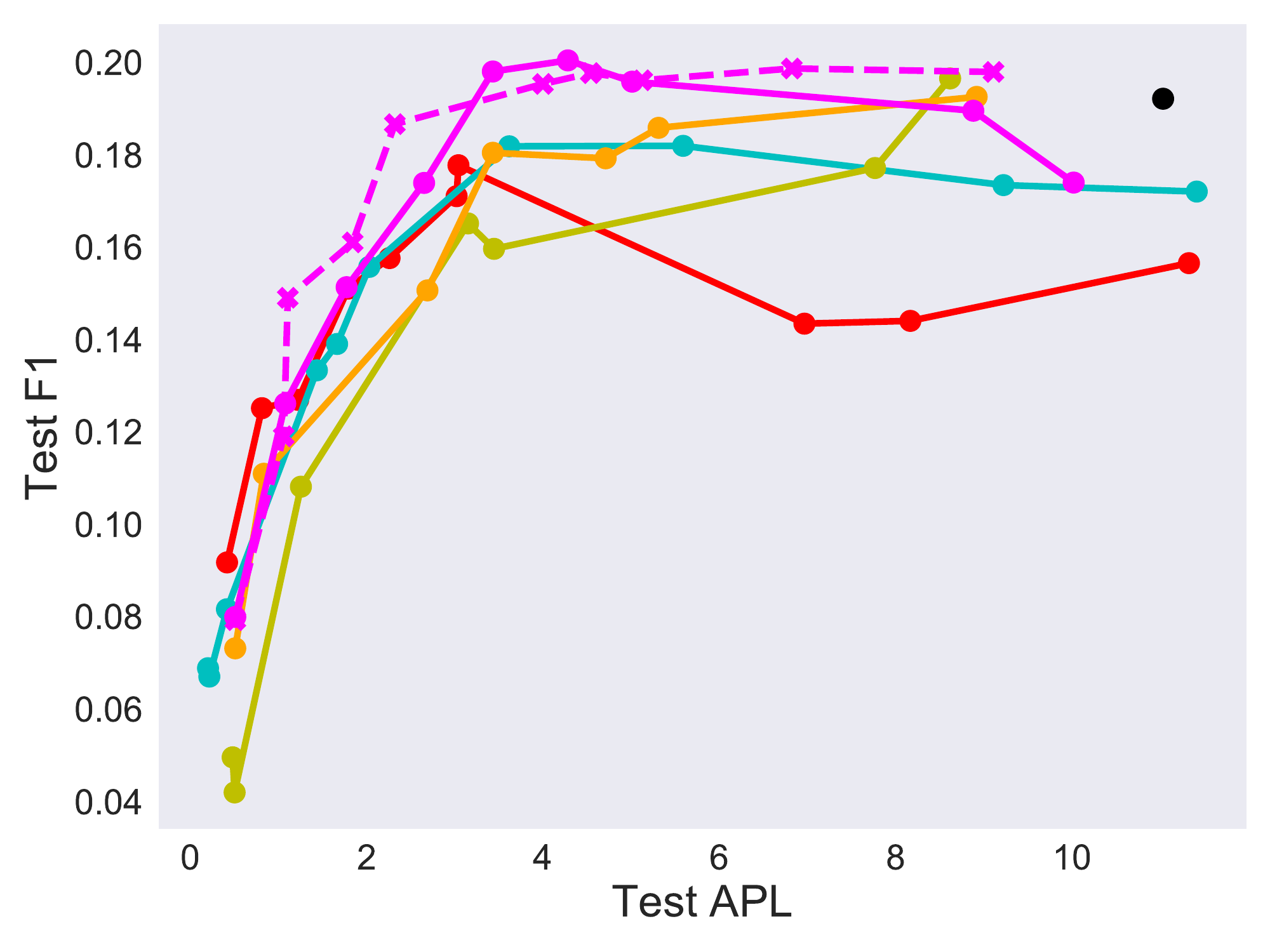}
    \caption{Bank}
  \end{subfigure}
  \begin{subfigure}[b]{0.24\textwidth}
    \includegraphics[width=\textwidth]{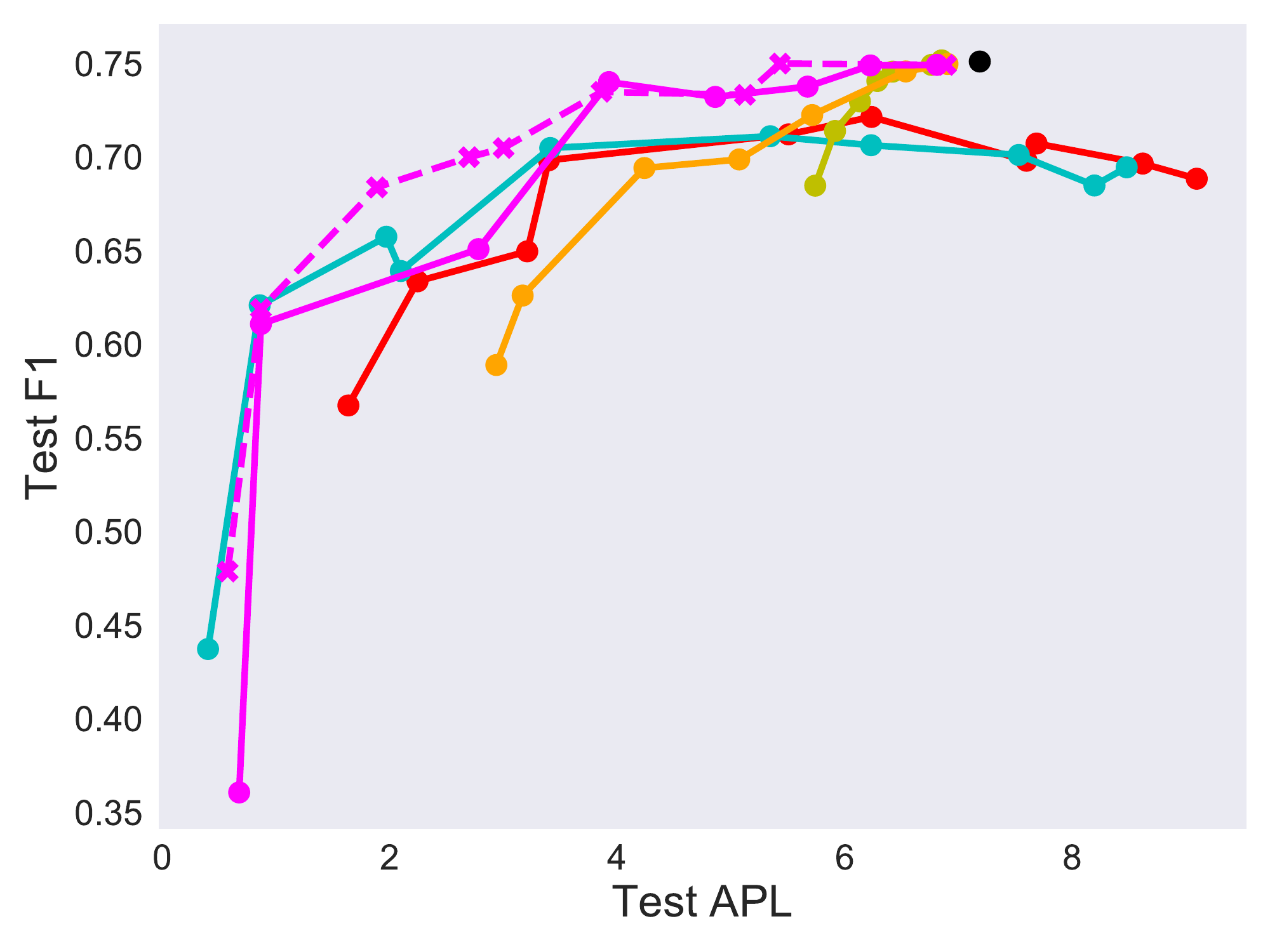}
    \caption{Gamma}
  \end{subfigure}
  \begin{subfigure}[b]{0.24\textwidth}
    \includegraphics[width=\textwidth]{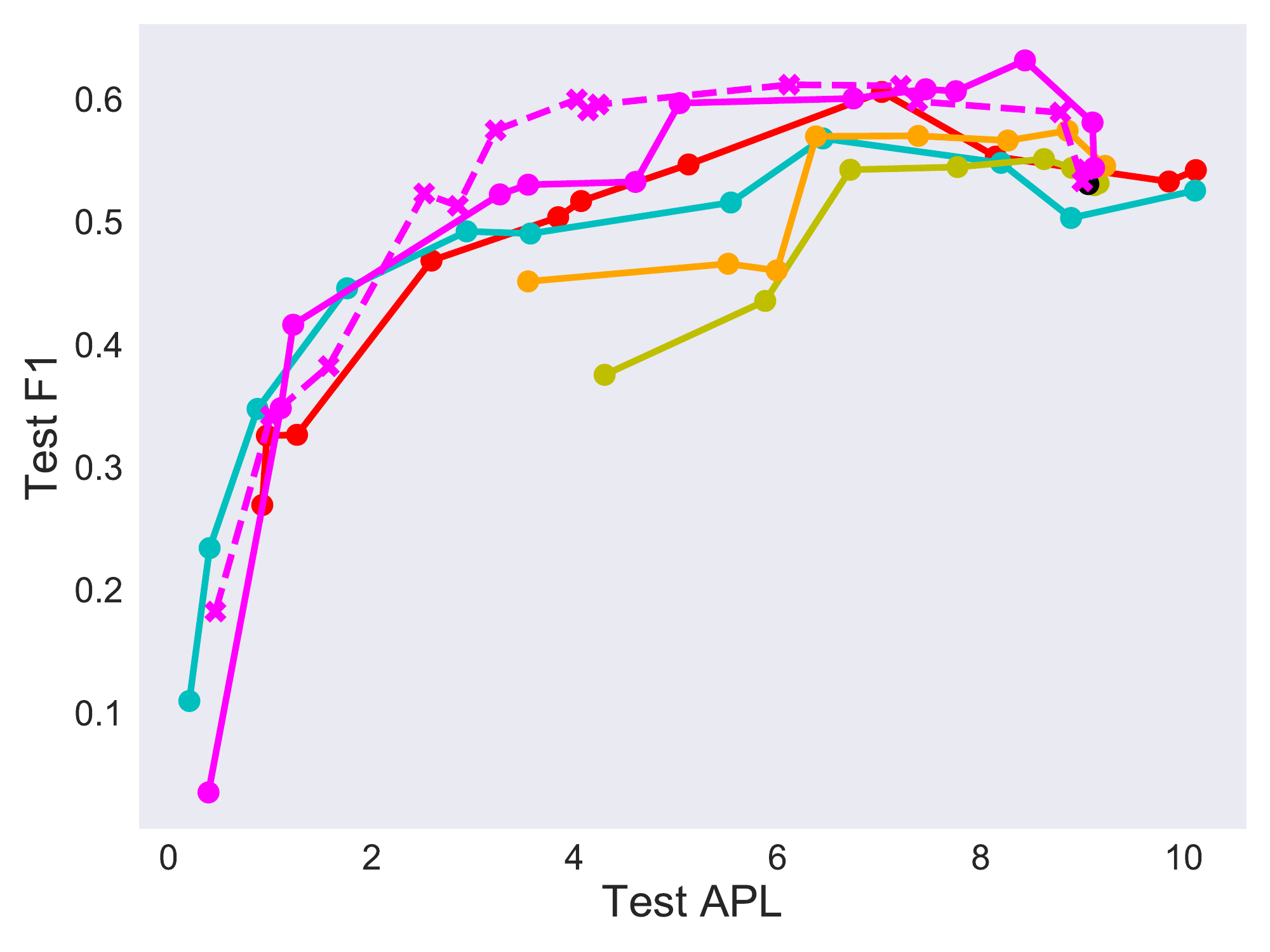}
    \caption{Adult}
  \end{subfigure}
  \begin{subfigure}[b]{0.24\textwidth}
    \includegraphics[width=\textwidth]{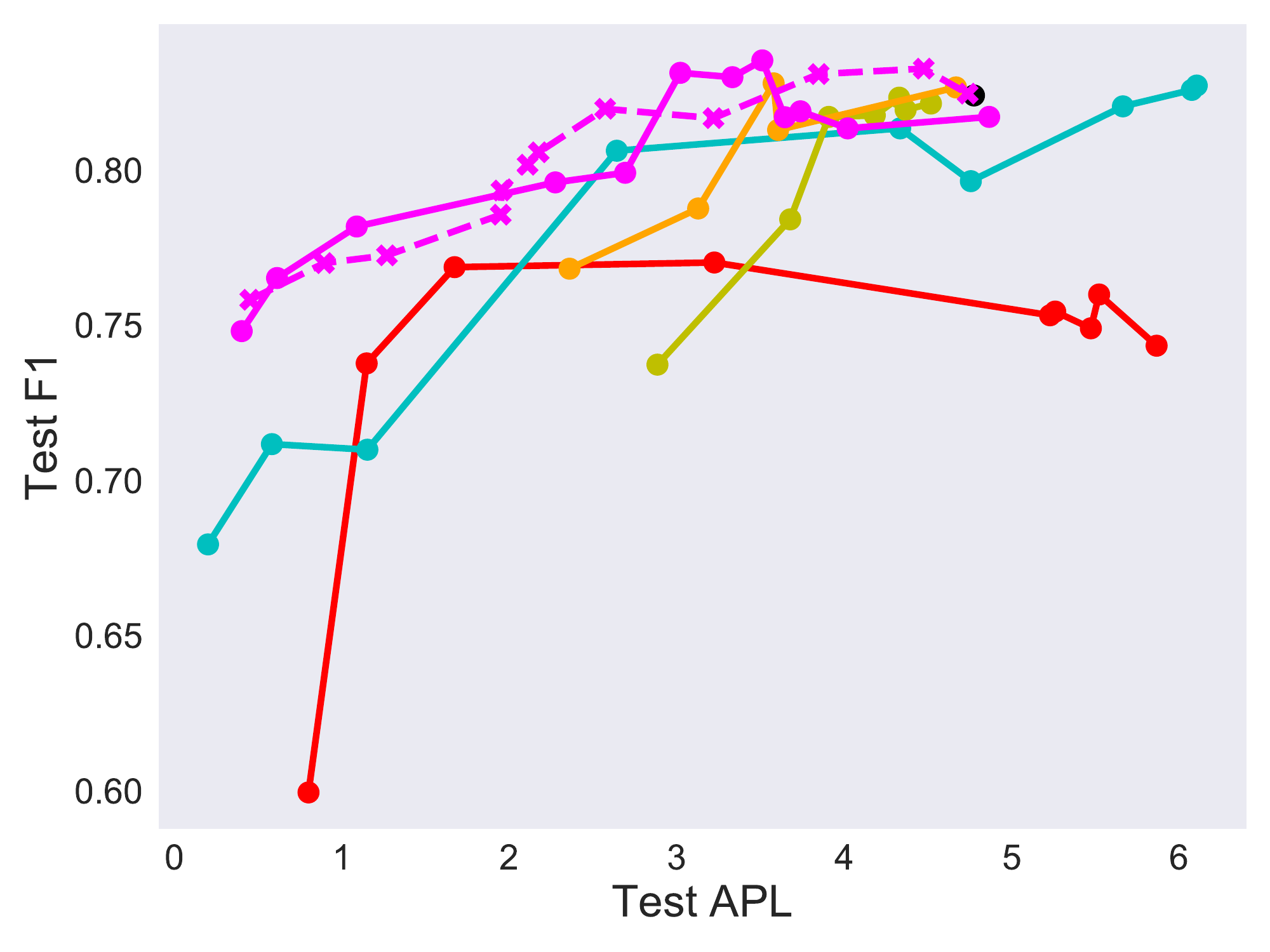}
    \caption{Wine}
  \end{subfigure}
  \caption{(a-d) Comparison of regularizers (L2, global tree, regional tree) on four datasets from the UCI repository. Each subfigure plots the average APL over 5 regions (computed on a held-out test set) against the test F1 score. The ideal model is with high accuracy and low APL i.e. the upper left diagonal of each plot. In each setting, regional tree regularized models are able to find more low APL minima than global explanations and consistently achieves the highest performance at low APL. In contrast, the performance of global tree and L2 regularization quickly decays as the regularization strength increases.}
  \label{fig:uci}
\end{figure*}

\section{Application: UC Irvine Prediction Tasks}
Having seen a synthetic dataset, we transition to more realistic machine learning settings. Without loss of generality, we focus on feedforward networks, or MLPs. The same ideas of regional explanation using decision trees can be trivially extended to sequential models (like the GRU used above) or convolutional models. For the experiments below, we set the target neural model to a 6 layer MLP with 128, 128, 128, 64, 64, and $Q$ dimensional hidden layers respectively. The final layer contains a node for each output dimension. We use leaky ReLU nonlinearities in between each layer. Each surrogate remains a very shallow MLP.

\subsection{Evaluation Metrics}
We wish to compare models with global and regional explanations. However, given $\theta\in\Theta$, $\Omega^{\texttt{regional}}(\theta)$ and $\Omega^{\texttt{global}}(\theta)$ are not directly comparable: subtly, the APL of a global tree is often an overestimate for data points in a single region. To reconcile this, for any globally regularized model, we separately compute $\Omega^{\texttt{regional}}(\theta)$ as an evaluation criterion. In this context, $\Omega^{\texttt{regional}}$ is used only for evaluation; it does not appear in the objective nor training. We do the same for baseline models, L2 regularized models, and unregularized models. From this point on, if we refer to average path length (e.g. Test APL, APL, path length) outside of the objective, we are referring to the evaluation metric, $\Omega^{\texttt{regional}}(\theta)$.

\subsection{Datasets}

We apply regional tree regularization to a suite of four popular machine learning datasets from UC Irvine repository \cite{Dua:2017}.
We briefly provide context for each dataset and show results comparing the regularization methods in effectiveness. We choose a generic method for defining regions to showcase the wide applicability of regional regularization: we use $\mathcal{D}$ to fit a $k$-means clustering model with $k=5$. Each example $\mathbf{x}_n \in \mathcal{D}$ is then assigned a number, $s_n \in \{1,2,3,4,5\}$. We define $X_r = \{ \mathbf{x}_n | s_n = r \} \subseteq \mathcal{X}^P$.
\vspace{1.4mm}

\noindent\textbf{Bank Marketing} (Bank): 45,211 rows collected from marketing campaigns for a bank \cite{moro2014data}. $\mathbf{x}_n$ has 17 features describing a recipient of the campaign (age, education, etc). There is one binary ouput indicating whether the recipient subscribed.
\vspace{1.4mm}

\noindent\textbf{MAGIC Gamma Telescope} (Gamma): 19,020 samples from a simulator of high energy Gamma particles in an Cherenkov telescope. There are 11 input features for afterimages of photon pulses, and one binary output discriminating between signal and background.
\vspace{1.4mm}

\noindent\textbf{Adult Income} (Adult): 48,842 data points with 14 input features (age, sex, etc.), and a binary output indicating if an individual's income exceeds \$50,000 per year \cite{kohavi1996scaling}.
\vspace{1.4mm}

\noindent\textbf{Wine Quality} (Wine): 4,898 examples describing wine from Portugal. Each row has a quality score from 0 to 10 and eleven variables based on physicochemical tests for acidity, sugar, pH, etc. We binarize the target where a positive label indicates a score of at least 5.
\vspace{1.4mm}

In each dataset, the target neural model is trained for 500 epochs with 1e-4 learning rate using Adam \cite{kingma2014adam} and a minibatch size of 128. We train under 20 different $\lambda$ between 0.0001 and 10.0. We do not do early stopping to preserve overfitting effects. We use 250 samples from the convex hull and retrain every 50 gradient steps. We set $C=25$ for Wine and $C=100$ otherwise. Figure~\ref{fig:uci} (a-d) compare L2, global tree, and regional tree regularization with varying strengths. The points plotted show minima from 3 independent runs. We include three baselines: an unregularized model, a decision tree trained on $\mathcal{D}$ and, a set of  trees with one for each region (we call this: regional decision tree). For baseline trees, we vary $h$ where a higher $h$ is a more regularized decision tree.

\subsection{Results}

Some patterns are apparent. First, an unregularized model (black) does poorly due to overfitting to a complex decision boundary, as the neural network is over-parameterized. Second, we find that L2 is \textit{not} a desirable regularizer for simulatability as it is unable to find many minima in the low APL region (see Gamma, Adult, and Wine under roughly 5 APL). Any increase in regularization strength quickly causes the target neural model to decay to an F1 score of 0, in other words, one that predict a single label. We see similar behavior with global tree regularization, suggesting that finding low complexity minima is challenging under global constraints. Third, regional tree regularization achieves the highest test accuracy in all datasets. We find that in the lower APL area, regional explanations surpasses global explanations in performance. For example, in Bank, Gamma, Adult, and Wine, we can see this at 3-6, 4-7, 5-8, 3-4 APL respectively. This suggests, like in the toy example, that it is easier to regularize groups rather than the entire input space as a whole. In fact, unlike global regularization, models constrained regionally are able to reach a wealth of minima in the low APL area. Lastly, we note that with high regularization strengths, regional tree regularization mostly converges in performance with regional decision trees, which is sensible as the neural network prioritizes distillation over performance.

\begin{figure}[h!]
  \centering
  \begin{subfigure}[b]{0.9\textwidth}
    \includegraphics[width=\textwidth]{v2/uci/legend.pdf}
  \end{subfigure}
  \begin{subfigure}[b]{0.24\textwidth}
    \includegraphics[width=\textwidth]{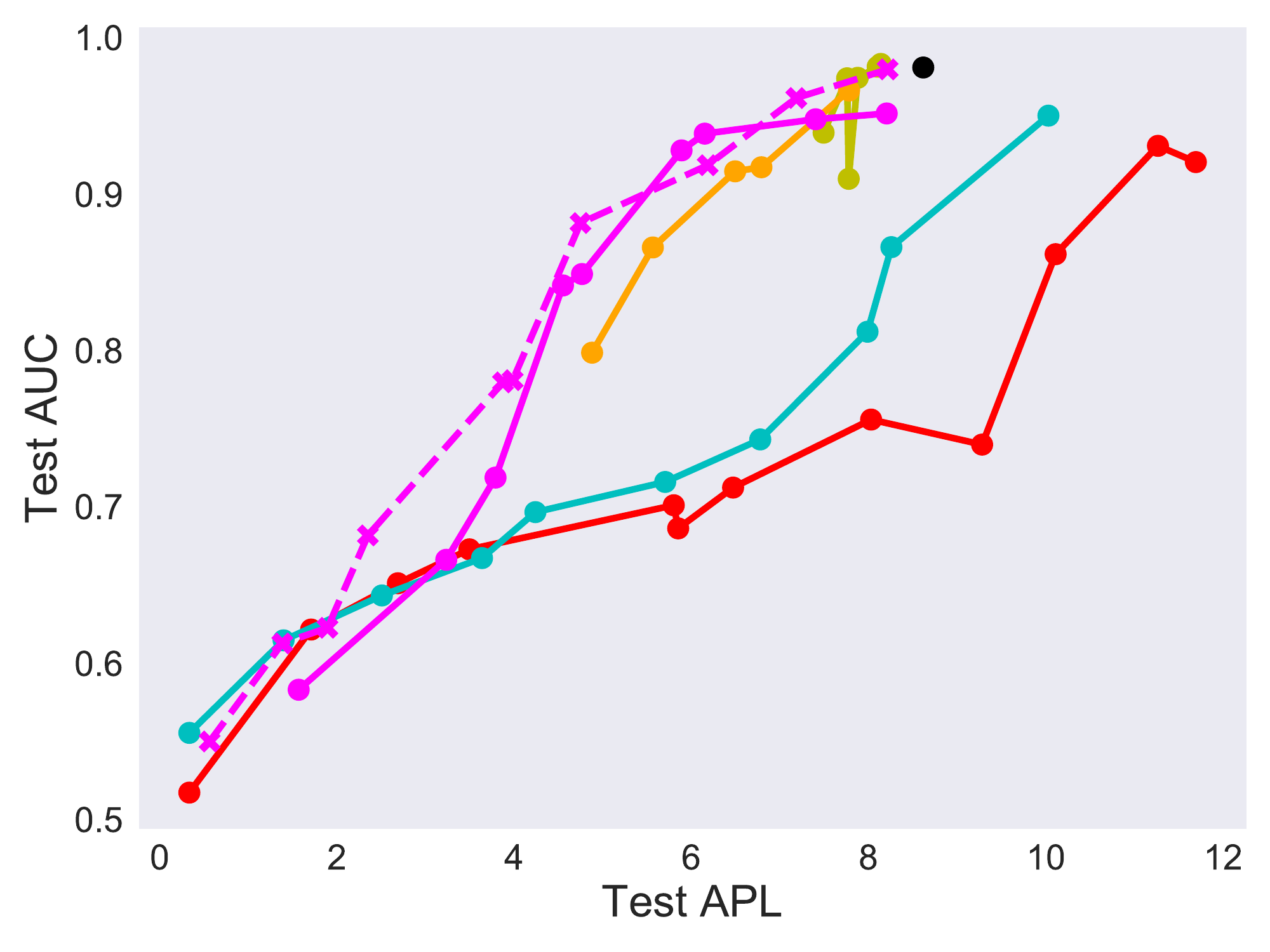}
    \caption{SOFA: Vaso.}
  \end{subfigure}
  \begin{subfigure}[b]{0.24\textwidth}
    \includegraphics[width=\textwidth]{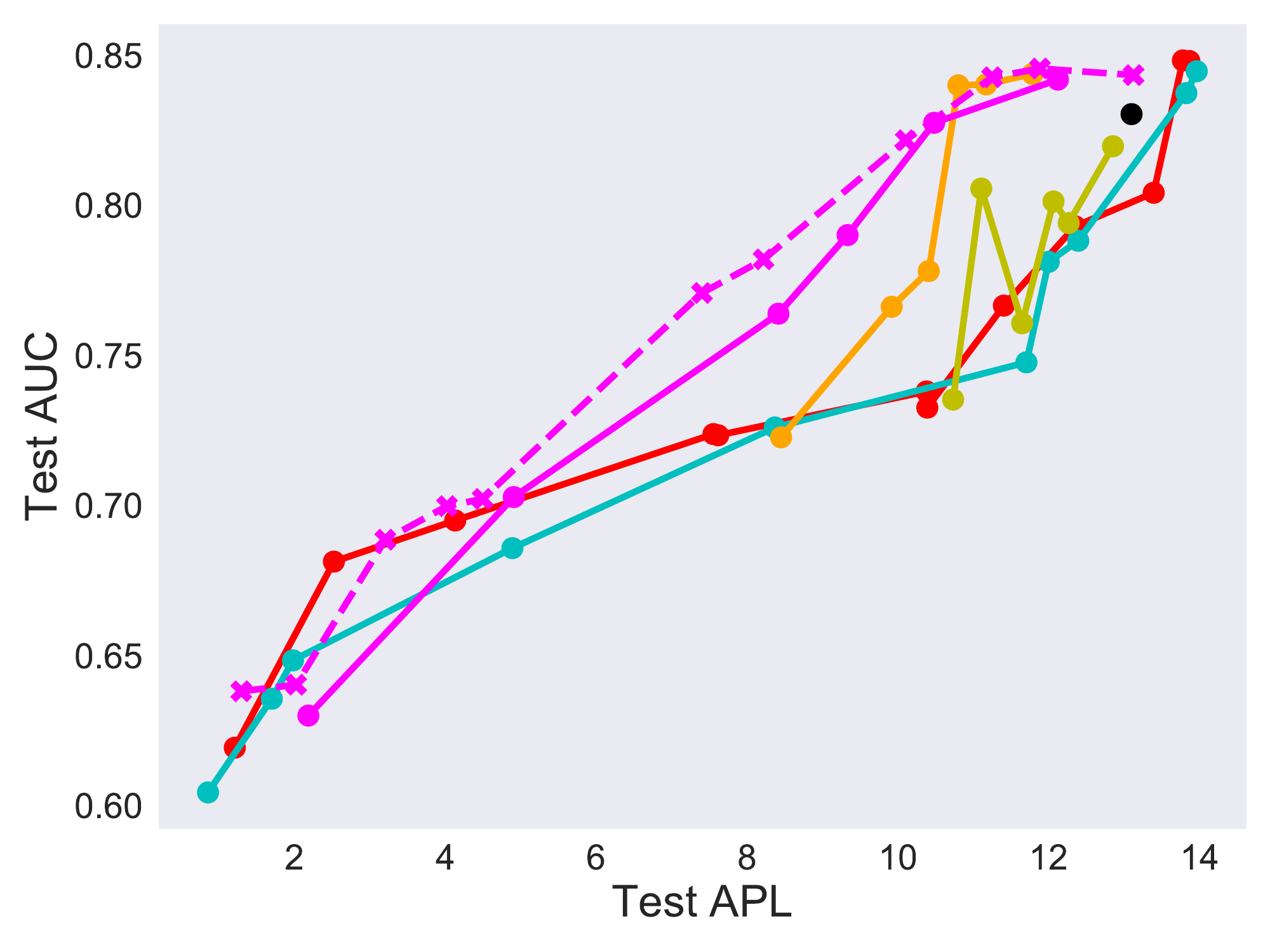}
    \caption{SOFA: Sedation}
  \end{subfigure}
  \begin{subfigure}[b]{0.24\textwidth}
    \includegraphics[width=\textwidth]{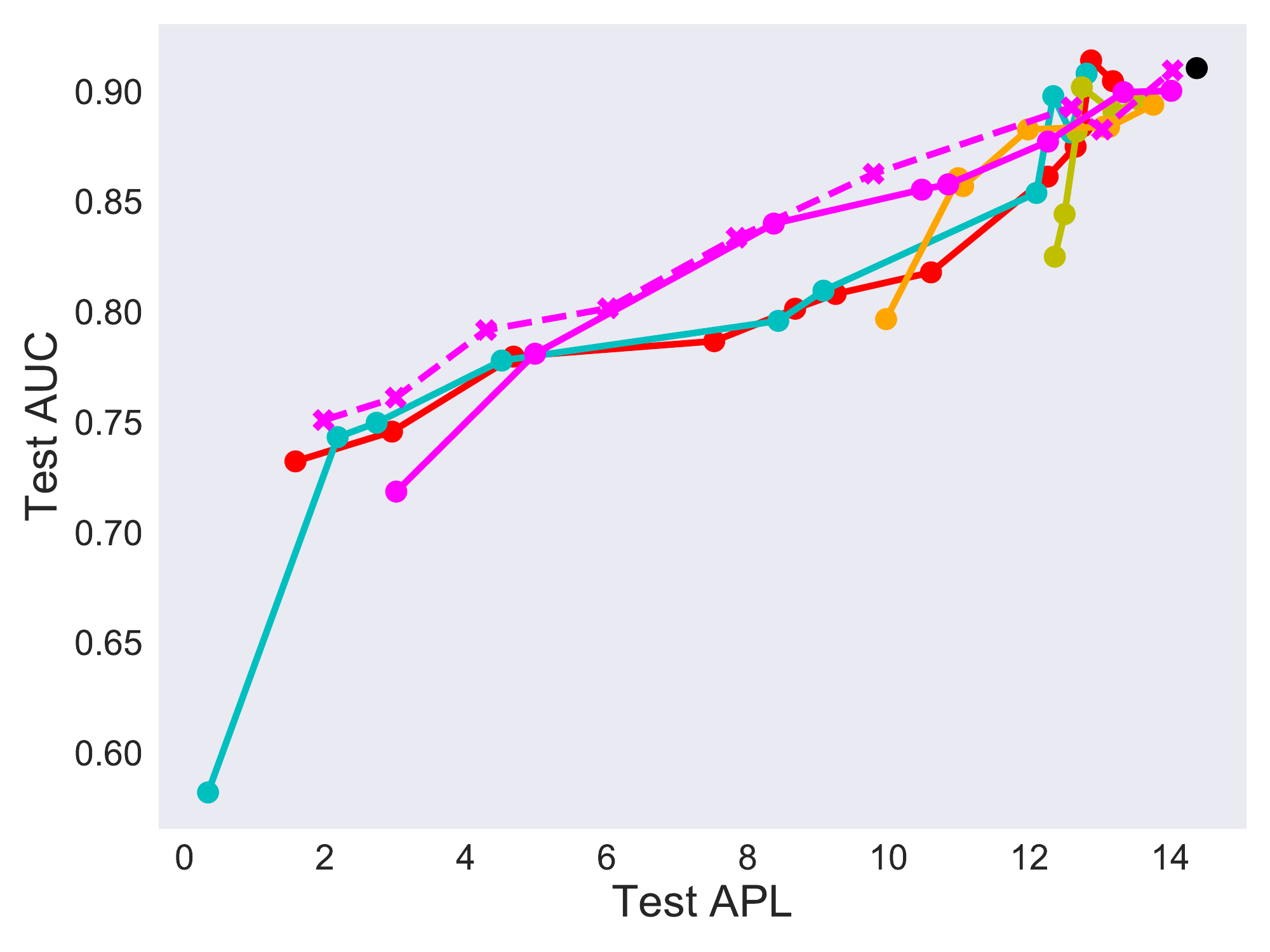}
    \caption{SOFA: Vent.}
  \end{subfigure}
  \begin{subfigure}[b]{0.24\textwidth}
    \includegraphics[width=\textwidth]{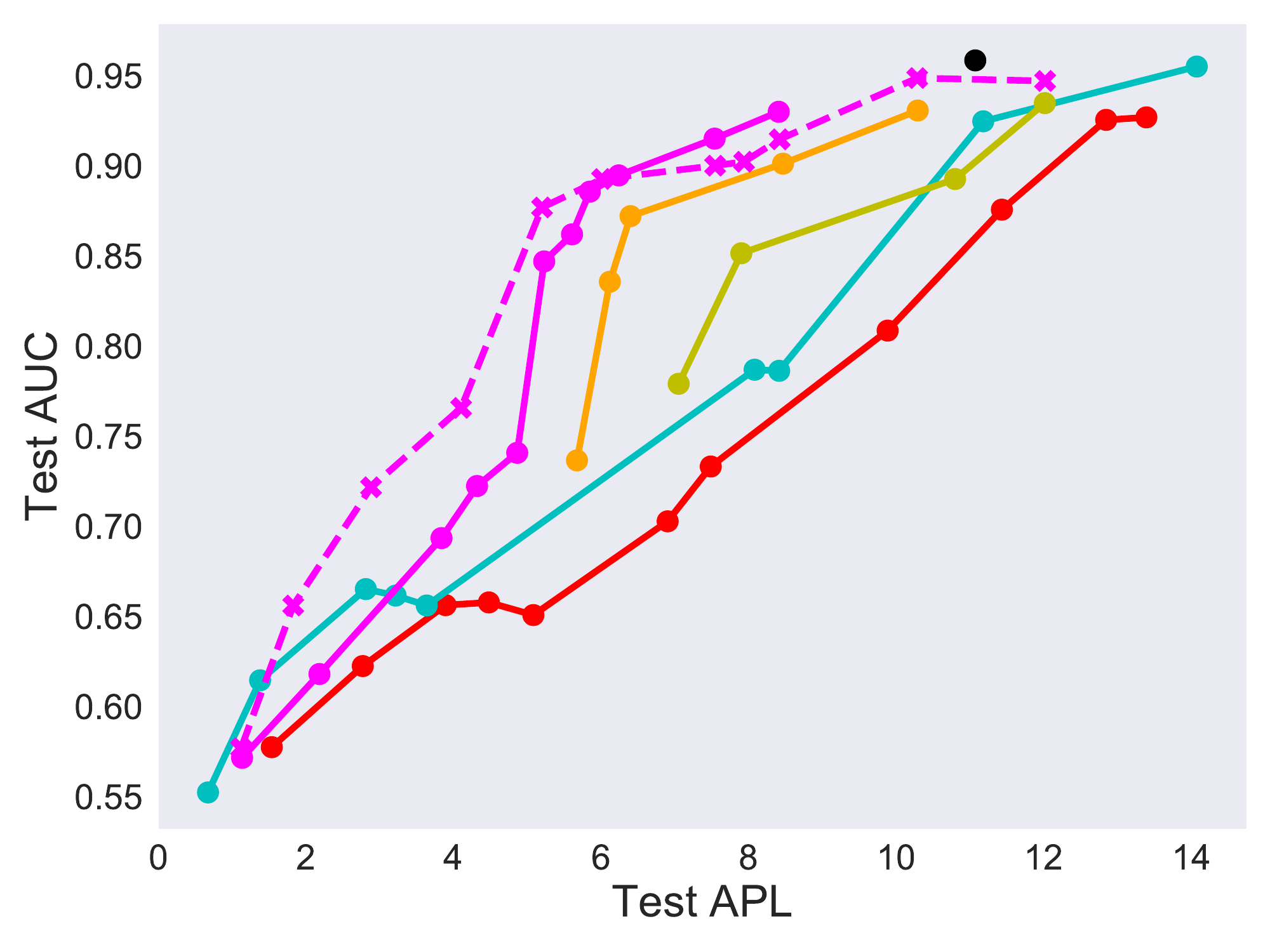}
    \caption{SOFA: Renal}
  \end{subfigure}
  \begin{subfigure}[b]{0.24\textwidth}
    \includegraphics[width=\textwidth]{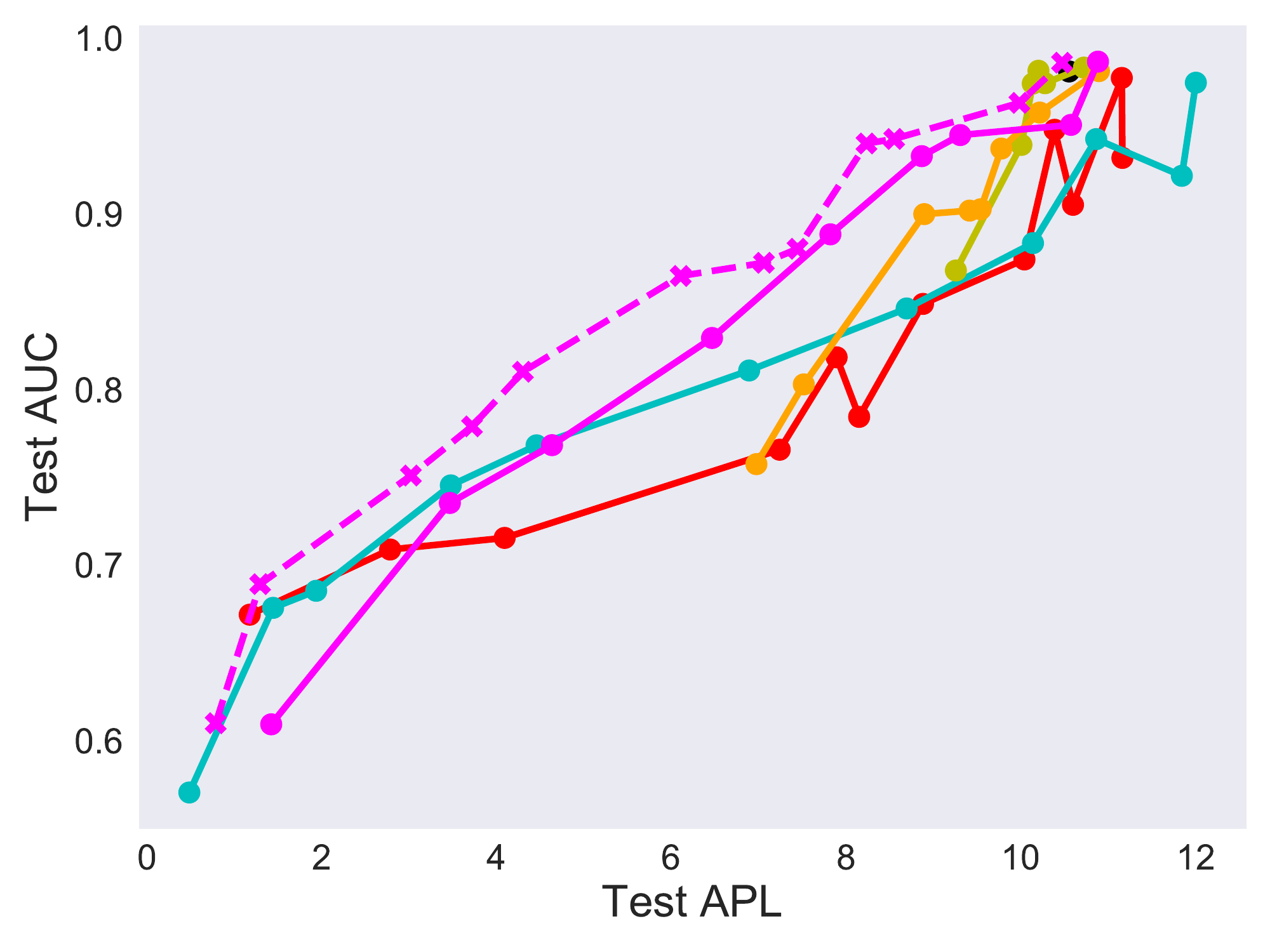}
    \caption{Careunit: Vaso.}
  \end{subfigure}
  \begin{subfigure}[b]{0.24\textwidth}
    \includegraphics[width=\textwidth]{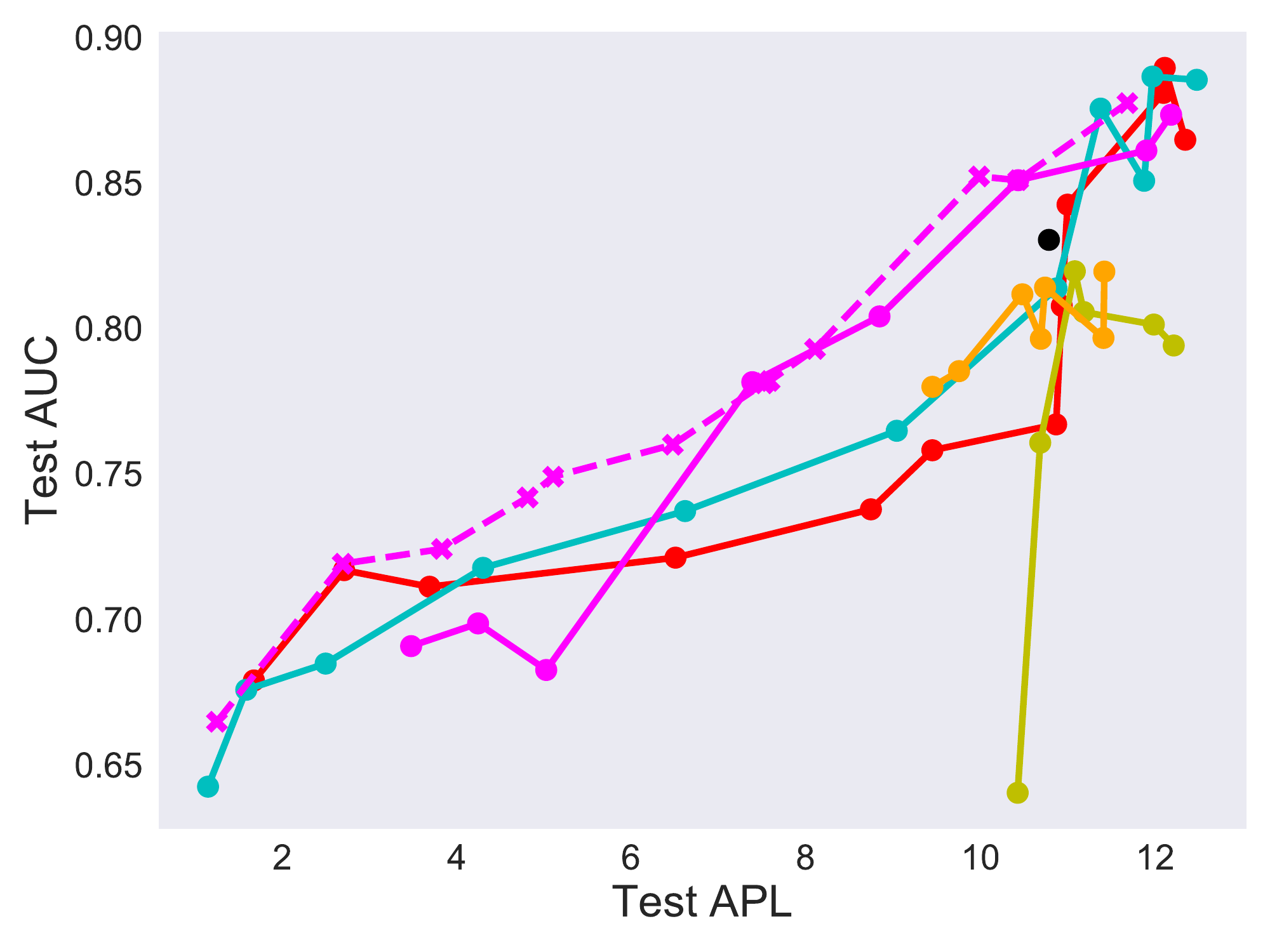}
    \caption{Careunit: Sedation}
  \end{subfigure}
  \begin{subfigure}[b]{0.24\textwidth}
    \includegraphics[width=\textwidth]{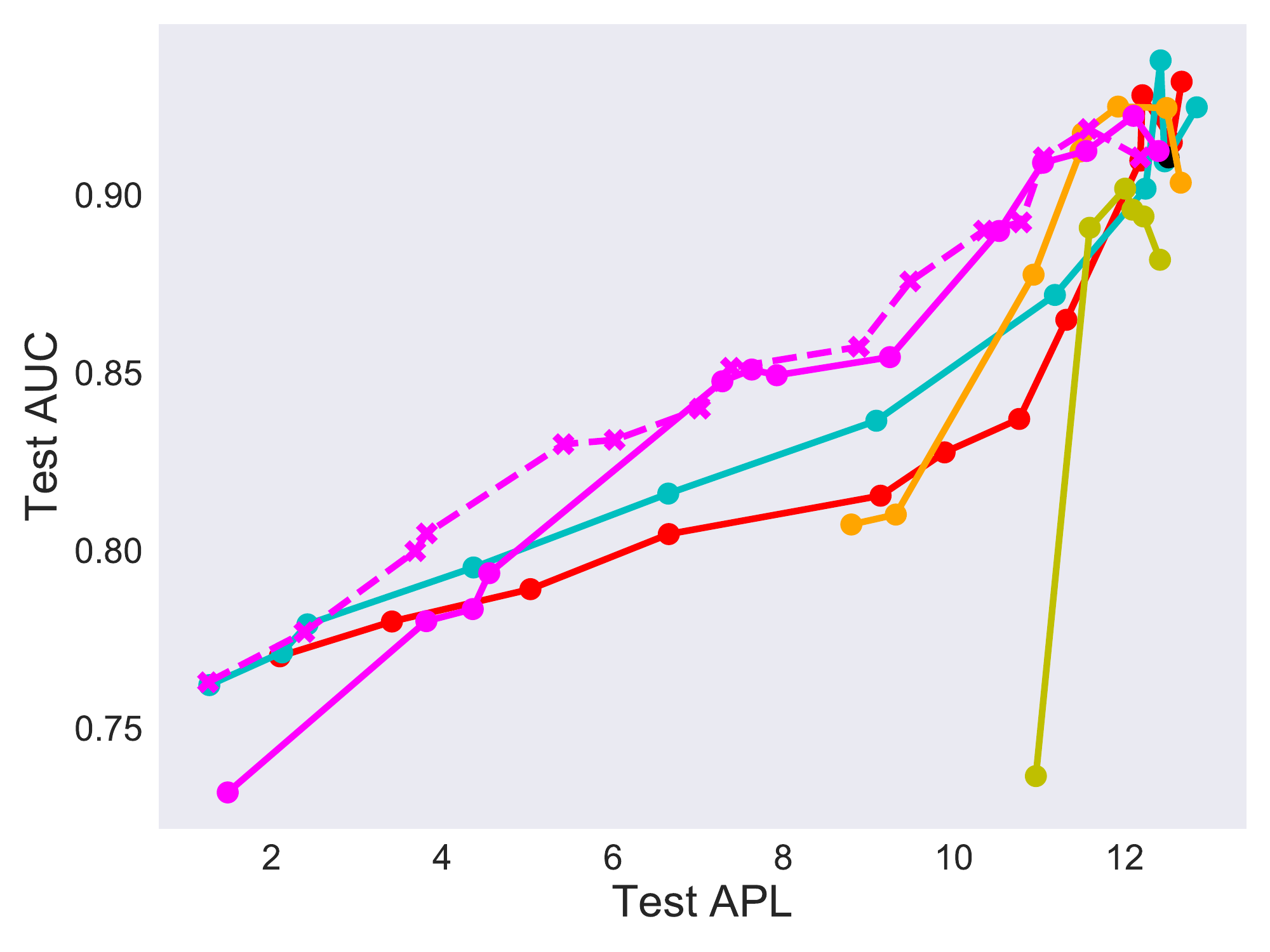}
    \caption{Careunit: Vent.}
  \end{subfigure}
  \begin{subfigure}[b]{0.24\textwidth}
    \includegraphics[width=\textwidth]{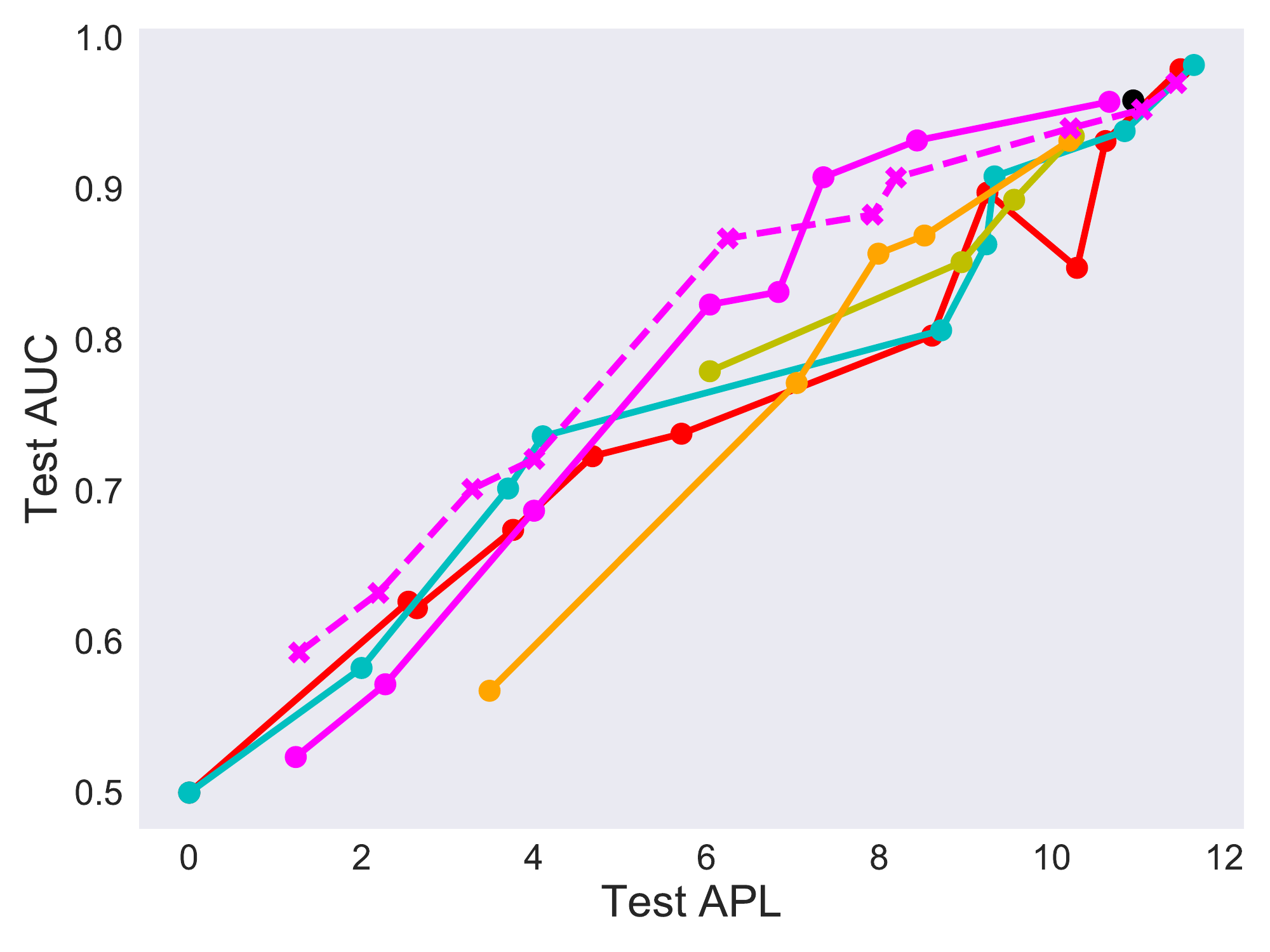}
    \caption{Careunit: Renal}
  \end{subfigure}
  \begin{subfigure}[b]{0.24\textwidth}
    \includegraphics[width=\textwidth]{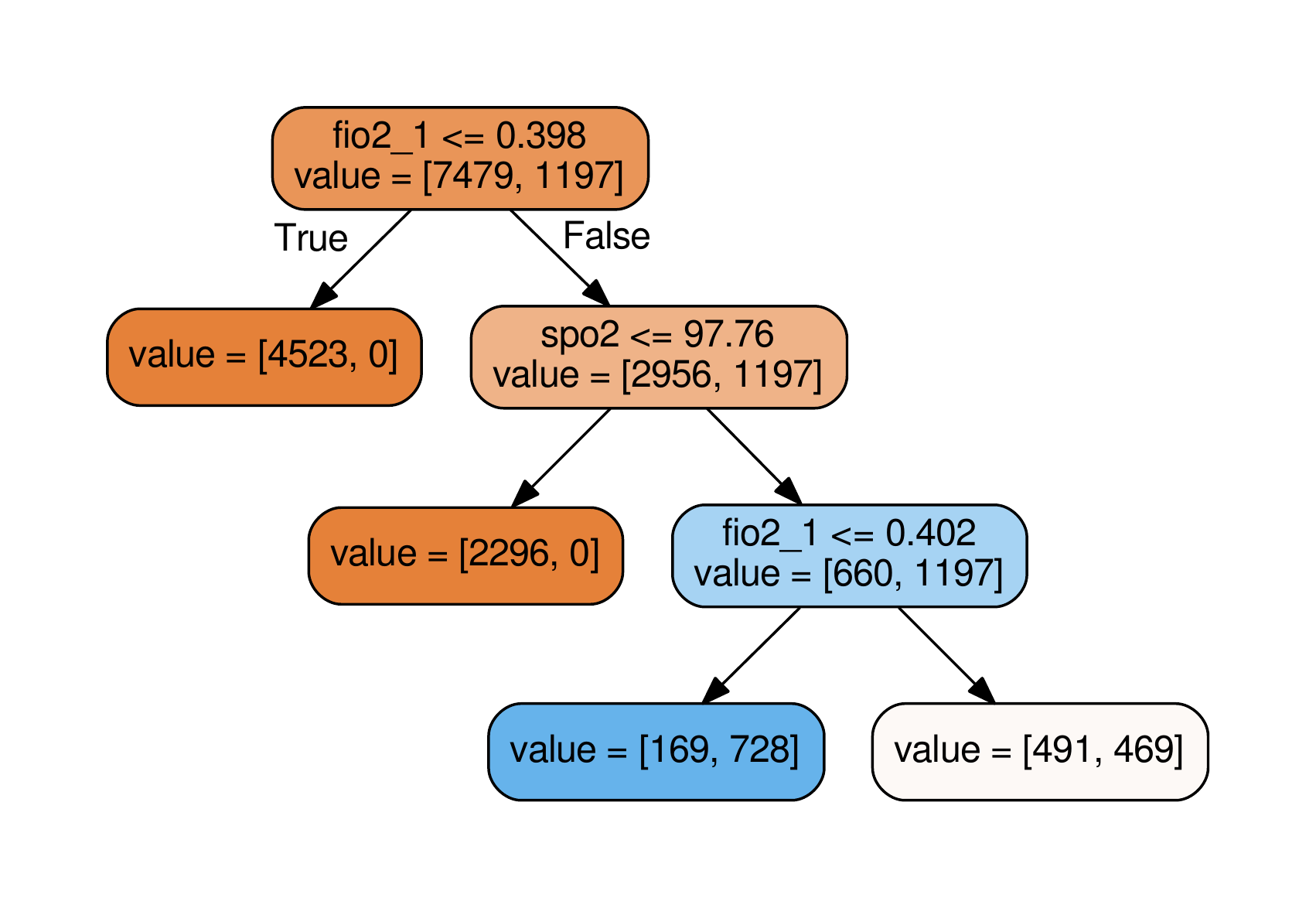}
    \caption{Low SOFA: Vent.}
  \end{subfigure}
  \begin{subfigure}[b]{0.24\textwidth}
    \includegraphics[width=\textwidth]{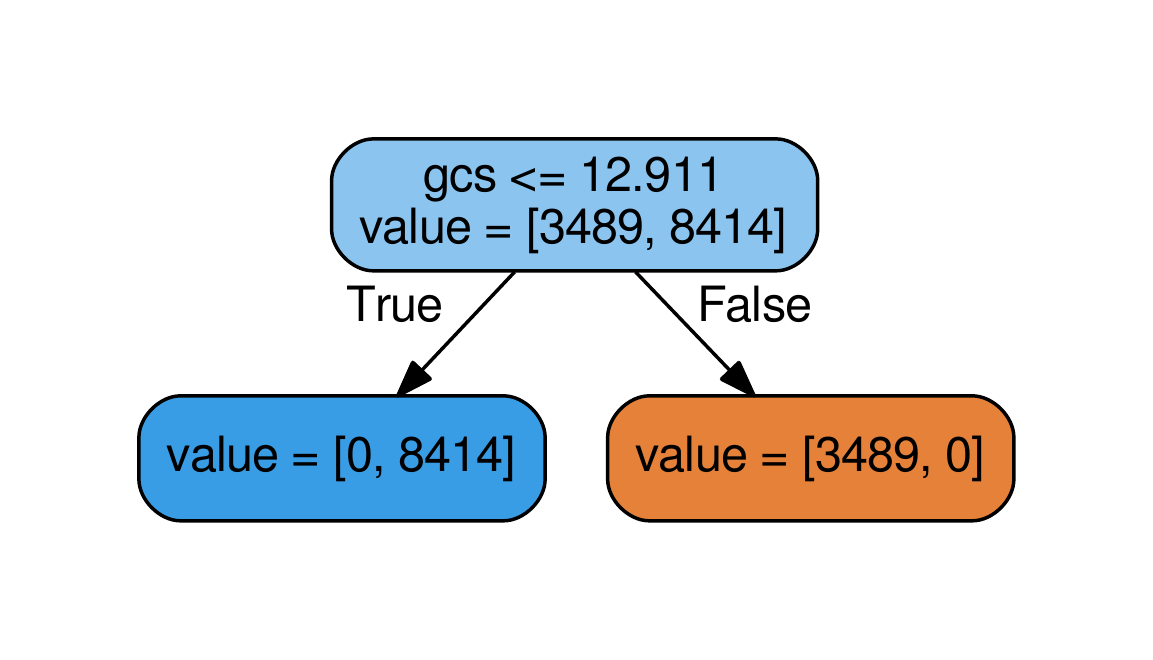}
    \caption{High SOFA: Vent.}
  \end{subfigure}
  \begin{subfigure}[b]{0.24\textwidth}
    \includegraphics[width=\textwidth]{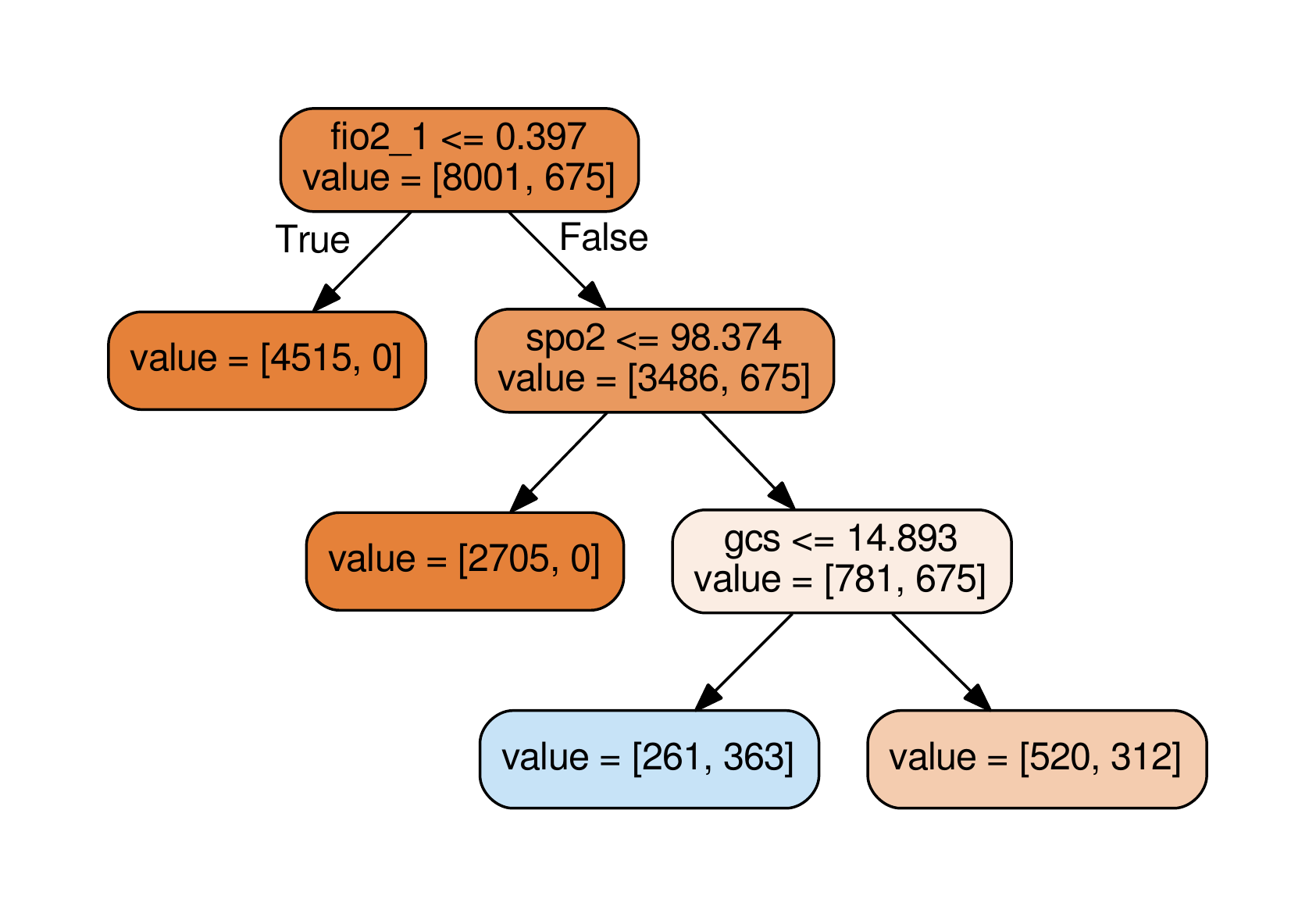}
    \caption{Low SOFA: Sedation}
  \end{subfigure}
  \begin{subfigure}[b]{0.24\textwidth}
    \includegraphics[width=\textwidth]{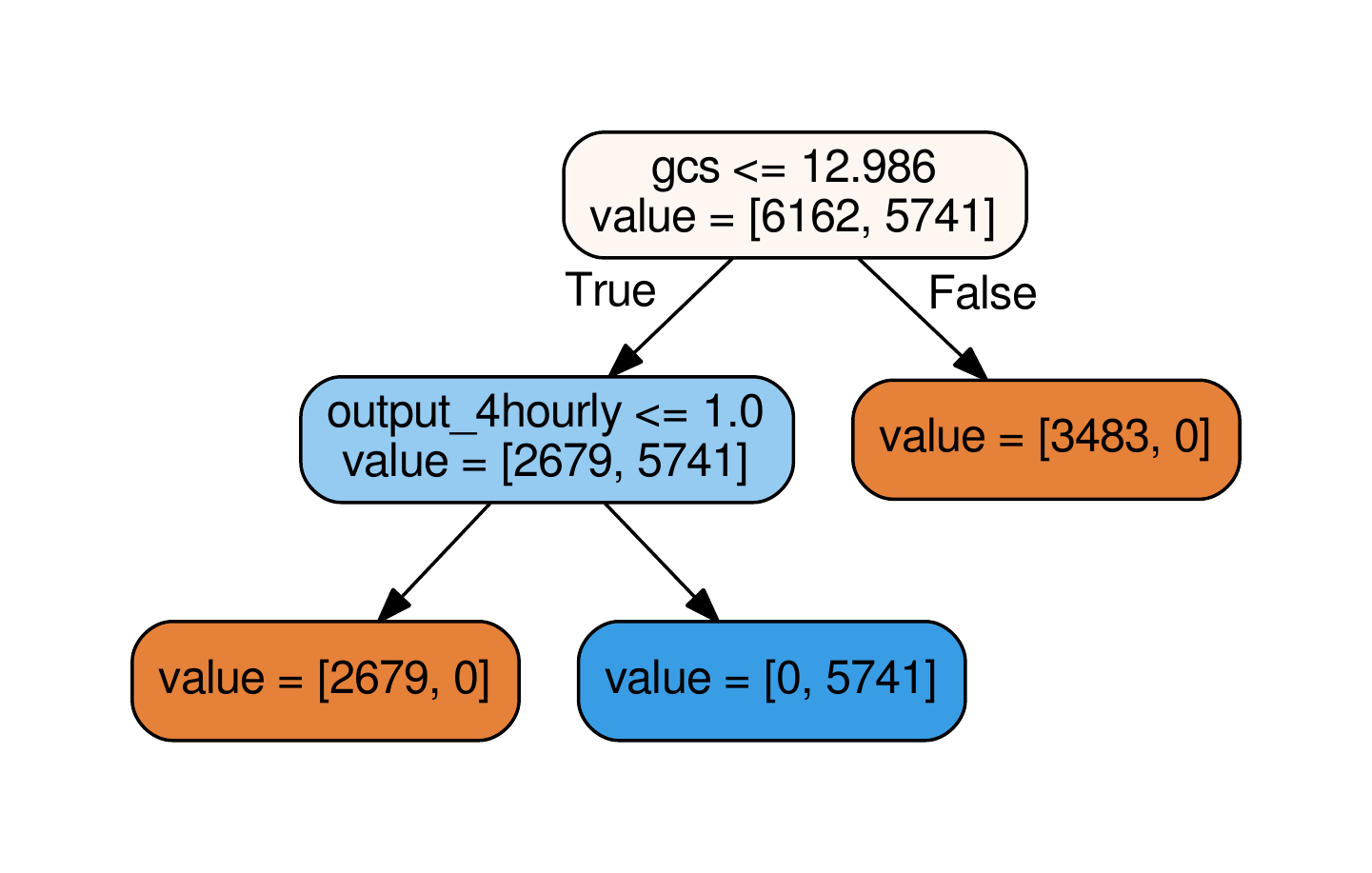}
    \caption{High SOFA: Sedation}
  \end{subfigure}
  \caption{Comparison of regularization methods on the Critical Care dataset. Each output represents a form of medication given in the ICU (e.g. vasopressor, sedation, mechanical ventilation, and renal replacement therapy). Each subfigure compares APL and test accuracy. \emph{(a-d)} compute APL based on three regions defined using SOFA scores; \emph{(e-h)} instead, compute APL on five regions, one for each careunit (e.g. medical vs. surgical ICU). In each experiment, regional tree regularized finds the best performing models at low complexity. Finally, \emph{(i-l)} show distilled decision trees (split by SOFA) that best approximate a regionally regularized target neural model with a low APL and good test accuracy. As confirmed by a physician in the ICU, distilled trees are simulable and capture statistical nuances specific to a region.}
  \label{fig:sepsis}
\end{figure}

\section{Application: Sepsis (ICU)}
We revisit the Sepsis Critical Care dataset, only this time we apply regional tree regularization and compare to other regularizers, including global tree regularization.

\paragraph{APL for multiple outputs.} Previous datasets had only 1 binary output while Critical Care has 5. Fortunately, the definition of APL generalizes: compute the APL for each output dimension, and take the sum as the measure of complexity. This requires fitting $Q \times R$  trees.

\paragraph{Defining regions.} We explore two  methods of defining regions, both suggested by ICU physicians. The first defines three regions by sequential organ failure assessment (SOFA), a summary statistic that has historically been used for predicting ICU mortality. Using $\mathcal{D}$, the groups are defined by more than one standard deviation below the mean, one standard deviation from the mean, and more than one standard deviation above the mean. Intuitively, each group should encapsulate a very different type of patient. The second method clusters patients by the his/her careunit into five groups: MICU (medical), SICU (surgical), TSICU (trauma surgical), CCU (cardiac non-surgical), and CSRU (cardiac surgical). Again, patients who undergo surgery should behave differently than those with less-invasive operations.

\paragraph{Regularization results.} Figure~\ref{fig:sepsis} compares different regularization schemes against baseline models for SOFA regions (a-d) and careunit regions (e-h). Overall, the patterns we discussed in the UCI datasets are consistent in this application. We especially highlight the inability (across the board) of global explanation to find many low complexity solutions. For example, in Figure~\ref{fig:sepsis} (a,c,e), the minima from global constraints stay very close to the unregularized minima. In other cases (f, g), global regularization finds very poor optima: reaching low accuracy with high APL. In contrast, region regularization consistently finds a good compromise between complexity and performance. In each subfigure, we can point to a span of APL at which the pink curve is much higher than all others. These results are from three runs, each with 20 different  strengths.

\paragraph{Distilled decision trees.} A consequence of tree regularization is that every minima is associated with a set of trained trees. We can extract the trees that best approximate the target neural model, and rely on it for explanation. Figure~\ref{fig:sepsis} (i,j) show an example of two trees predicting ventilation plucked from a low APL - high AUC minima of a regional tree regularized model. We note that the composition of the trees are different, suggesting that they each capture a decision function biased to a region. Moreover, we can see that while Figure~\ref{fig:sepsis} (i) mostly predicts 0, Figure~\ref{fig:sepsis} (j) mostly predicts 1; this agrees with our intuition that SOFA scores are correlated with risk of mortality. Figure~\ref{fig:sepsis} (k,l) show similar findings for sedation. If we were to capture this behavior with a single decision tree, we would either lose granularity or be left with a very large tree.

\paragraph{Feedback from physicians.} We presented a set of 9 distilled trees from regional tree regularized models (1 for each output and SOFA region) to an expert intensivist for interpretation.  Broadly, he found the regions beneficial as it allowed him to connect the model to his cognitive categories of patients---including those unlikely to need interventions.  He verified that for predicting ventilation, GCS (mental status) should have been a key factor, and for predicting vasopressor use, the logic supported cases when vasopressors would likely be used versus other interventions (e.g. fluids if urine output is low).  He was also able to make requests: for example, he asked if the effect of oxygen could have been a higher branch in the tree to better understand its effects on ventilation choices, and, noticing the similarities between the sedation and ventilation trees, pointed out that they were correlated and suggested defining new regions by both SOFA and ventilation status.\newline

\noindent We highlight that this kind of reasoning about what the model is learning and how it can be improved is very valuable. Very few notions of interpretability in deep models offer the level of granularity and simulatability as regional tree explanations do.

\begin{figure}[h!]
  \centering
  \begin{subfigure}[b]{0.9\textwidth}
    \includegraphics[width=\textwidth]{v2/uci/legend.pdf}
  \end{subfigure}
  \begin{subfigure}[b]{0.24\textwidth}
    \includegraphics[width=\textwidth]{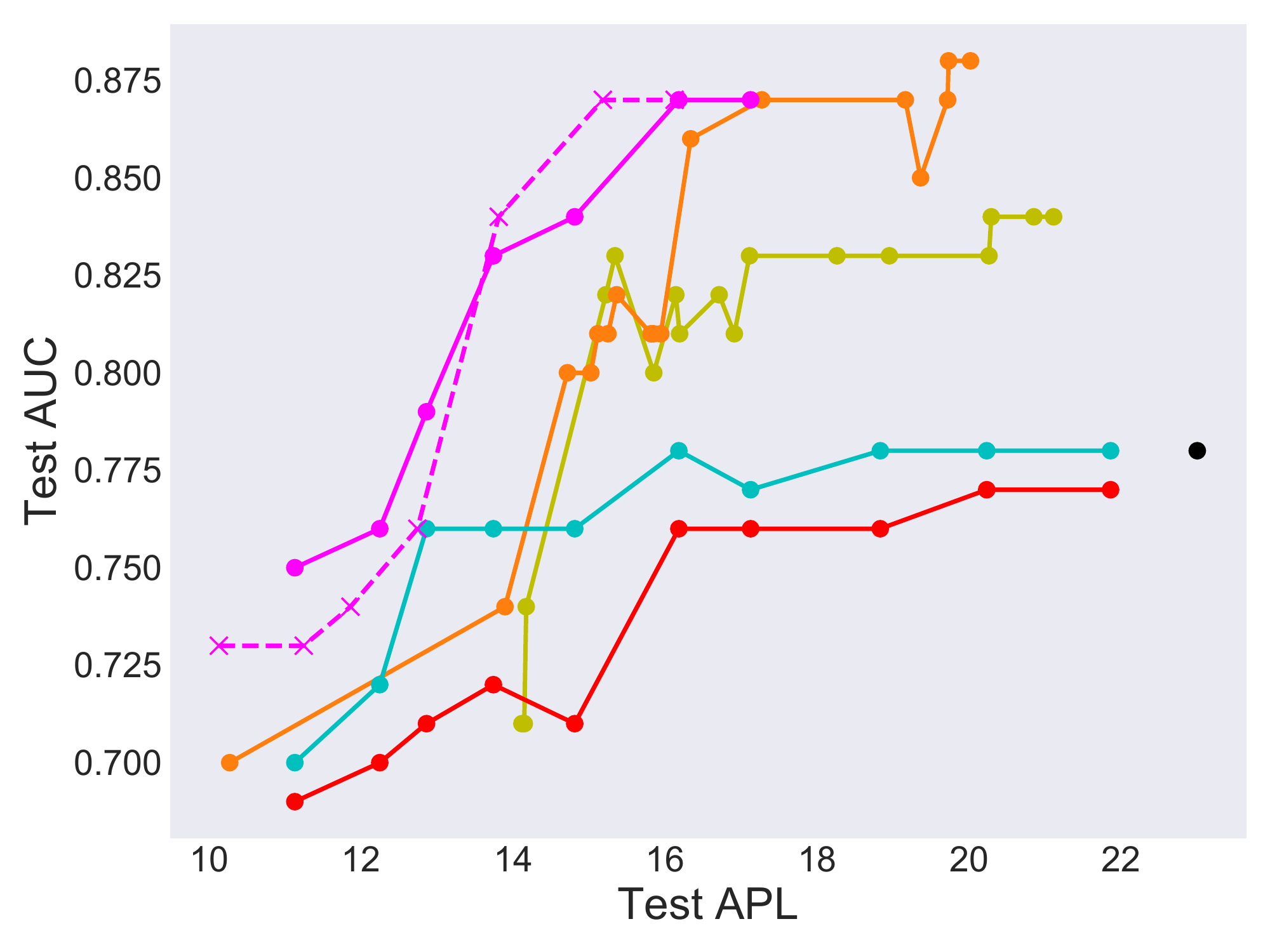}
    \caption{Immunity: Mortality}
     \label{mortalsubfig}
  \end{subfigure}
  \begin{subfigure}[b]{0.24\textwidth}
    \includegraphics[width=\textwidth]{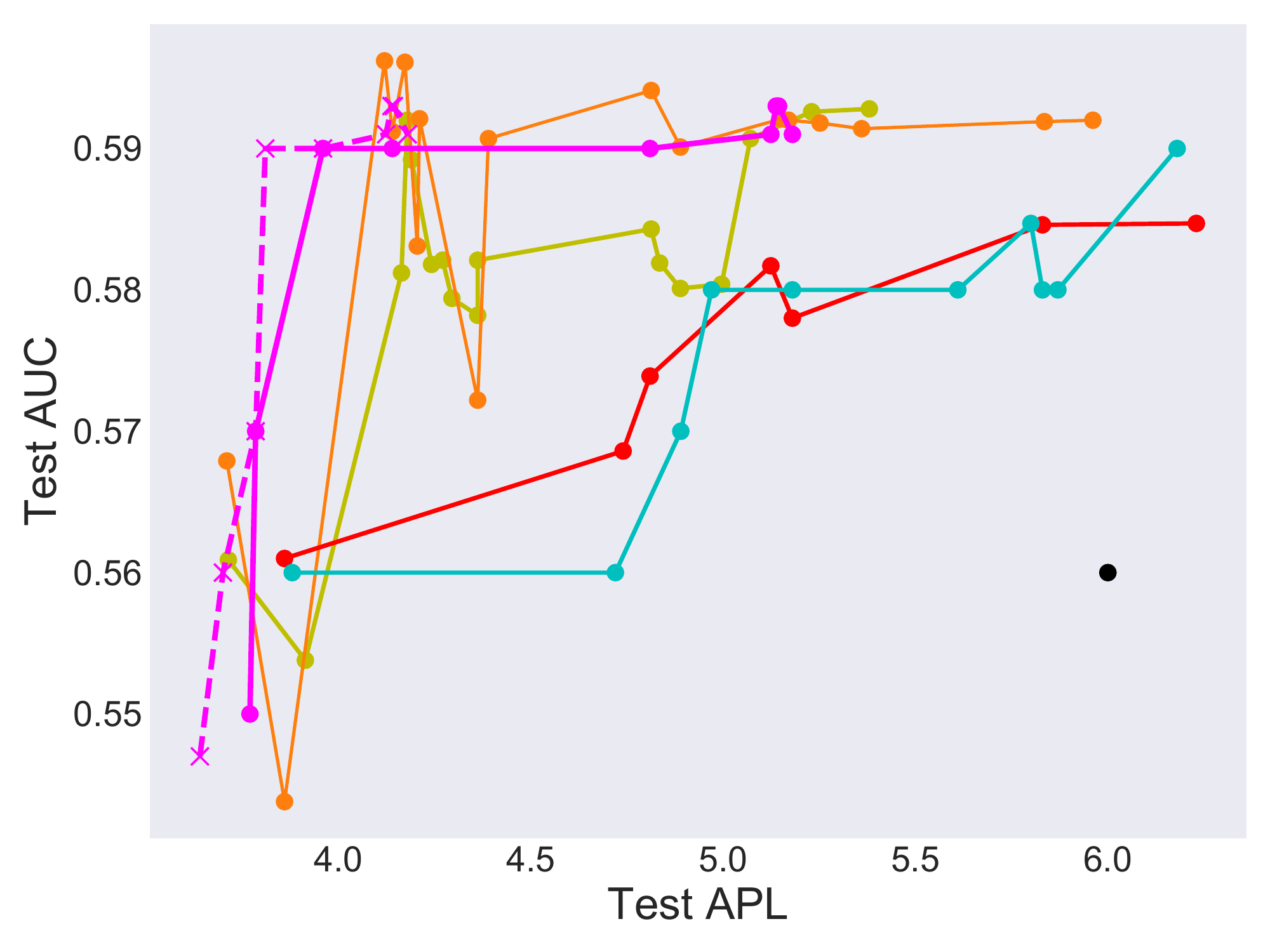}
    \caption{Immunity: AIDS Onset}
  \end{subfigure}
  \begin{subfigure}[b]{0.24\textwidth}
    \includegraphics[width=\textwidth]{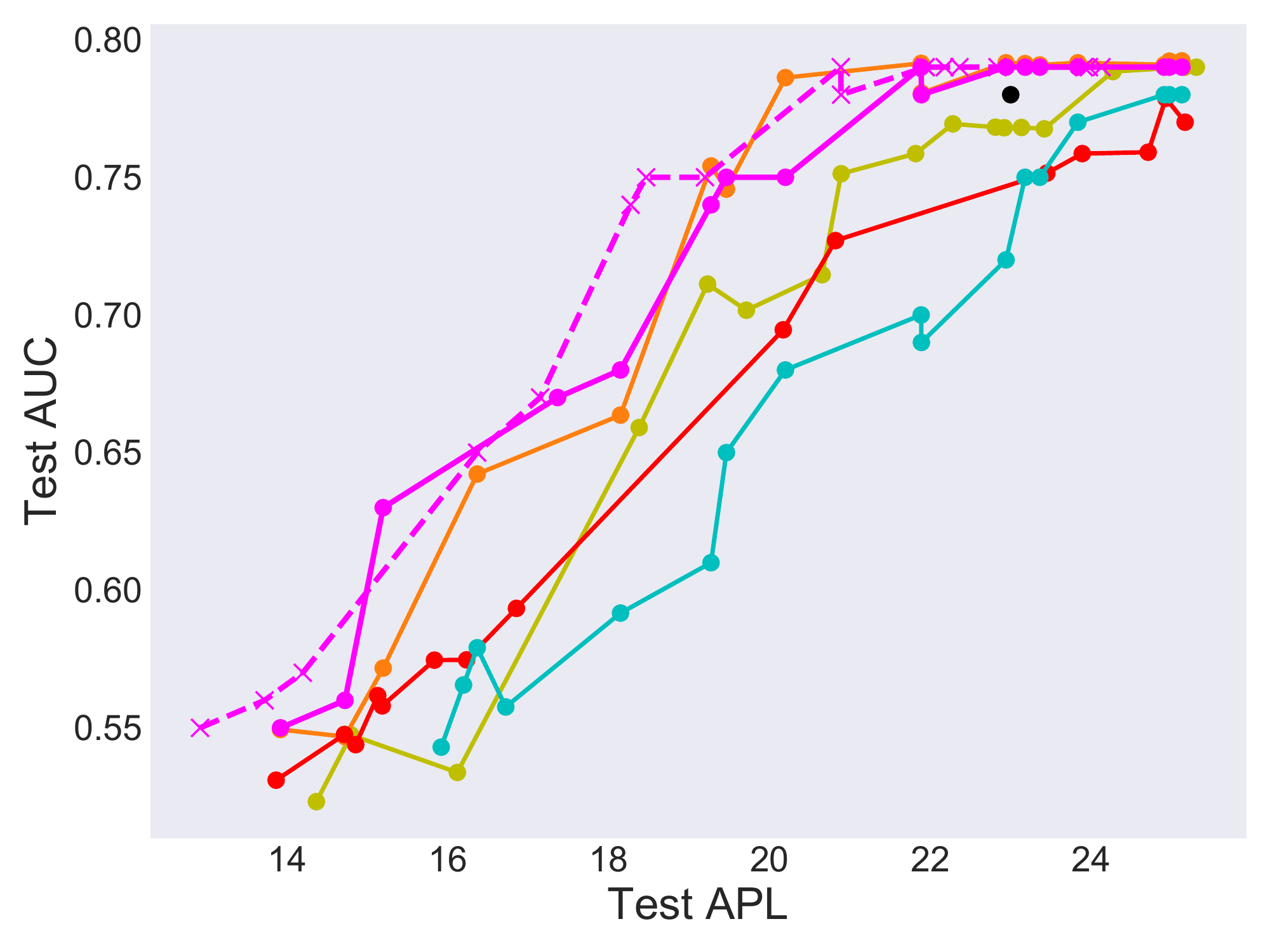}
    \caption{Immunity: Adherence}
  \end{subfigure}
  \begin{subfigure}[b]{0.24\textwidth}
    \includegraphics[width=\textwidth]{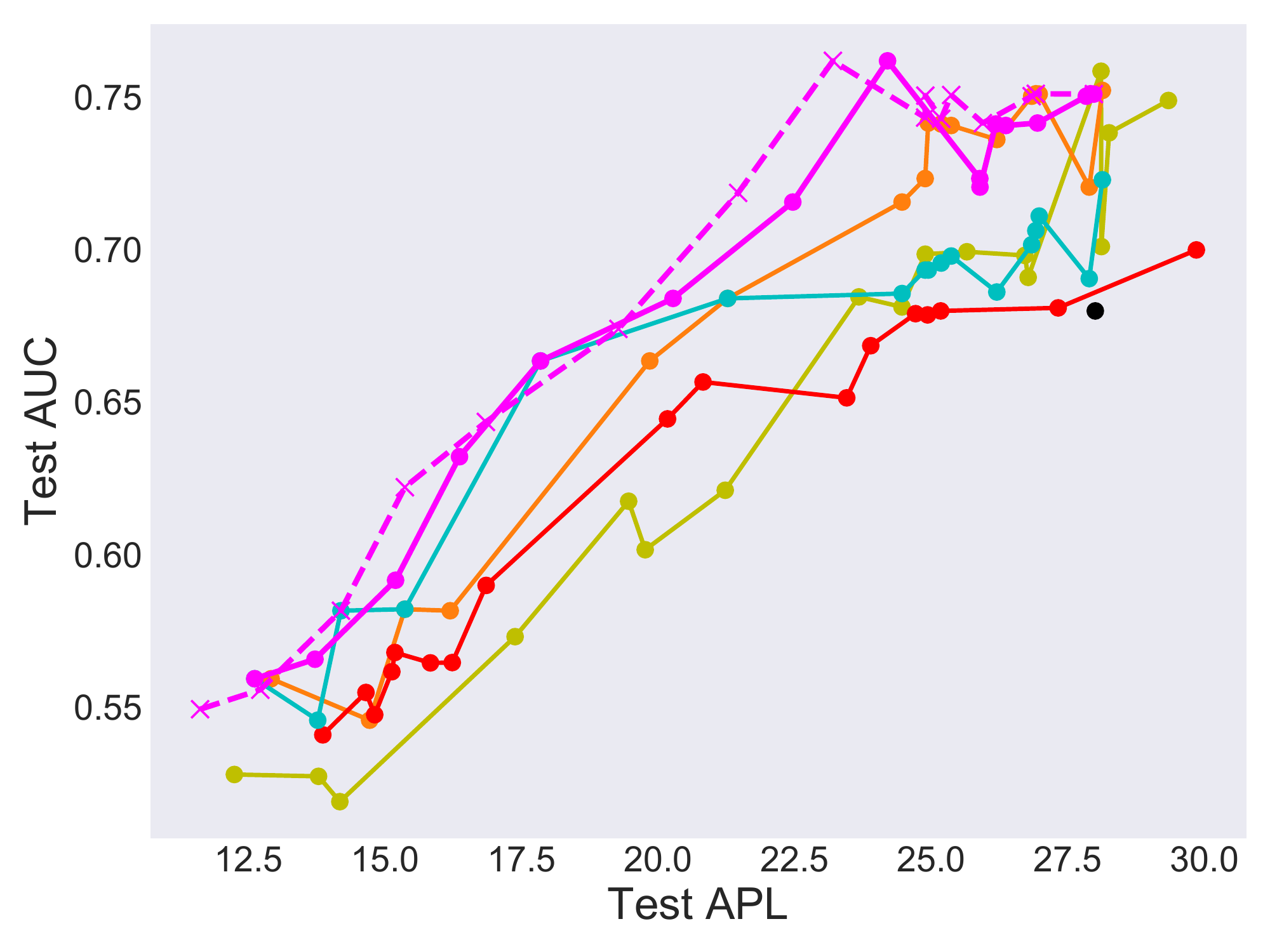}
    \caption{Immunity: Viral Suppression }
    \end{subfigure}
     \begin{subfigure}[b]{0.32\textwidth}
    \includegraphics[width=\textwidth]{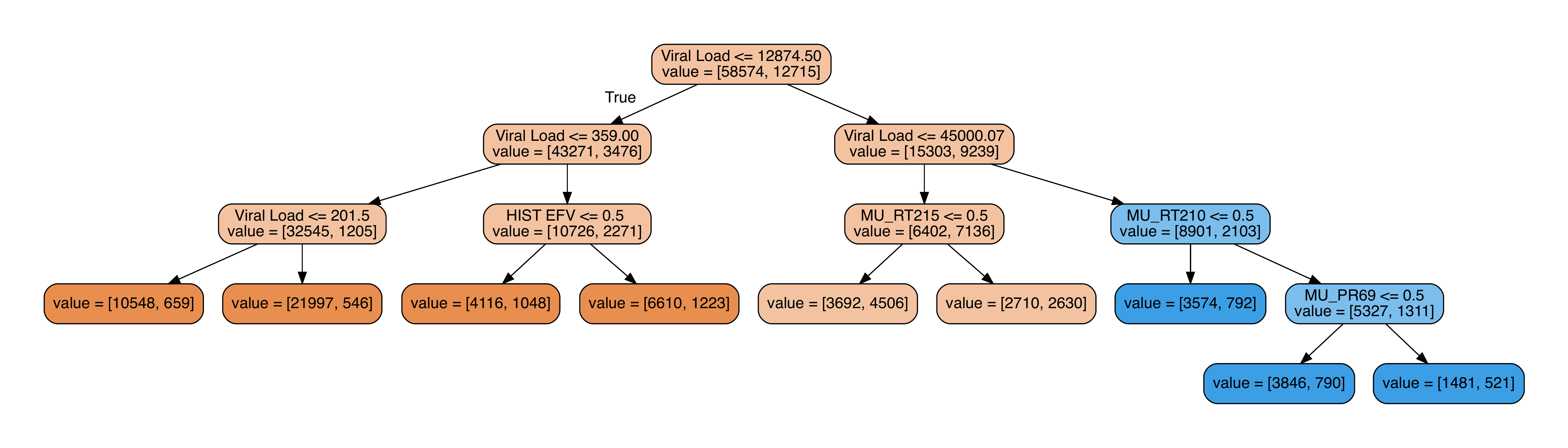}
    \caption{High Immunity: Mortality}
  \end{subfigure}
  \begin{subfigure}[b]{0.32\textwidth}
    \includegraphics[width=\textwidth]{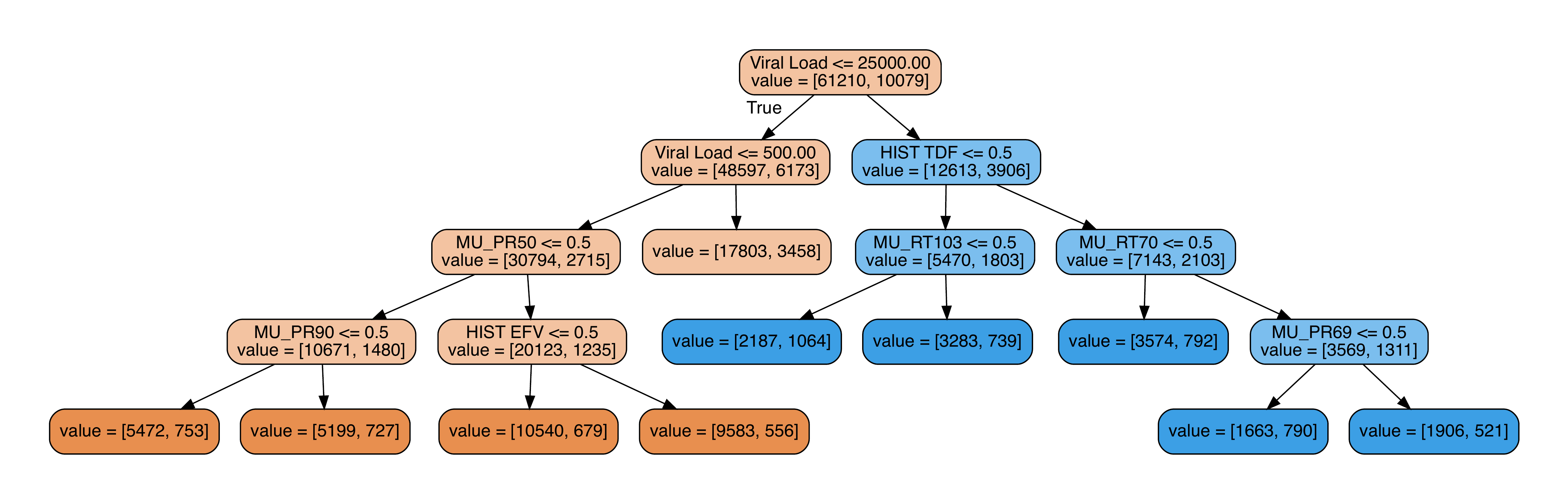}
    \caption{Mid Immunity: Mortality}
  \end{subfigure}
  \begin{subfigure}[b]{0.32\textwidth}
    \includegraphics[width=\textwidth]{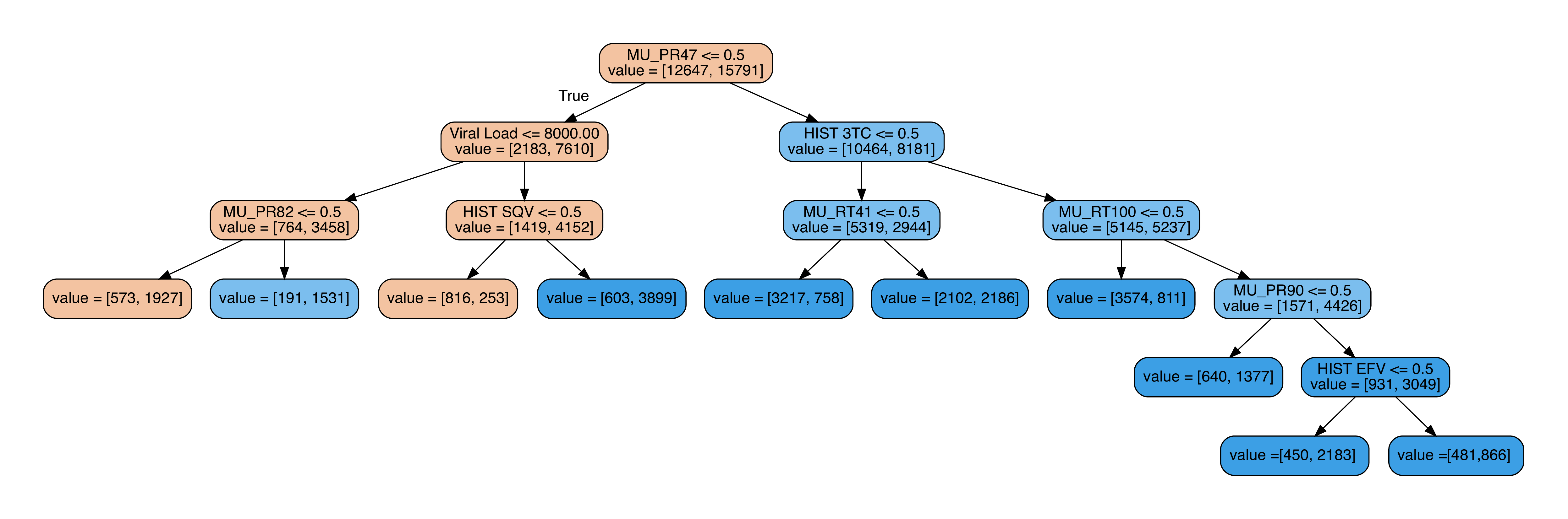}
    \caption{Low Immunity: Mortality}
  \end{subfigure}
  \caption{Comparison of regularization methods on 15 output dimensions of the HIV dataset (4 of which are shown). Each subfigure compares APL and test accuracy. Subfigures (a-d) base the metric on four regions corresponding to the level of immunosuppression (abbreviated to immunity) at baseline (e.g. $<$200 cells/mm$^3$).  Subfigures (e-g) show distilled decision trees (split by degrees of immunity) that best approximate a regionally regularized target neural model with a low APL.}
  \label{fig:hiv}
\end{figure}

\section{Application: EuResist (HIV)}
We again revisit the HIV dataset to compare global and regional explanations.

\paragraph{Defining regions in HIV.} We define regions based on the advice of medical experts. This is performed using a patient's degree of immunosuppression at baseline (known as CDC staging). These groups are defined as: $< $200 cells/mm$^3$, 200 - 300 cells/mm$^3$, 300 - 500 cells/mm$^3$ and $>$500 cells/mm$^3$ \cite{world2005interim}. This choice of regions should characterize patients based on the initial severity of their infection; the lower the initial cell count, the more severe the infection.

\paragraph{Regularization results.} Figure~\ref{fig:hiv} compares different regularization schemes against baseline models across levels of immunosuppression. Overall, regional tree regularization produces more accurate predictions and provides simpler explanations across all outputs. For the case of predicting patient mortality in Fig~\ref{mortalsubfig}, we tend to find more suitable optima across different patient groupings and can provide better regional explanations for these patients as a result. Here, we observe that patients with lower levels of immunosuppression tend to have lower risk of mortality. We also observe that patients with lower immunity at baseline are more likely to progress to AIDS. Similar inferences can be made for the other outputs. In each subfigure, we reiterate that there is a span of APL at which the pink curve is much higher than all others.

\paragraph{Distilled decision trees.} We extract decision trees that approximate the target model for multiple minima and use these as explanations. Fig~\ref{fig:hiv} (e-g) show three trees where we have low APL and high AUC minima from a regional tree regularized model. Again, the trees look significantly different based on the decision function in a particular region. In particular, we observe that lower levels of immunity at baseline are associated with higher viral loads (lower viral suppression) and higher risk of mortality.

\paragraph{Feedback from physicians.} The trees were shown to a physician specializing in HIV treatment.  He was able to simulate the model's logic, and confirmed our observations about relationships between viral loads and mortality. In addition, he noted that when patients have lower baseline immunity, the trees for mortality contain several more drugs. This is consistent with medical knowledge, since patients with lower immunity tend to have more severe infections, and require more aggressive therapies to combat drug resistance.

\section{Analysis for Regional Tree Regularization}

We now summarize a few important outcomes from the regional experiments:

\paragraph{The most effective minima are found in the low APL, high AUC regime.} The ideal model is one that is highly performant and simulable. This translates to high F1/AUC scores near medium APL. Too large of an APL would be hard for an expert to understand. Too small of an APL would be too restrictive, resulting in no benefit from using a deep model. Across all experiments, we see that L$_0$ region regularization is most adept at finding low APL and high AUC minima.

\paragraph{Global and local regularization are two extreme forms of regional regularization.}
If $R=1$, the full training dataset is contained in a single region, enforcing global explainability. If $R=N$, then every data point $\mathbf{x}_n \in \mathcal{D}$ has its own region i.e. local explainability.

\begin{table}[h!]
\centering
\begin{tabular}{c c c c c c c}
\toprule
 & Bank & Gamma & Adult & Wine & Crit. Care & HIV \\
\midrule
Fidelity & 0.892 & 0.881 & 0.910 & 0.876 & 0.900 & 0.897 \\
\bottomrule
\end{tabular}
\caption{Fidelity is the percentage of examples on which the prediction made by a tree agrees with the deep model \cite{craven1996extracting}. }
\label{table:fidelity}
\end{table}

\paragraph{Regularized deep models outperform trees.} Comparing regional tree-regularized models and regional decision trees, the former reach much higher AUC at equal APL.

\paragraph{Regional tree regularization produces regionally faithful decision trees.} Table~\ref{table:fidelity} shows the fidelity of a deep model to its distilled tree. A score of 1.0 indicates that both models learned the same decision function. With a fidelity of 89\%, the regularized model is ``simple" in most cases, but can take advantage of deep nonlinearity with difficult examples.

\paragraph{Regional tree regularization is not computationally expensive.}
Over 100 trials on Sepsis, an L2 model takes $2.393 \pm 0.258$ sec. per epoch; a global tree model takes $5.903 \pm 0.452$ sec. and $21.422\pm0.619$ sec. to (1) sample 1000 convex samples, (2) compute APL for $\mathcal{D}^\theta$, (3) train a surrogate model for 100 epochs; a regional tree model takes $6.603\pm0.271$ sec. and $39.878\pm0.512$ sec. for (1), (2), and training 5 surrogates. The increase in base cost is due to the extra forward pass through $R$ surrogate models to predict APL. The surrogate cost(s) are customizable depending on the size of $\mathcal{D}^\theta$, the number of training epochs, and the frequency of re-training. If $R$ is large, we need not re-train each surrogate. The choice of which regions to prioritize can be treated as a bandit problem.

\paragraph{Distilled decision trees are interpretable by domain experts.} We asked physicians in Critical Care and HIV to analyze the distilled decision trees from regional regularization. They were able to quickly understand the learned decision function per region, suggest improvements, and verify the logic.

\paragraph{Optimizing surrogates is much faster and more stable than gradient-free methods.} We tried alternative optimization methods that do not require differentiating through training a decision tree: (1) estimate gradients by perturbing inputs, (2) search algorithms like Nelder-Mead. However, we found these methods to either be unreasonably expensive, or easily stuck in local minima based on initialization.

\paragraph{Sparsity over regions is important.} We experimented with different ``dense" norms: L$_1$, L$_2$, and a softmax approximation to L$_0$, all of which faced issues where regions with simpler decision boundaries a priori were over-regularized to trivial decision functions. Only with L$_0$ (i.e. \texttt{sparsemax}) did we avoid this problem. As a consequence, in toy examples, we observe that \texttt{sparsemax} finds minima with more axis-aligned boundaries. In real world studies, we find \texttt{sparsemax} to lead to better performance in low/mid APL regimes.

\section{Conclusion}
Interpretability is a bottleneck preventing widespread acceptance of deep learning. We have introduced a family of novel tree-regularization techniques that encourages the complex decision boundaries of any differentiable model to be well-approximated by human-simulable
functions, allowing domain experts to quickly understand and approximately compute what the model is doing. Overall, our training procedure is robust and efficient. Across three complex, real-world domains (HIV treatment, sepsis treatment, and human speech processing) our tree-regularized models provide gains in prediction accuracy in the regime of simpler, human-simulatable
models. Finally, we then showed how to extend tree regularization to more regional-specific approximations of a loss, where experts can add prior knowledge about the structure of their domain. More broadly, our general training procedure could apply tree-regularization or other procedure-regularization to a wide class of popular models, helping us move beyond sparsity toward models humans can easily simulate and thus trust.
\bibliography{interp_draft}
\bibliographystyle{theapa}

\end{document}